\documentclass[a4paper,fleqn]{cas-sc}
\usepackage[round,authoryear]{natbib}
\usepackage{amsmath}
\usepackage[linesnumbered, ruled]{algorithm2e}
\SetKwRepeat{Do}{do}{while}%
\usepackage{soul}
\usepackage{bm}
\usepackage{color, xcolor}

\newcounter{subsubsubsection}[subsubsection]
\renewcommand{\thesubsubsubsection}{\thesubsubsection.\arabic{subsubsubsection}}
\newcommand{\subsubsubsection}[1]{%
  \par\addvspace{1ex plus .1ex}%
  \refstepcounter{subsubsubsection}%
  \noindent{\normalfont\normalsize\itshape
    \thesubsubsubsection\hspace{1em}#1}\par
  \addvspace{1ex plus .1ex}\nobreak\noindent\ignorespaces}
\usepackage[normalem]{ulem} 

\soulregister\cite7
\soulregister\citep7
\soulregister\ref7

\usepackage{graphicx}
\usepackage{subcaption}
\usepackage{caption}
\usepackage{float}      
\usepackage{placeins}   

\usepackage{multirow}
\usepackage{rotating}
\usepackage{makecell}

\usepackage{longtable}
\usepackage{siunitx}
\usepackage{tabularx}
\usepackage{diagbox}
\def\tsc#1{\csdef{#1}{\textsc{\lowercase{#1}}\xspace}}
\tsc{WGM}
\tsc{QE}

\begin{document}
\let\WriteBookmarks\relax

\setcounter{topnumber}{5}
\setcounter{bottomnumber}{5}
\setcounter{totalnumber}{10}

\renewcommand{\topfraction}{0.95}     
\renewcommand{\bottomfraction}{0.90}  
\renewcommand{\textfraction}{0.05}    
\renewcommand{\floatpagefraction}{0.85} 

\setlength{\floatsep}{6pt plus 2pt minus 2pt}
\setlength{\textfloatsep}{8pt plus 2pt minus 2pt}
\setlength{\intextsep}{6pt plus 2pt minus 2pt}

\makeatletter
\def\fps@table{tbp}
\makeatother

\shortauthors{Yilin Wang et~al.}
\shorttitle{}


\title [mode = title]{RoboSense: Leveraging Robotaxi Fleets as Drive-by Sensors for Urban Traffic Monitoring}

\author[1]{Yilin Wang}[style=chinese]
\ead{wang4517@purdue.edu}

\author[1]{Yiheng Feng}[style=chinese]
\cormark[1]
\ead{feng333@purdue.edu}

\affiliation[1]{organization={Lyles School of Civil and Construction Engineering, Purdue University},
  city={West Lafayette},
  country={USA}}

\cortext[1]{Corresponding author.}

\begin{abstract}
\noindent
Urban traffic monitoring plays a critical role in safety analysis, congestion management, and incident response. The growing deployment of robotaxis creates a new opportunity for network-level traffic monitoring. Although robotaxis are primarily designed to serve passengers, they can also be leveraged as drive-by sensors to collect traffic data. Compared to conventional infrastructure sensors or probe vehicles, a fleet of robotaxis forms a cooperative perception environment, which can collectively gather spatially and temporally continuous traffic information. This paper proposes a novel dynamic robotaxi routing framework that explicitly incorporates traffic monitoring tasks as an objective. The framework introduces: (1) a cell-based network representation that aligns with sensing capabilities of robotaxis; (2) a cell-level monitoring metric to quantify spatiotemporal robotaxi coverage; and (3) a mixed-integer linear programming (MILP) formulation that jointly minimizes time-dependent travel time and maximizes traffic monitoring performance. A $5\times 5$ urban grid network is built in SUMO to evaluate the framework under three robotaxi market penetration rates (2\%, 5\%, and 10\%) with a range of objective weight combinations. Results show that incorporating spatiotemporal network coverage in the objective function can effectively improve the traffic monitoring performance. Interestingly, with appropriate weights between the two objectives, monitoring performance and robotaxi average speed can be improved simultaneously. This suggests better network monitoring leads to more accurate traffic state prediction and improved mobility. This win-win situation could incentivize robotaxi operators to contribute their vehicles as drive-by sensors for traffic monitoring. 
\end{abstract}

\date{}
\maketitle

\section{Introduction}

Traffic monitoring is a critical component of Intelligent Transportation Systems (ITS), supporting a wide range of operational and planning applications such as safety analysis \citep{yuan2018utilizing}, congestion management \citep{zhang2011data, christofa2013arterial}, and emission control \citep{guerrero2018sensor}. Owing to the inherently spatiotemporal nature of traffic dynamics, effective monitoring requires sensing mechanisms that jointly provide adequate temporal and spatial coverage. Conventional infrastructure-based sensing approaches, including inductive loop detectors and roadside surveillance cameras, offer temporally continuous measurements but remain constrained by their fixed deployment locations. As a result, they often exhibit limited spatial coverage, high deployment and maintenance costs, and reduced performance in capturing spatiotemporal traffic dynamics.

To overcome limitations of fixed-location sensors, recent studies have explored using probe vehicles for traffic monitoring \citep{herrera2010evaluation, zheng2017estimating}. Unlike stationary detectors, probe vehicles provide spatially continuous trajectories, enabling richer and more detailed traffic information across the network. However, probe vehicle data still suffer from two major limitations. First, probe vehicles only report their own trajectories. As a result, a sufficiently high market penetration rate (MPR) is required to ensure adequate spatiotemporal coverage for reliable traffic state estimation and imputation (e.g., speed, density, and travel time reconstruction) \citep{comert2009queue, zhao2019estimation}. Second, probe vehicle data are typically obtained from private vehicles through commercial providers such as INRIX \citep{cookson2017inrix} and TomTom \citep{tomtom2023index}, and their movements are primarily driven by personal travel needs and cannot be controlled. 

Recent deployment of robotaxis creates a new opportunity to solve these challenges. Robotaxis are equipped with onboard perception sensors and are capable of perceiving surrounding traffic with high precision. They can share perception data in real time, which creates a cooperative perception environment \citep{xiang2024v2x,chen2022cooperative}. Compared with probe vehicles which only report their own status, cooperative perception based traffic monitoring requires a much lower MPR. Moreover,
unlike individually operated vehicles, robotaxis operate as centrally managed fleets. Their routing behavior can be coordinated by the fleet operator to enhance spatiotemporal data coverage for traffic monitoring, while still meeting passenger service requirements (e.g., travel time). As a result, traffic monitoring can be transformed into a purposefully managed data-collection effort, rather than purely as a byproduct of individual vehicle travel behavior. 



\begin{figure}[pos=tbp]
    \centering
    \begin{subfigure}[b]{0.45\textwidth}
        \centering
        \includegraphics[width=\textwidth]{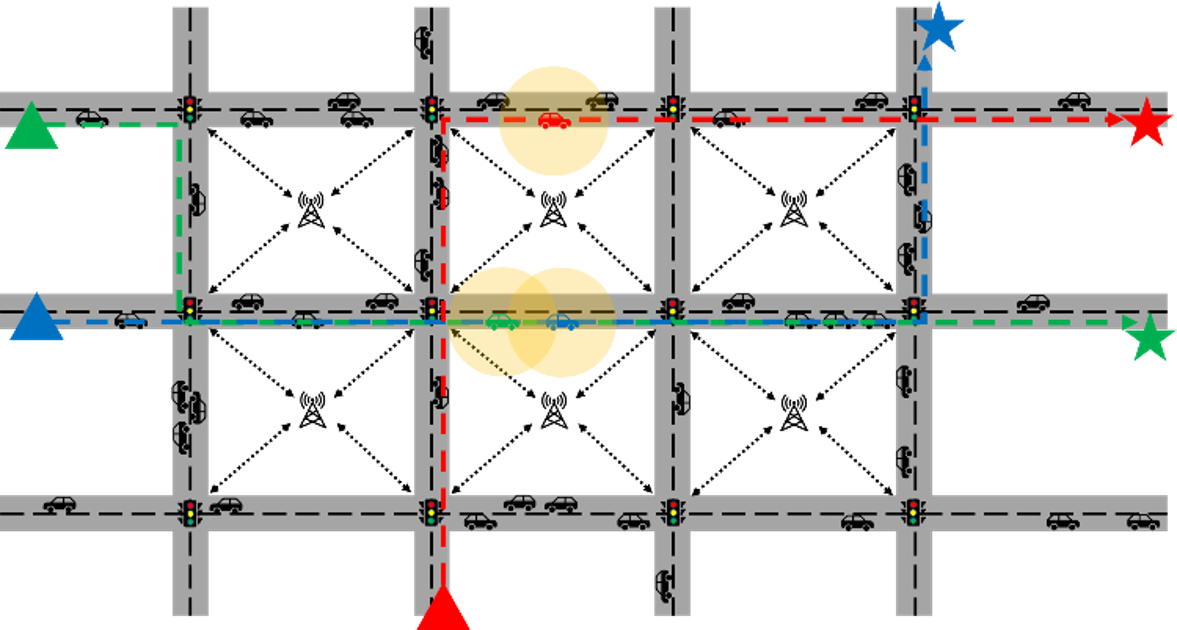}
        \caption{Fleet routing based on travel time}
        \label{fig:route_pattern1}
    \end{subfigure}
    \hfill
    \begin{subfigure}[b]{0.45\textwidth}
        \centering
        \includegraphics[width=\textwidth]{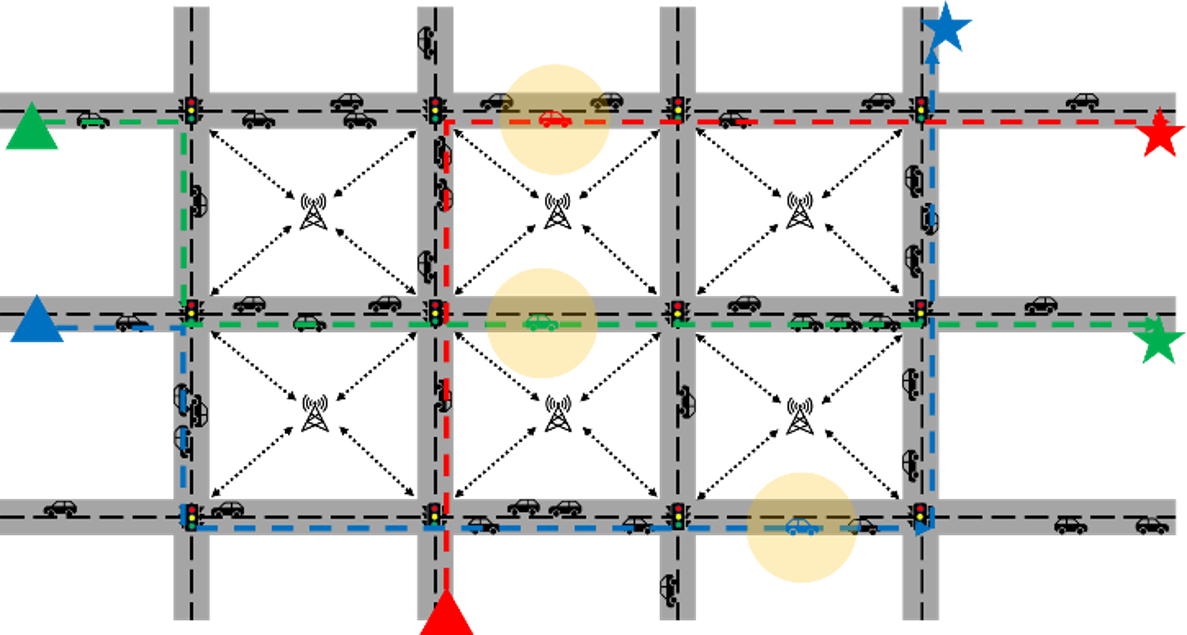}
        \caption{Fleet routing considering traffic monitoring}
        \label{fig:route_pattern2}
    \end{subfigure}
    \caption{An illustrative example on routing selection with different considerations}
    \label{fig:route_pattern}
\end{figure}

When traffic monitoring becomes an explicit objective, fleet mobility (e.g., travel time or distance) and information acquisition (e.g., spatiotemporal coverage) are no longer separable. Decisions regarding where and when robotaxis travel directly determine which parts of the network are observed, making vehicle routing a central mechanism through which monitoring performance is shaped. Figure \ref{fig:route_pattern} provides an intuitive example. Assuming three robotaxi vehicles enter and leave the network at the same time, Figure \ref{fig:route_pattern1} shows the routing pattern when all three vehicles choose the shortest path in terms of travel time or distance. During a certain time period, the green and the blue vehicles travel on the same link at the same time, generating redundant traffic data, and leaving the bottom corridor unobserved. However, if the blue vehicle chooses an alternative route by turning right when entering the network as shown in Figure \ref{fig:route_pattern2}, the bottom corridor can also be observed when the blue vehicle passes through, which improves monitoring performance by increasing spatiotemporal coverage. In order to integrate the traffic monitoring task with robotaxi routing operations, the following challenges need to be addressed.

\begin{itemize}

\item \textbf{Network Representation:} 
Most existing vehicle routing methods adopt either path-based \citep{bezzi2023route} or link-based \citep{ran1996link, grosso2018mathematical} network representations, as comprehensively reviewed in \citep{munari2016generalized}. However, such representations are insufficient for robotaxi-based traffic monitoring due to the spatial mismatch between sensing resolution and network abstraction. Empirical studies suggest that typical onboard sensing ranges of autonomous vehicles are on the order of $60\,\mathrm{m}$ to $120\,\mathrm{m}$ \citep{Arnold2019_3ddetection, xu2023v2v4real}, which are significantly shorter than the length of a typical road link in urban networks. Consequently, when a robotaxi traverses a link, only a localized portion of that link is observed. Accurately capturing which parts of the network are observed at any time therefore requires a higher-resolution network representation, which can explicitly reflect partial link observations.

\item \textbf{Performance Metric for Monitoring:} 
While minimizing travel time and/or travel distance remains the dominant objective in existing vehicle routing formulations, traffic monitoring needs a new performance metric. Such a metric should (i) directly reflect the quality of traffic monitoring rather than travel efficiency, (ii) be computable based on observations collected by individual robotaxis, and (iii) aggregate meaningfully to characterize monitoring performance at the network level. Designing metrics that simultaneously satisfy these requirements is essential, as monitoring effectiveness depends on both the spatiotemporal distribution of observations and their collective coverage of the network.

\item \textbf{Time-Dependent Dynamic Decision Making:} 
Routing robotaxi fleets for traffic monitoring is inherently a dynamic decision-making problem. Routing decisions must be updated over time to adapt to evolving traffic conditions, such as time-dependent travel times and the current network observability. This leads to a high-dimensional decision space spanning time, vehicles, and network states. When combined with multiple objectives, the resulting problem becomes significantly more complex. 

\end{itemize}

In this paper, we develop a dynamic routing framework for robotaxi fleet operations that integrates traffic monitoring as an explicit network level objective. The proposed framework jointly models vehicle mobility, network observability, and traffic dynamics through a closed-loop interaction among: \textit{Microscopic Simulation}, \textit{Traffic State Estimation} and \textit{Dynamic Vehicle Routing}. The spatiotemporal distribution of observable network regions is determined by the real-time locations of robotaxis and their sensing ranges provided by the microscopic simulator, while the traffic states are continuously predicted using vehicle-based observations. These predicted states, in turn, inform robotaxi routing decisions by capturing the evolving conditions of the network.

The dynamic routing algorithm is solved repeatedly over time to adapt to changes in both traffic states and network observability.
Numerical experiments in an urban grid network illustrate how incorporating traffic monitoring tasks into robotaxi routing alters fleet behaviors. The results demonstrate that routing strategies accounting for monitoring objectives can substantially enhance spatiotemporal network coverage compared to routing decisions only considering travel time. 
Interestingly, with appropriate weighting of the monitoring objective, the robotaxi fleet’s average speed also increases, indicating potential mobility benefits that could incentivize robotaxi operators to support the traffic monitoring task.
These findings highlight the potential of coordinated robotaxi fleets to serve as effective drive-by sensors for urban traffic monitoring.

To the best of our knowledge, this is among the earliest studies to explicitly integrate traffic monitoring objectives into dynamic routing for robotaxi fleets. The main contributions of this paper are summarized as follows:

\begin{enumerate}
    \item We propose an integrated optimization framework that embeds traffic monitoring performance as an explicit objective within dynamic robotaxi fleet routing, capturing the structural coupling between vehicle mobility and network observability.

    \item We adopt a cell-based network representation to jointly support vehicle-level sensing and detection, traffic state estimation and prediction, and monitoring performance evaluation, and design a spatiotemporal monitoring metric based on cell-level observability over time.

    \item We demonstrate the proposed framework in a microscopic simulation environment and evaluate its performance under a range of robotaxi MPR scenarios, illustrating how fleet-level routing decisions can influence both monitoring effectiveness and travel efficiency.
\end{enumerate}

The rest of this paper is organized as follows. Section~\ref{sec:literature review} reviews related literature on traffic monitoring and vehicle routing. Section~\ref{sec:methodology} presents the proposed methodology, including an overview of the proposed framework, cell-based traffic state estimation and prediction, and the formulation of the dynamic robotaxi routing problem. Section~\ref{sec:case study} introduces a case study based on an urban grid network and evaluates the proposed framework under different robotaxi fleet MPRs and objective weights. Section~\ref{sec:implication} discusses the implications of the results and potentials for real-world implementation. Finally, Section~\ref{sec:Conclusion and Future Work} concludes the paper and outlines directions for future research.




\section{Literature Review}
\label{sec:literature review}
\subsection{Traffic Monitoring}

\begin{figure}[pos=tbp]
    \centering
    \includegraphics[width=0.8\linewidth]{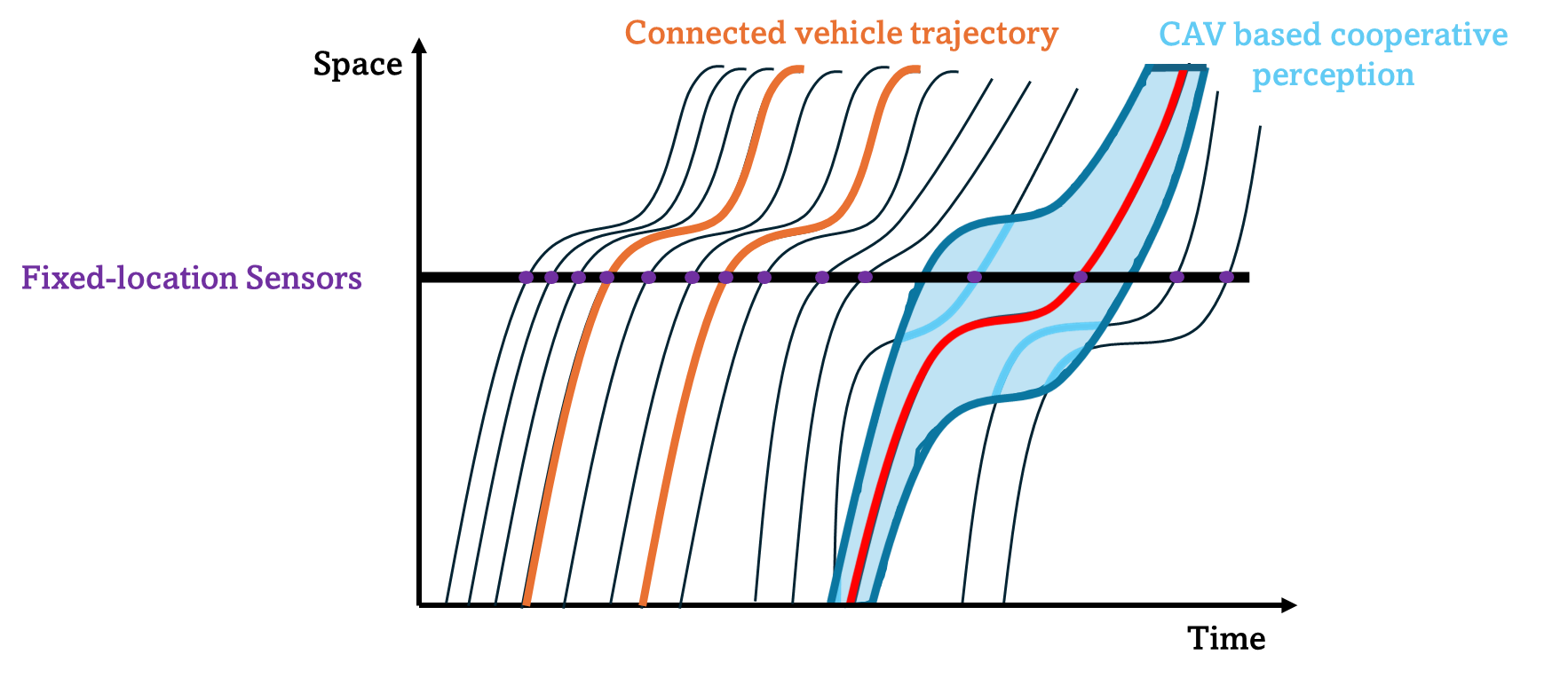}
    \caption{Data modalities in traffic monitoring}
    \label{fig:data modality of traffic monitoring}
\end{figure}

Traffic monitoring aims to estimate, predict, reconstruct traffic states (e.g., flow, density and speed) by leveraging data from single or different sources. Based on data modality, existing monitoring methodologies can be classified into three paradigms: fixed-location sensors, probe vehicles, and perception sensors on connected and automated vehicles (CAV). As illustrated in Figure \ref{fig:data modality of traffic monitoring}, fixed-location sensors typically provide temporally continuous data but remain spatially sparse due to their stationary nature. In contrast, probe vehicle trajectories offer spatiotemporal continuous data, but only for individual vehicles. Finally, perception sensors equipped on CAVs (e.g., robotaxis) capture multiple vehicle trajectories simultaneously within the detection range. Furthermore, multiple CAVs can create a cooperative perception environment. However, unlike probe vehicle trajectories, data quality and coverage patterns collected by perception sensors are highly dependent on vehicle location distribution across the network (e.g., overlapping). The following provides a brief review of prior studies across the three monitoring methodologies.

Despite the differences in sensing mechanics, fixed location sensors share a fundamental data modality: they provide high-frequency, temporally continuous observations at specific locations \citep{zhang2011data}. Consequently, the primary challenge has been inferring continuous spatial traffic states (e.g., link-level density or travel time) from these discrete, point-based measurements. Classic methodologies addressed this challenge through statistical inference and probabilistic modeling. Study by Dailey \citep{dailey1993cross} utilized cross-correlation techniques on flow sequences to estimate travel times, establishing the viability of measuring propagation delay from point sensors. Later on, \citep{coifman2002estimating} utilized kinematic wave theory and a simplified triangular flow-density relationship to estimate link travel times and vehicle trajectories using data from a single dual loop detector. By integrating local vehicle velocities with assumed signal propagation speeds, this study reconstructs trajectories without requiring vehicle re-identification or hardware at multiple locations.
With the raise of machine learning techniques, recent studies have explored learning-based methods. For example, to address the inherent noise and non-linear dynamics in sensor data, 
Shi et al. \citep{shi2021physics} applied Physics-Informed Deep Learning (PIDL) to embed traffic flow models (e.g., LWR conservation laws) into deep neural networks, enabling accurate traffic state and fundamental diagram estimation even observational data is sparse.
Meanwhile, vision-based monitoring systems have evolved to extract vehicle kinematic data from infrastructure cameras. \citep{coifman1998real} pioneered a real-time computer vision framework capable of extracting vehicle trajectories from roadside cameras, extending data modality from simple counts to detailed driving behaviors.  

Over the past decade, trajectory data from probe vehicles has been extensively applied to traffic monitoring and state estimation (TSE). \citep{seo2015estimation} developed a Lagrangian estimation method that directly derives flow and density from vehicle spacing data, eliminating the need for exogenous assumptions such as a fundamental diagram or a known MPR. \citep{zheng2017estimating} demonstrated that GPS trajectories from CVs could be used to estimate traffic volumes at signalized intersections by modeling vehicle arrivals as time-dependent Poisson processes, providing critical inputs for signal control. However, the effectiveness of this approach is often hindered by low MPRs. To mitigate this, various methods have been proposed to estimate macroscopic traffic states from partial observations \citep{han2021estimation, Yuan2024GAI} or to reconstruct full trajectories from limited samples \citep{sun2013vehicle, chen2022vehicle}. While effective, these methods treat probe vehicles as passive sensors, whose sensing results are a byproduct of exogenous travel demands rather than a coordinated effort.


The emergence of CAVs introduces a shift in sensing granularity. CAV fleets such as robotaxis could share their detection results with each other or with the infrastructure, which greatly improve the traffic monitoring efficiency. A much lower MPR is needed to collect equivalent data than those from probe vehicles \citep{Li_Han_Ma_2022, Xia_Meng_Han_Li_Tsukiji_Xu_Zheng_Ma_2023, Cai_Qu_Gao_Chen_2023}. \citep{cao2022analytical} showed that less than 10\% of CAVs can observe more than 50\% of total vehicles on average. \citep{chen2022cooperative} also obtained similar results that 5\% of CAVs plus an infrastructure Lidar sensor could reach nearly 50\% of equivalent probe vehicle MPR. The cooperative perception based data collection mechanism also brings two new challenges. First, the observed vehicles are clustered around the CAV rather than randomly distributed across the road segments as seen in data collected from probe vehicles, generates new data patterns and thus requiring new traffic state estimation and prediction methodologies. Second, if multiple CAVs are located close to each other, a great portion of data collected from each CAV is redundant. To improve traffic monitoring efficiency, CAVs need to be distributed sparsely within the network, each covering non-overlapping spatial areas and temporal intervals. Few studies have explored how to actively manage fleet mobility to achieve this objective. The growing deployment of robotaxis motivates this study, because unlike individually operated vehicles, robotaxis can be centrally coordinated by a fleet manager to implement the proposed routing for monitoring framework. 



\subsection{Vehicle Routing Models}
In a mixed traffic condition with CAVs and human driven vehicles (HDVs), existing studies aim to optimize the routes of CAVs to improve the traffic network performance \citep{zhang2018mitigating, guo2021mixed}. For example, \citep{zhang2018mitigating} proposed a routing scheme to control a small portion of CAVs and bring the system close to system optimum. The work proposed in \citep{guo2021mixed} has a similar objective but further considers instantaneous dynamic user equilibrium (IDUE) for HDV and dynamic system optimal (DSO) for CAVs. This work, along with others such as \citep{psaraftis1995dynamic, ouertani2023dynamic}, incorporate time-dependent link travel times into the routing framework. Similarly, in \citep{Li_Mirchandani_Zhou_2015}, the time-dependent travel time is represented by a space-phase-time (SPT) hypernetwork. Notably, this study not only optimizes the routes of vehicles, but also jointly optimizes traffic signals. More work on time-dependent routing problems can be found in \citep{gendreau2015time}. 

From the modeling perspective, \citep{munari2016generalized} reviewed two main forms of formulating vehicle routing formulations, link-based and path-based. The link-based formulation considers link travel time and uses decision variables (i.e., whether to traverse a specific link) to associate links in a network graph \citep{Li_Mirchandani_Zhou_2015}, while the path-based formulation defines the decision variable based on the paths and optimizes routes from a feasible path set. 

In terms of the objective functions, most studies focus on reducing total travel cost (e.g., distance, travel time) \citep{zhang2018mitigating, guo2021mixed, Li_Mirchandani_Zhou_2015} or network congestion level (e.g., total queue length \citep{de2020multi}). Some works start to include other objectives such as sustainability (e.g., energy consumption) and fairness (equity). For instance, in the context of electric vehicles, \citep{cai2023delivery} introduced a multi-objective vehicle routing problem (VRP) variant that addresses energy consumption under traffic congestion, time windows, and simultaneous pickup-and-delivery settings. This model jointly minimizes total cost and maximizes customer satisfaction, while incorporating a dynamic energy consumption model considering vehicle speed and load. In terms of equity, fairness has been included as an objective. \citep{kang2024promoting} proposed a two-sided fairness framework for dynamic VRP, introducing a fairness-aware genetic algorithm (2FairGA) that balances utilities between service providers and customers. Additionally, workload equity (i.e., equally distributed workload among all vehicles) in VRP has been studied. \citep{matl2018workload} analyzed various equity functions for balanced route workloads, highlighting how non-monotonic measures can produce inconsistent outcomes, while \citep{nekooghadirli2026workload} showed that multi-period workload equity can coexist with near-optimal routing costs over longer planning horizons. For the traffic monitoring perspective, utilizing robotaxis as drive-by sensors exploits the spatial flexibility of vehicle trajectories to capture traffic dynamics across wider geographic areas. For example, \citep{zhu2014mobile} reformulated traffic sensor deployment as an information-acquisition-oriented VRP, where sensing benefits are modeled as nonlinear functions of stay time and link importance. This work highlights the coupling between mobility decisions and estimation performance, but treats traffic conditions and sensing structure as exogenous inputs.

In summary, existing studies treat traffic demand, vehicle interactions, and service constraints as external to the sensing problem. In contrast, the emergency of robotaxi fleets fundamentally changes the problem structure. When vehicle movements can be directly coordinated, sensing and routing decisions become intrinsically coupled . This coupling gives rise to a new class of endogenous information acquisition problems, where mobility decisions simultaneously affect service performance and network observability. To the best of our knowledge, there are very few studies that consider traffic monitoring in the dynamic vehicle routing framework, especially under the robotaxi and cooperative perception settings. This study is among the earliest efforts to address this research gap.

\section{Methodology}
\label{sec:methodology}
\subsection{Framework Overview}

\begin{figure}[pos=tbp]
    \centering
    \begin{subcaptionbox}{Link-level cell typology\label{fig:ctm_link}}[0.45\linewidth]
        {\includegraphics[width=\linewidth]{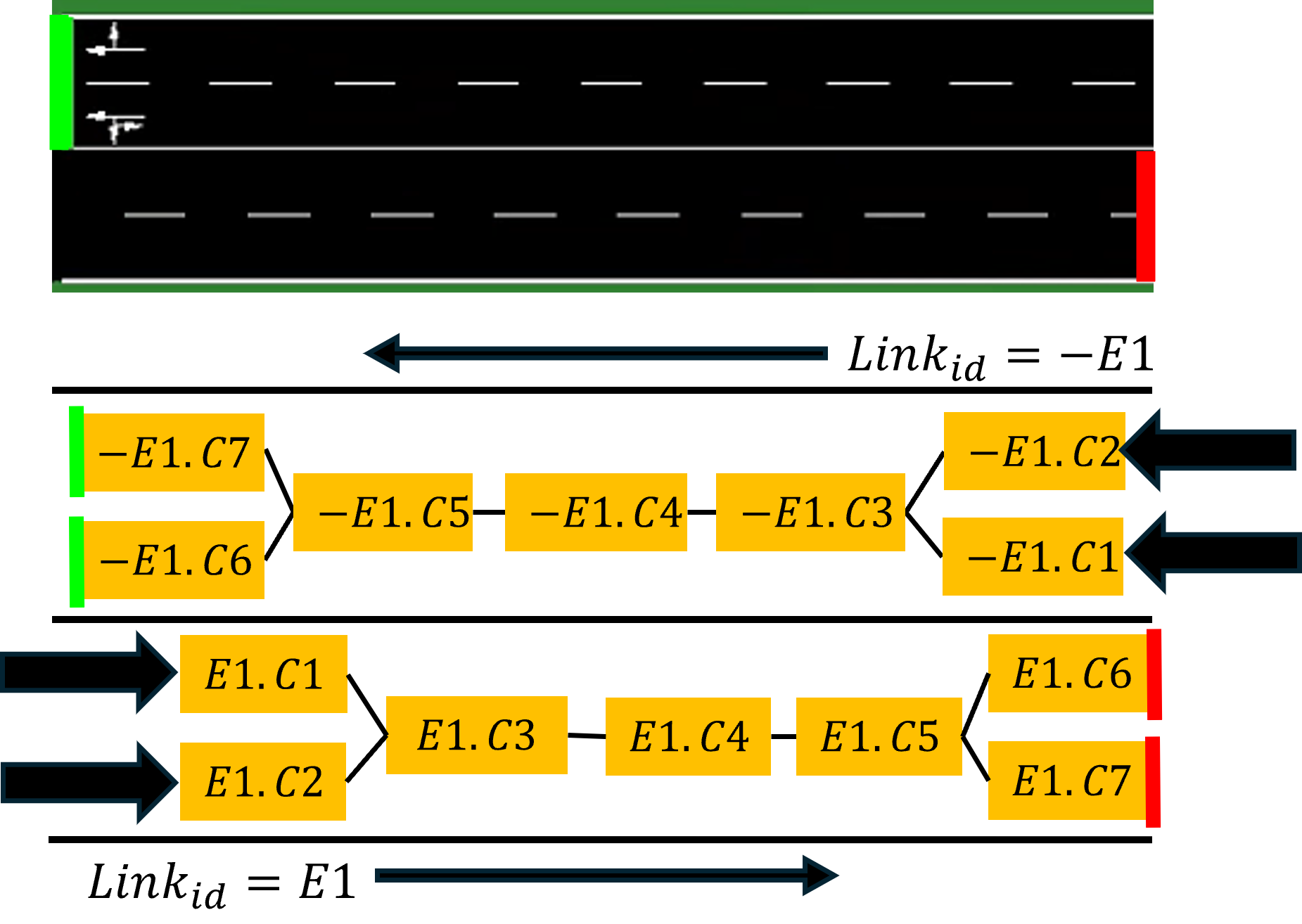}}
    \end{subcaptionbox}
    \hfill
    \begin{subcaptionbox}{Intersection-level cell typology\label{fig:ctm_inter}}[0.45\linewidth]
        {\includegraphics[width=\linewidth]{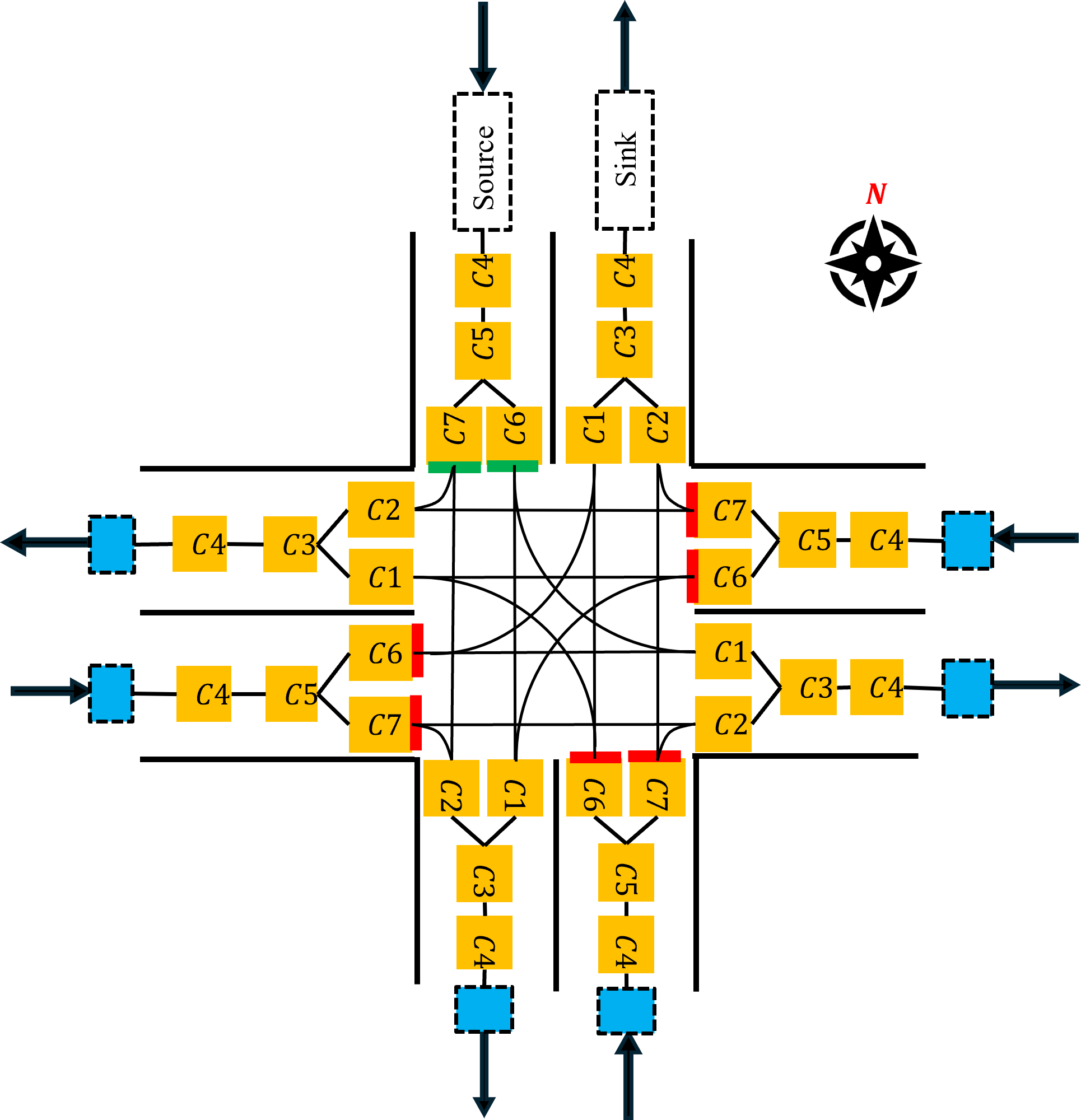}}
    \end{subcaptionbox}
    \caption{Network CTM representation}
    \label{fig:ctm_typology}
\end{figure}
\begin{figure}[pos=tbp]
    \centering
    \includegraphics[width=\linewidth]{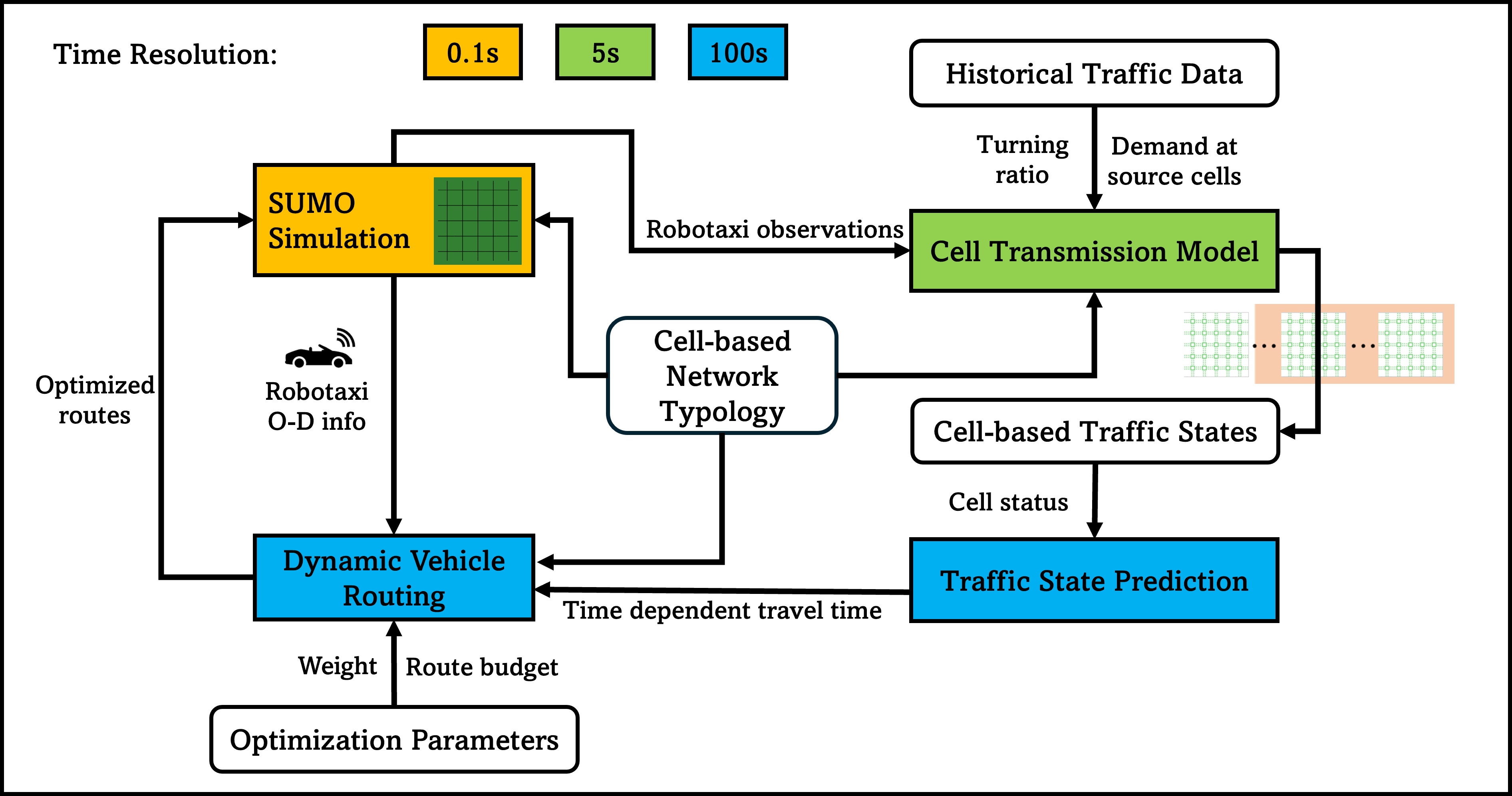}
    \caption{Overview of the routing for traffic monitoring framework}
    \label{fig:main_workflow}
\end{figure}

An overview of the proposed robotaxi routing for monitoring framework is illustrated in Figure~\ref{fig:main_workflow}. We use SUMO as the microscopic simulation platform in this study. Two types of vehicles are generated in SUMO, background vehicles (BV) and robotaxis. BVs serve as background traffic while robotaxis are able to perform traffic monitoring tasks. Robotaxis can observe surrounding vehicles within a predefined detection range. A cell-based network typology is designed and each link is divided into several cells as shown in Figure~\ref{fig:ctm_link}. This cell-based network typology is combined with the robotaxis' locations in SUMO to determine whether a cell is occupied. If a cell is occupied by one or more robotaxis, it is considered to be observed. An observed cell provides complete traffic data (e.g., average speed, and density/number of vehicles). The same cell-based network typology is also applied to construct the CTM for traffic state estimation. The cell length is set to be the same as the typical detection range of a robotaxi (e.g., 80 meters). The CTM runs in parallel with the SUMO simulation and estimates current traffic states based on historical traffic data (e.g., turning ratio at the intersection cells) and robotaxi observations. In addition, the CTM is also responsible for predicting future traffic states. It runs beyond the current time to generate cell-based traffic states for $T$ more time steps, where $T$ is the planning horizon in the Dynamic Vehicle Routing model. The predicted traffic states (i.e., density of each cell at each time step) are stored in a spatiotemporal matrix by cell and time indexes, from which the time-dependent travel time can be derived. Finally, when the Dynamic Vehicle Routing model is activated, it requires inputs including the network typology, current robotaxi locations and O-D information, time-dependent cell travel time, and predetermined optimization hyper-parameters. The Dynamic Vehicle Routing model is formulated as a mixed-integer linear programming (MILP) model to find optimal routes for all robotaxis in the network. The objective function aims to minimize the total robotaxi travel time while maximizing network coverage or equivalently maximizing the total number of observed cells during the planning horizon. A rolling horizon scheme is implemented to solve the MILP iteratively, allowing the model to accommodate changing traffic conditions and update the monitoring and travel time prediction results. The optimized robotaxi routing decisions are then applied to the SUMO simulation after each planning horizon for execution.

Major assumptions of this study are listed below.


\begin{enumerate}
    
    \item It is assumed that robotaxis' routing decisions do not affect the background traffic's routing decisions and corresponding time dependent travel time, considering low MPRs (e.g., less than $10\%$).
    \item The travel time within the intersection area is ignored when calculating the total travel time of a specific route, but the delay caused by queuing and traffic signals are considered. 
    \item A central controller has access to all robotaxis' real-time information (i.e., location at each time step) and is responsible for assigning routes to the robotaxis. All robotaxis follow the exact routes assigned by the central controller. 

 
\end{enumerate}

\subsection{Cell Transmission Model for Urban Network}
The CTM is a first order discrete approximation of the LWR model \citep{lighthill1955kinematic, richards1956shock}, proposed by \citep{daganzo1994ctm1}. The model assumes a triangle or trapezoidal fundamental diagram and represents a road segment with consecutive homogeneous sections. 
Later, Daganzo extended the model to represent network traffic, which allows it to model urban networks with signalized intersections \citep{daganzo1995ctm2}.

\begin{figure}[pos=tbp]
    \centering
    \includegraphics[width=0.5\linewidth]{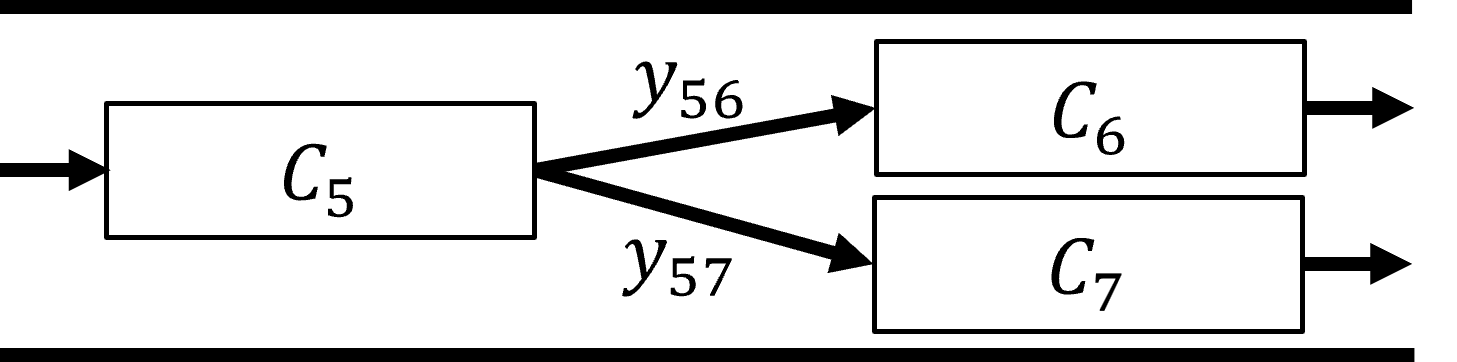}
    \caption{Diverging cell structure}
    \label{fig:ctm_diverge_structure}
\end{figure}

Six different types of cells are defined to present the urban network: ordinary cells, merging cells, diverging cells, source cells, sink cells, and intersection cells. Cell typologies for each intersection and each link are shown in Figure \ref{fig:ctm_typology}. We use two parallel cells at the entrance ($C1$ and $C2$) to receive flows coming from the upstream links. Then the two flows merge in the middle of the link ($C3$, $C4$, and $C5$). Then, $C5$ diverges to $C6$ and $C7$, mimicking the channelization at the intersection. Note that the capacity of $C3$, $C4$ and, $C5$ is twice that of other cells. 
The flow dynamics between different types of cells follow the definitions in \citep{daganzo1994ctm1, daganzo1995ctm2}, with one exception. In the original CTM, the diverging cells follow the First-In-First-Out (FIFO) principle. Whenever one of the downstream cells is congested, the diverge cell stops distributing flows even the other cell still has capacity. This setting is valid when the downstream cells represent different road segments or the lane change maneuver is forbidden. However, in our setting, through traffic can choose either left lane or right lane to proceed, which causes frequent lane changes in the SUMO simulation to balance the queue length. To better reflect the queuing dynamics in SUMO, we release the FIFO principle and allow flows from the diverging cell to enter downstream cells independently. Given the cell typology in Figure \ref{fig:ctm_diverge_structure}, the outgoing flows $y_{56}$ and $y_{57}$ of the diverging cell $C5$ is expressed in Equations \ref{eqa: cell5_div_logic1} and \ref{eqa:cell5_div_logic2}.

\begin{equation}
    y_{56} = \text{min} \{R_{6}, S_{5}\times p_{56}\}
    \label{eqa: cell5_div_logic1}
\end{equation}

\begin{equation}
    y_{57} = \text{min} \{R_{7}, S_{5}\times p_{57}\}
    \label{eqa:cell5_div_logic2}
\end{equation}

Where, $R_{6}$ and $R_{7}$ are the receiving functions of cells $C6$ and $C7$ respectively; $S_{5}$ is the sending function of cell $C5$; and $p_{56}$ and $p_{57}$ are the diverging ratios and $p_{56}+p_{57}=1$. The diverging ratios are determined in Equation \ref{eqa:cell5_turnratio}.

\begin{align}
    \begin{cases}
        p_{56} = \frac{n_{l}+n_{tr}\times \frac{R6}{R6+R7}}{n_{l} + n_{th} + n_{r}} \\
        p_{57} = \frac{n_{r}+n_{tr}\times \frac{R7}{R6+R7}}{n_{l} + n_{th} + n_{r}}
    \end{cases}
\label{eqa:cell5_turnratio}
\end{align}

Where, $n_l$, $n_{tr}$, and $n_r$ are the historical left turn, through and right turn flow for the given approach. Based on the lane approach configuration in Figure \ref{fig:ctm_link}, the left turn flow must enter cell 6 while the right turn flow must enter cell 7. The through flow will be distributed by the remaining capacities of the two downstream cells respectively. Thus the diverging ratios contain $R_{6}$ and $R_{7}$ and are dynamic over time.

\subsection{Dynamic Vehicle Routing Model Formulation}
The objective of the dynamic vehicle routing model aims to generate optimal routes for all robotaxis in the network considering the maximization of spatiotemporal network coverage and minimization of total robotaxi travel time. Main notations used in the formulation are shown in Table \ref{Table:notations}.

\begin{table}
\begin{flushleft}
\caption{Variables and notations}
\label{Table:notations}
\begin{tabular}{lp{14cm}}
\hline\hline
\textbf{Notation} & \textbf{Meaning} \\
\hline
$A$ & Set of robotaxis\\
$I$ & Set of cells in the network\\
$E$ & Set of edges in the network \\
$T$ & Planning horizon \\
$x_{i, t}^{a}$               & Binary variable, equals to 1 if robotaxi $a$ enters cell $i$ at time $t$, and 0 otherwise\\
$z_{i, j, t, s}^{a}$            & Binary variable, equals to 1 if robotaxi $a$ enters cell $i$ at time $t$ and leaves cell $i$ at time $s$, heading to cell $j$, and 0 otherwise\\
$y_{i}^{t}$ & Binary variable, equals to 1 if cell $i$ is occupied by any robotaxi(s) at time $t$, and 0 otherwise\\
$\omega_{i, t}^{a}$ & Binary variable, equals to 1 if robotaxi $a$ occupies cell $i$ at time $t$, and 0 otherwise\\

$c_{i}^{t}$ & Parameter, time-dependent travel time of cell $i$ for robotaxis entering at time $t$ \\ 
$\Pi_{i, j}$ & Binary parameter for network typology, equals to 1 if cell $i$ is connected to cell $j$, and 0 otherwise, $(i, j) \in E$ \\

$o(a)$ & Origin cell or the starting cell at the beginning of the planning horizon of robotaxi $a$ \\
$d(a)$ & Destination cell of robotaxi $a$ \\
$M$ & A sufficiently big number \\
\hline\hline
\end{tabular}
\end{flushleft}
\end{table}

We model the urban traffic network as a directed graph $G(I, E)$. Each cell in the CTM representation $i\in I$ is considered as a node and the connection between cells $i$ and $j$, $(i,j) \in E$ is considered as a directed edge. Time is discretized into steps, and the planning horizon $T$ represents total number of time steps. The time resolution in the dynamic routing model is set to be the same as the CTM and the planning horizon is predefined and should be long enough for all robotaxis in the network to reach their destinations.
Four primary variables are designed to formulate the problem. $x_{i,t}^{a}$ is used to represent selected cells in robotaxi $a$'s route and the entry time $t$ is used to locate the corresponding time-dependent travel time $c_i^t$. 
Variable $z_{i,j,t,s}^{a}$ further expands the dimension of $x_{i,t}^{a}$ to include the connecting cell(s) $j$ and the departure time $s$, which is mainly used to formulate the flow conservation, ensuring the selected cells consist of a feasible path from origin to destination.
Variable $\omega_{i, t}^{a}$ is designed for vehicle-level cell occupation and $y_{i}^{t}$ indicates the cell-level spatiotemporal occupation, which is used to calculate the coverage term in the objective function.
\subsubsection{Objective function}
The objective function contains two terms which represent total robotaxi travel time and spatiotemporal network coverage, with corresponding weights $\alpha_{1}$ and $\alpha_{2}$. The total robotaxi travel time is calculated as the summation of all cell traverse time of all robotaxis in the network. The network coverage is calculated by the summation of cell occupation $y_i^{t}$ over time, and normalized by total time steps and total number of cells. Note that since both $T$ and $I$ are constants, the normalization will not impact the solution. The overarching goal is to maximize network coverage and minimizing total travel time at the same time. The values of $\alpha_{1}$ and $\alpha_{2}$ should be properly selected to keep a balance between the two objectives. The objective function is shown in Equation~\ref{eqa:obj_function}.

\begin{equation}
        \text{min} \ \  \frac{\alpha_{1}}{|A|}\sum_{a\in A}\sum_{i\in I, t\in T}{c^{t}_{i} \times x_{i, t}^{a}} - \alpha_{2}\frac{\sum_{i \in I, t \in T}{y_{i}^{t}}}{|I| \times |T|}
        \label{eqa:obj_function}
\end{equation}

\subsubsection{Variable \texorpdfstring{$x$}{x} - \texorpdfstring{$z$}{z} connection constraints}
As variable  $x_{i,t}^{a}$ only determines the entry time of cell $i$, $z_{i, j, t, s}^{a}$ is introduced to represent the departure time $s$ and the downstream cell $j$ of cell $i$. Equation \ref{eqa:x-z condition} depicts that when a robotaxi enters cell $i$, it can only have one combination of departure time and downstream cell. Equation \ref{eqa:cell connection} considers the network typology $\Pi_{i, j}$. Only a connected downstream cell of cell $i$ can be chosen. 
\begin{equation}
    x_{i, t}^{a} = \sum_{j}\sum_{s}{z_{i, j, t, s}^{a}} \quad \forall a \in A\quad i\in I \quad t\in T
\label{eqa:x-z condition}
\end{equation}


\begin{equation}
    z_{i,j,t,s}^{a} \leq \Pi_{i, j} * x_{i,t}^{a} \quad \forall a \in A \quad i \in I \quad t \in T \\
\label{eqa:cell connection}
\end{equation}

\subsubsection{Network flow constraints}

The flow conservation constraint in Equation \ref{eqa:flow_conservation2} guarantees that selected cells with associated entry and departure times consist of a time-continuous feasible path. 
Note that this constraint also applies to intersection cells because the travel time within the intersection is not considered, and waiting time at traffic signals is also included in the cell travel time $c_i^t$. This means all cells are connected without any transition time (i.e., no edge cost). Equations \ref{eqa:flow_conservation1} and \ref{eqa:flow_conservation3} ensure the robotaxi starts at the origin cell and reaches the destination cell.
Furthermore, Equation \ref{eqa:temporal uniqueness} constraints that one robotaxi cannot enter multiple cells at the same time and Equation \ref{eqa:single visit} limits each cell can be visited by each robotaxi only one time (no subtour). 


\begin{equation}
    \sum_{i\in I, \ t \in T}z_{i, j, t, s}^{a} - \sum_{k \in I,\ r \in T}z_{j, k, s, r}^{a} =  0  \quad \quad \text{if} \ j \neq o(a) \ \& \ j \neq d(a)  \quad \quad \quad \quad \forall a\in A \quad t,s,r\in T
    \label{eqa:flow_conservation2}
\end{equation}

\begin{equation}
    \sum_{j\in I,\ s\in T}z_{i, j, t, s}^{a} = 1  \quad \quad \text{if} \ i=o(a) \ \& \ t=0 \quad \quad \quad \quad \forall a\in A \quad t,s\in T
    \label{eqa:flow_conservation1}
\end{equation}

\begin{equation}
    \sum_{i\in I,\ t,s \in T}z_{i, j, t, s}^{a} =1 \quad \quad \text{if} \ j = d(a)  \quad \quad \quad \quad \forall a\in A \quad t,s\in T
    \label{eqa:flow_conservation3}
\end{equation}




\begin{equation}
    \sum_{i \in I}{x_{i, t}^{a}} \leq 1 \quad \forall a \in A, \ t \in T
\label{eqa:temporal uniqueness}
\end{equation}

\begin{equation}
    \sum_{t \in T}x_{i, t}^{a} \leq 1 \quad \forall a \in A, \ i \in I
\label{eqa:single visit}
\end{equation}

\subsubsection{Vehicle-level network occupation constraints}

\begin{figure}[pos=tbp]
    \centering
    \includegraphics[width=0.7\linewidth]{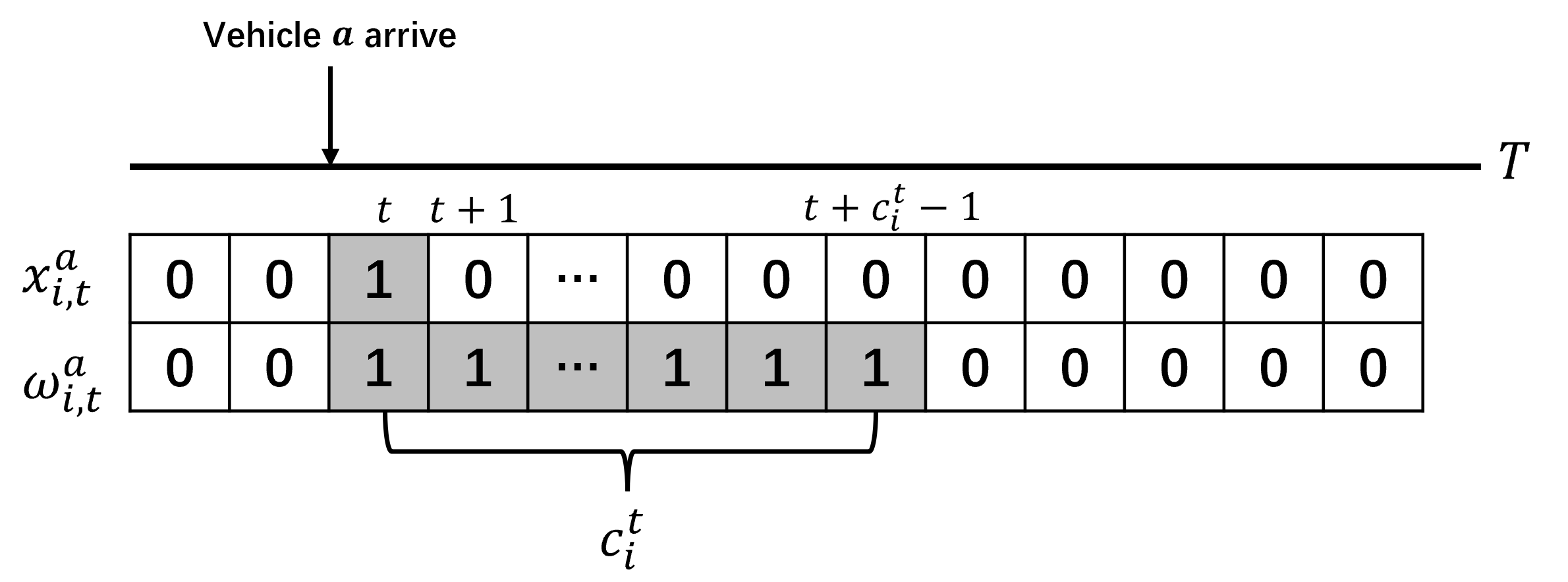}
    \caption{Relationship between $\omega$ and $x$}
    \label{fig:omg-x relation}
\end{figure}

We use variable $\omega$ to model the vehicle-level occupation of cells. The value of $\omega_{i,t}^{a}$ is determined by both $x_{i, t}^{a}$ and time-dependent travel time $c_{i}^{t}$. The relation between $\omega$ and $x$ is shown in Figure \ref{fig:omg-x relation}. Equation \ref{eqa:omg-x cond0} illustrates if robotaxi $a$ never visits cell $i$, then the occupation should be set to 0. If robotaxi $a$ visits cell $i$ at a certain time point $t$ (i.e., $x_{i, t}^{a}=1$), then this constraint is released and Equations \ref{eqa:omg-x cond1} to \ref{eqa:omg-x cond3} become effective. These three constraints collectively regulate the values of $\omega$ should be one from time $t$ to time $t+c_i^t-1$, and zero otherwise. 
These constraints are further linearized using the big-M method as shown in Equations \ref{eqa:omg-x-1:linear1} to \ref{eqa:omg-x-3:linear2}.

\begin{equation}
    \omega_{i, t}^{a} \leq \sum_{t \in T}{x_{i, t}^{a}} \quad \forall a \in A, i \in I, t \in T
    \label{eqa:omg-x cond0}
\end{equation}


\begin{equation}
    \sum_{0\leq k \leq t-1}{\omega_{i, k}^{a}}=0 \quad \forall i \in I, a \in A, t \in T
\label{eqa:omg-x cond1}
\end{equation}

\begin{equation}
    \sum_{t\leq k \leq t+c_{i}^{t}}{\omega_{i, k}^{a}}=c_{i}^{t} \quad \forall i \in I, a \in A, t \in T
\label{eqa:omg-x cond2}
\end{equation}

\begin{equation}
    \sum_{t+c_{i}^{t}+1\leq k < T}{\omega_{i, k}^{a}}=0 \quad \forall i \in I, a \in A, t \in T
\label{eqa:omg-x cond3}
\end{equation}

 

\begin{equation}
    \sum_{0\leq k \leq t-1}{\omega_{i, t}^{a}} \geq 0 
    \label{eqa:omg-x-1:linear1}
\end{equation}

\begin{equation}
    \sum_{0\leq k \leq t-1}{\omega_{i, t}^{a}} \leq  M\times (1-x_{i, t}^{a})
    \label{eqa:omg-x-1:linear2}
\end{equation}


\begin{equation}
    \sum_{t\leq k \leq t+c_{i}^{t}}{\omega_{i, t}^{a}} \geq x_{i,t}^{a}\times c_{i, t} 
    \label{eqa:omg-x-2:linear1}
\end{equation}

\begin{equation}
    \sum_{t\leq k \leq t+c_{i}^{t}}{\omega_{i, t}^{a}} \leq x_{i,t}^{a}\times c_{i, t} + M\times (1-x_{i, t}^{a})
    \label{eqa:omg-x-2:linear2}
\end{equation}


\begin{equation}
    \sum_{t+c_{i}^{t}+1\leq k < T}{\omega_{i, t}^{a}} \geq 0 
    \label{eqa:omg-x-3:linear1}
\end{equation}

\begin{equation}
    \sum_{t+c_{i}^{t}+1\leq k < T}{\omega_{i, t}^{a}} \leq  M\times (1-x_{i, t}^{a})
    \label{eqa:omg-x-3:linear2}
\end{equation}

\subsubsection{Cell-Level network occupation constraints (\texorpdfstring{$\omega$}{\omega} - \texorpdfstring{$y$}{y} relation)}

Converting vehicle-level occupation to cell-level occupation is necessary to handle the situation that multiple robotaxis occupy a cell at the same time. We use variable $y_{i}^{t}$ to indicate the occupation of cell $i$ at time $t$. Equation \ref{eqa:omg-y cond} states the relation and can be linearized to Equations \ref{eqa:omg-y:linear1} and \ref{eqa:omg-y:linear2} by the big-M method.
Note that in this study, we assume a perfect detection model so that all vehicles in the cell, if occupied by one or more robotaxis, are observed. This can be further extended to more complicated and accurate detection models that consider detection accuracy and uncertainty. 

    \begin{align}
        y_{i}^{t} = 
        \begin{cases}
            1 \quad \text{if} \ \sum_{a}{\omega_{i, t}^{a}}  \geq 1 \\
            0 \quad \text{if} \ \sum_{a}{\omega_{i, t}^{a}} =0
        \end{cases}
    \label{eqa:omg-y cond}
\end{align}

    \begin{equation}
        y_{i}^{t} \leq \sum_{a}{w_{i, t}^{a}}
        \label{eqa:omg-y:linear1}
    \end{equation}
        
    \begin{equation}
        M \times y_{i}^{t} \geq \sum_{a}{w_{i, t}^{a}}
        \label{eqa:omg-y:linear2}
    \end{equation}

In summary, the robotaxi dynamic routing model is formulated as an MILP problem (\textbf{P1}) as below:

    \textbf{Objective Function}: Equation \ref{eqa:obj_function}
    
    \textbf{Subject to}: Equations \ref{eqa:x-z condition} - \ref{eqa:single visit}, \ref{eqa:omg-x-1:linear1} - \ref{eqa:omg-x-3:linear2}, and \ref{eqa:omg-y:linear1}-\ref{eqa:omg-y:linear2}

One critical parameter that serves as the input to the model is the time-dependent travel time $c_{i}^{t}$, which indicates the total time to traverse cell $i$ if a robotaxi enters the cell at time $t$. It is calculated based on the predicted traffic states from the CTM. Assume $k_{i}^{t}$ denotes the predicted density of cell $i$ at time $t$. The speed of the cell $v_{i}^{t}$ can be calculated using the trapezoidal fundamental diagram. 

\begin{align}
    v_{i}^{t}=  
    \begin{cases}
        v_f \quad \text{if} \ 0 \leq k_{i}^{t} \leq k_1  \\
        q_{max}/k_{i}^{t} \quad \text{if} \ k_1 < k_{i}^{t} \leq k_2 \\
        |w|(k_j/k_{i}^{t}-1) \quad \text{if} \ k_2 < k_{i}^{t} \leq k_j
    \end{cases}
    \label{eqa: cell_speed}
\end{align}

Then the travel distance $d_{i}^{t}$ at time $t$ can be calculated as

\begin{equation}
    d_{i}^{t}= v_{i}^{t} \times \Delta{t}
\label{eqa:dis}
\end{equation}

Where $w$ is the wave speed; $v_f$ is the free flow speed; $k_1$ and $k_2$ are the lower and upper critical density; and $k_j$ is the jam density. Finally, the cell traverse time $c_{i}^{t}$ can be calculated as the total time steps elapsed when the summation of $d_{i}^{t}$ from $t$ is greater than or equal to the cell length $l$.



\subsubsection{Complexity Analysis}
The size of the proposed formulation and the dimensions of the problem are analyzed.
 In the MILP formulation, the sizes of the binary variables are shown below:
\begin{itemize}
    \item $x$–variables $x_{i,t}^a$: $|A||I||T|$,
    \item $z$–variables $z_{i,j,t,s}^a$: $|A||I|^{2}|T|^{2}$,
    \item $\omega$–variables $\omega_{i,t}^a$: $|A||I||T|$,
    \item $y$–variables $y_i^t$: $|I||T|$.
\end{itemize}

The primary source of the model complexity comes from $z_{i, j, t, s}^{a}$, which is a five dimensional matrix and its size increases quadratically with the network size and planning horizon. Notice that in equation ~\ref{eqa:cell connection}, $z_{i, j, t, s}^{a}$ could be none-zero only when $\Pi_{i, j} =1$. Given the network typology and limited cell connections, $\Pi_{i, j}$ is a very sparse matrix, and so is $z_{i, j, t, s}^{a}$. To reduce the model complexity and improve solving efficiency, we applied Equation \eqref{eqa: z-preset} to eliminate infeasible time-dependent arcs in advance. Denote $|E| = \Theta(|I|)$ to be the number of directed cell–to–cell connections, then the dimension of $z$–variables can be express as $\Theta(|A||E||T|)$, indicating linear scalability with respect to the number of robotaxi, cell connections, and planning horizon.

\begin{align}
    z_{i,j,t,s}^{a} 
    \begin{cases}
        \leq 1 \quad \text{if} \ s=t+c_{i}^{t} \quad \& \quad \Pi_{i,j}=1\\
        =0 \quad \text{otherwise}
    \end{cases}
    \label{eqa: z-preset}
\end{align}

Given there is no cell serves as a merging and diverging cell at the same time based on the CTM cell structure \citep{daganzo1994ctm1}, $|E|\leq3|I|$. Therefore, the total number of binary variables can be expressed as

\[
|A||E||T| \;+\; 2|A||I||T| \;+\; |I||T| = \Theta(|A||I||T|)
\]

The number of constraints exhibits a similar dimension. For the $x$–$z$ connection constraints, Equation \eqref{eqa:x-z condition} contribute $|A||E||T|$ equalities, while the topology constraints in Equation \eqref{eqa:cell connection} contribute at most $|A||E||T|$ inequalities. The flow conservation constraints in Equations \eqref{eqa:flow_conservation1}–\eqref{eqa:flow_conservation3} add $\Theta(|A||I||T|)$ additional constraints, and the temporal uniqueness and single-visit conditions in Equations ~\eqref{eqa:temporal uniqueness}–\eqref{eqa:single visit} contribute $|A||T| + |A||I|$ constraints. The mapping between $x$ and $\omega$ introduces additional $|A||I||T|$ inequalities in Equation \eqref{eqa:omg-x cond0} and the big-M linearization from Equations ~\eqref{eqa:omg-x-1:linear1}-\eqref{eqa:omg-x-3:linear2} adds another $6|A||I||T|$ inequalities. 

Finally, the cell-level occupation constraints in \eqref{eqa:omg-y:linear1}–\eqref{eqa:omg-y:linear2} add $2|I||T|$ inequalities. Combining all together, the total number of constraints can be expressed as
\[
\Theta\big(|A||E||T| + |A||I||T|\big),
\]

which again reduces to $\Theta(|A||I||T|)$ for sparse networks with $|E|=\Theta(|I|)$.

\section{Case Study}
\label{sec:case study}
In this section, a toy network is first used to demonstrate different routing strategies under the proposed routing framework. Then a $5\times 5$ urban network is constructed for comprehensive evaluation.

\subsection{Toy Network Analysis}

\begin{figure}[pos=tbp]
    \centering

    
    \begin{subfigure}[b]{0.18\textwidth}
        \includegraphics[width=\linewidth]{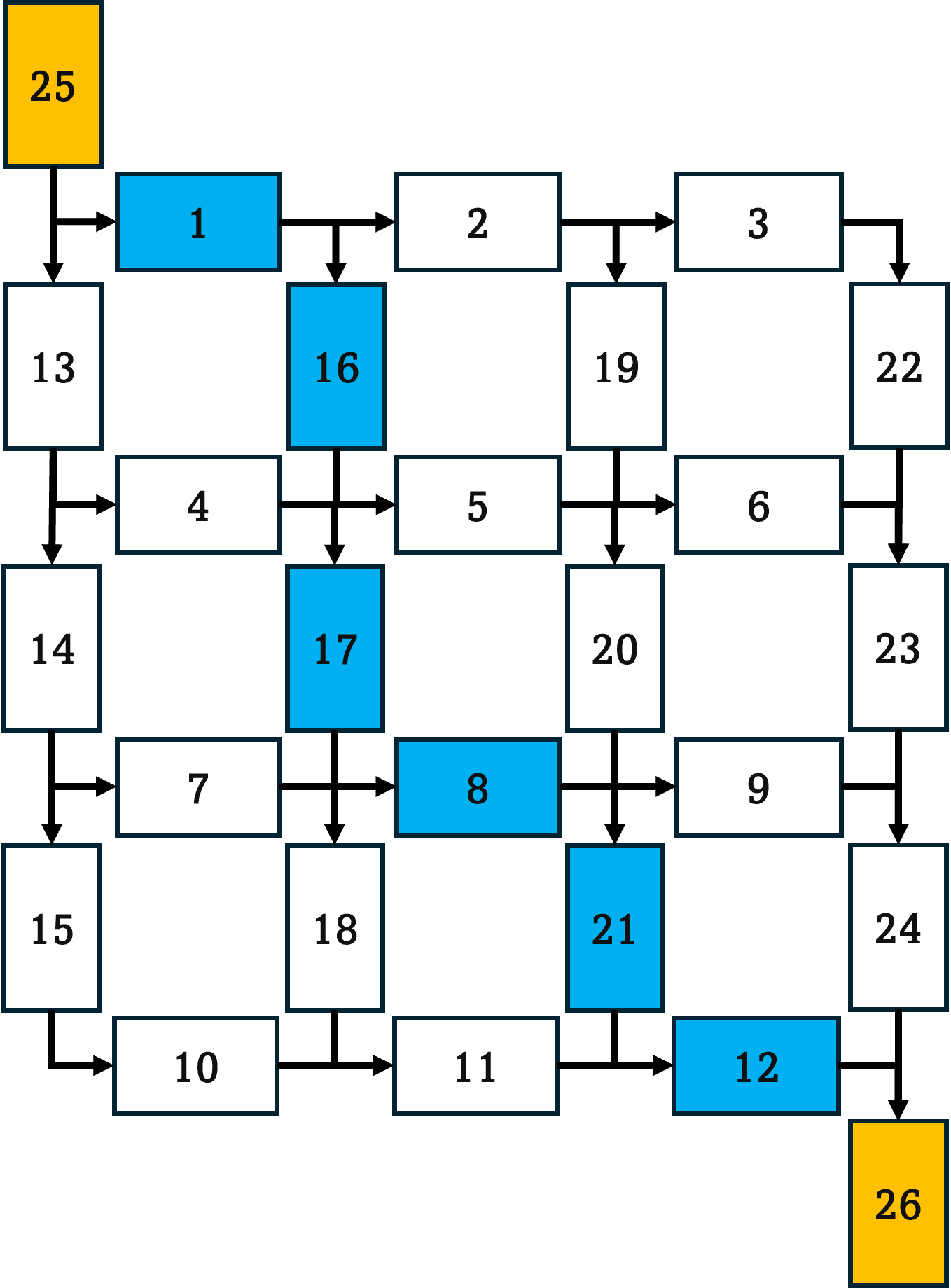}
        \caption{Vehicle 1}
        \label{fig:sub_toy_route1}
    \end{subfigure}
    \begin{subfigure}[b]{0.18\textwidth}
        \includegraphics[width=\linewidth]{figure/toy_net_result/toy_network_routing1-4.png}
        \caption{Vehicle 2}
        \label{fig:sub_toy_route2}
    \end{subfigure}
    \begin{subfigure}[b]{0.18\textwidth}
        \includegraphics[width=\linewidth]{figure/toy_net_result/toy_network_routing1-4.png}
        \caption{Vehicle 3}
        \label{fig:sub_toy_route3}
    \end{subfigure}
    \begin{subfigure}[b]{0.18\textwidth}
        \includegraphics[width=\linewidth]{figure/toy_net_result/toy_network_routing1-4.png}
        \caption{Vehicle 4}
        \label{fig:sub_toy_route4}
    \end{subfigure}
    \begin{subfigure}[b]{0.25\textwidth}
        \includegraphics[width=\linewidth]{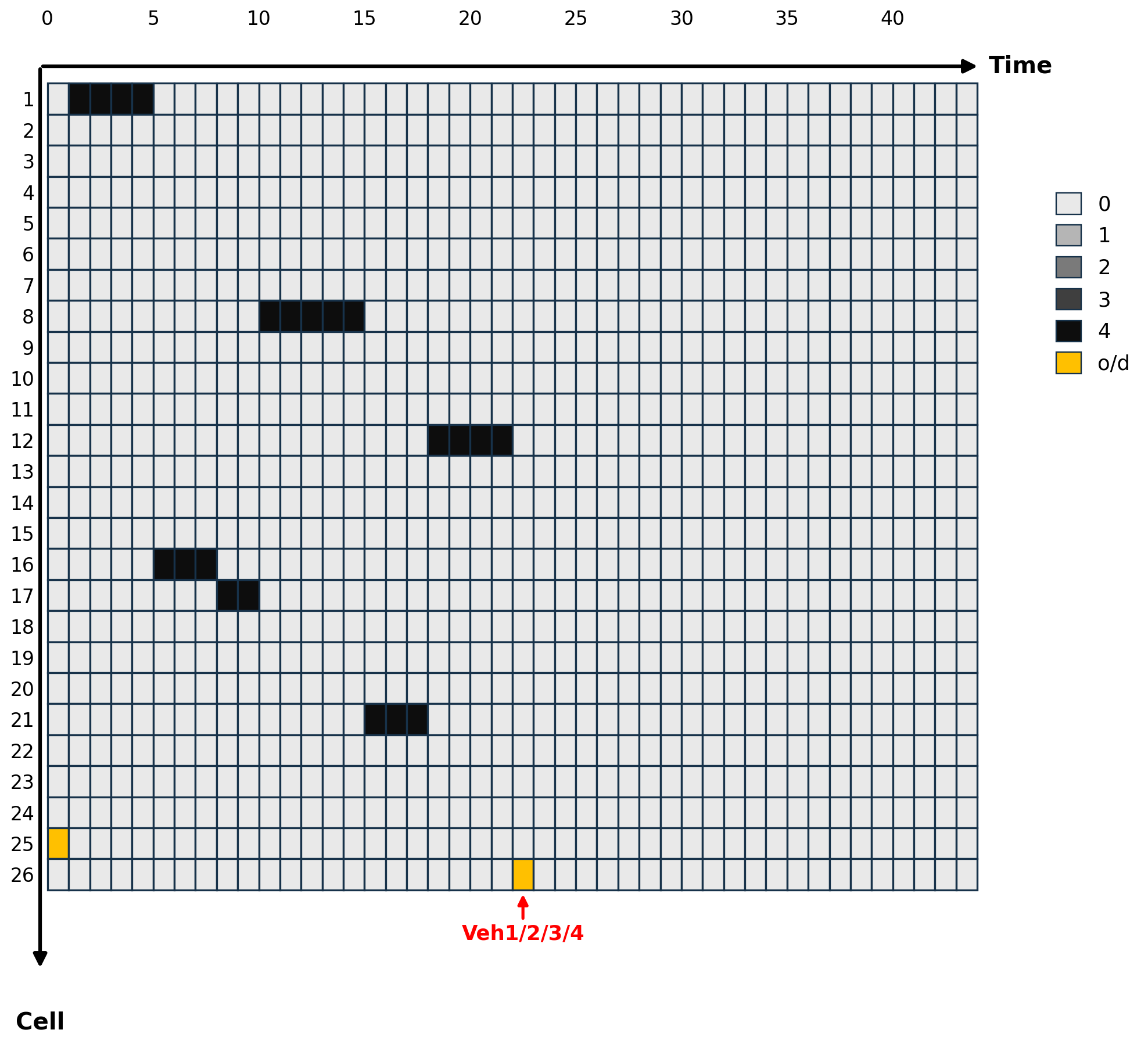}
        \caption{Spatiotemporal coverage}
        \label{fig:sub_toy_routeocc}
    \end{subfigure}

    \vspace{1em}


    \begin{subfigure}[b]{0.18\textwidth}
        \includegraphics[width=\linewidth]{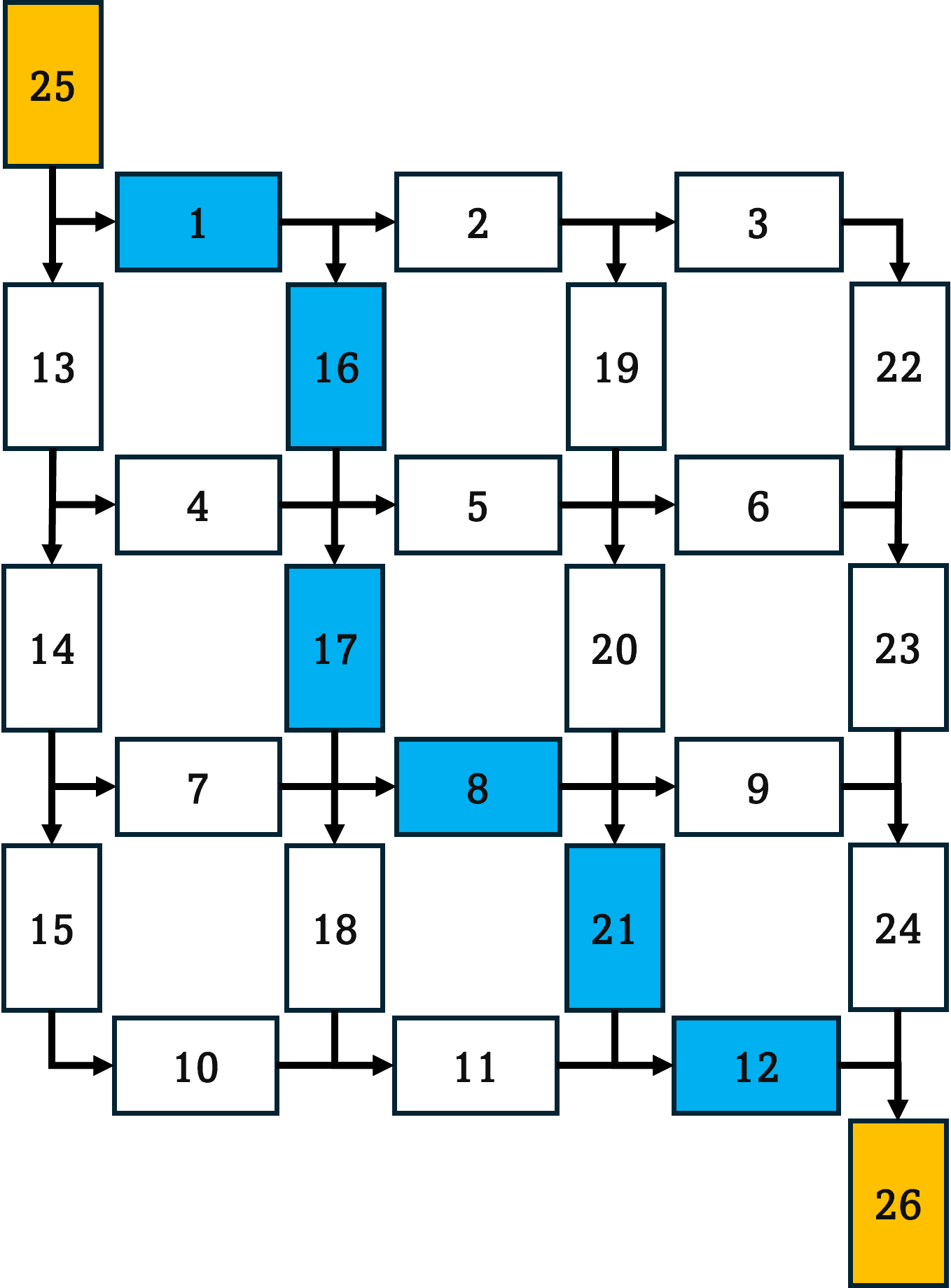}
        \caption{Vehicle 1}
        \label{fig:sub_toy_cover1}
    \end{subfigure}
    \begin{subfigure}[b]{0.18\textwidth}
        \includegraphics[width=\linewidth]{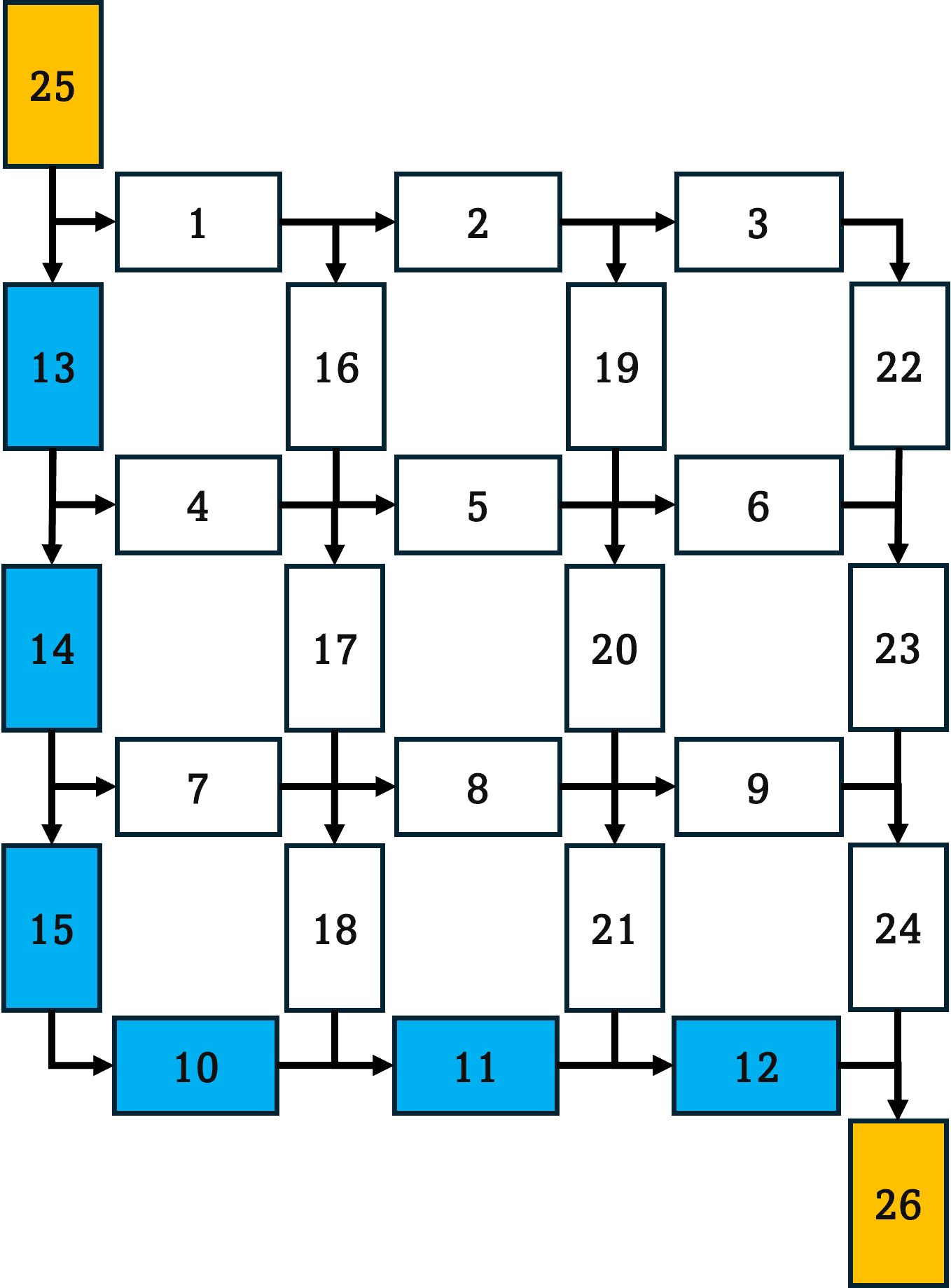}
        \caption{Vehicle 2}
        \label{fig:sub_toy_cover2}
    \end{subfigure}
    \begin{subfigure}[b]{0.18\textwidth}
        \includegraphics[width=\linewidth]{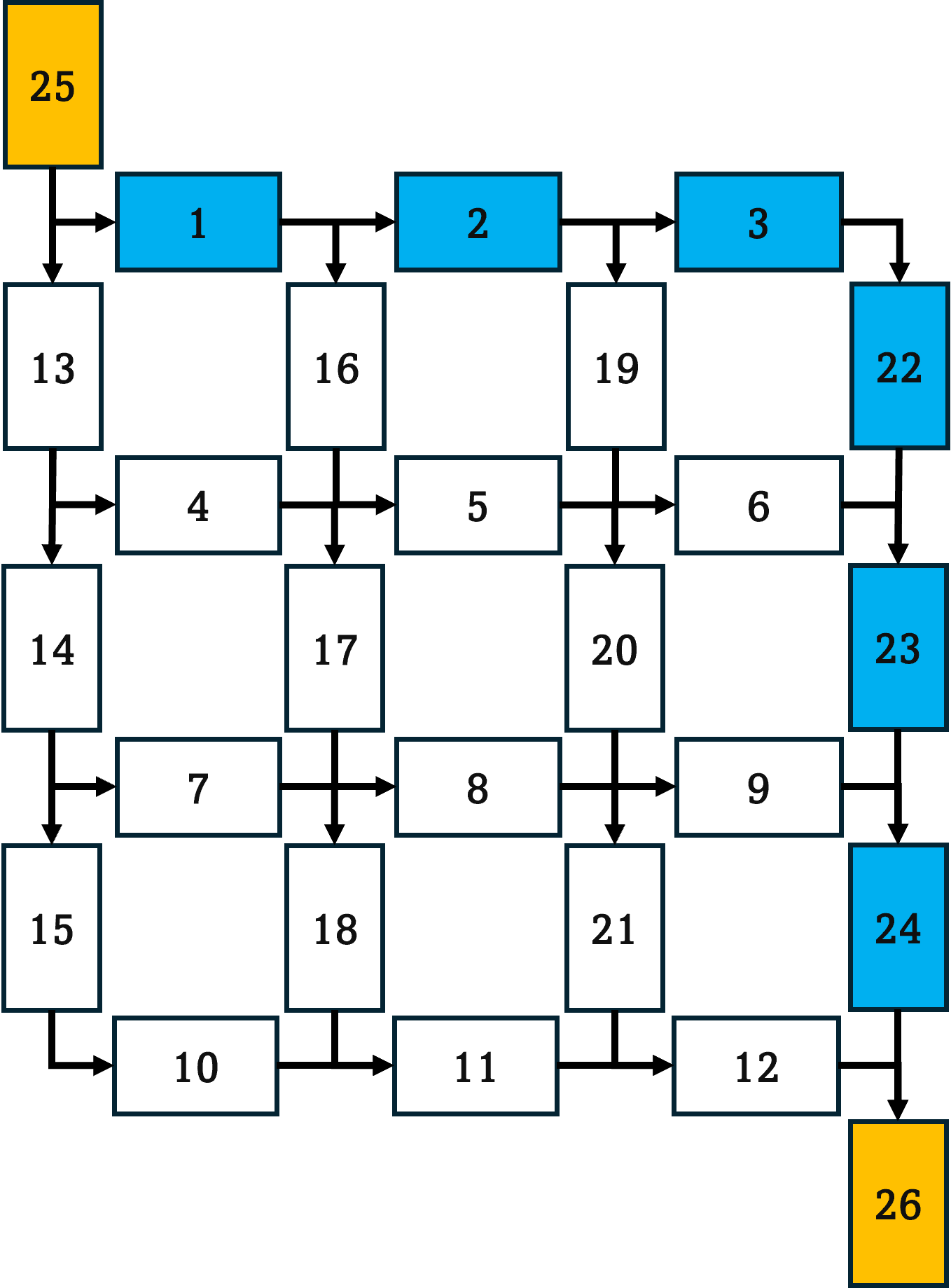}
        \caption{Vehicle 3}
        \label{fig:sub_toy_cover3}
    \end{subfigure}
    \begin{subfigure}[b]{0.18\textwidth}
        \includegraphics[width=\linewidth]{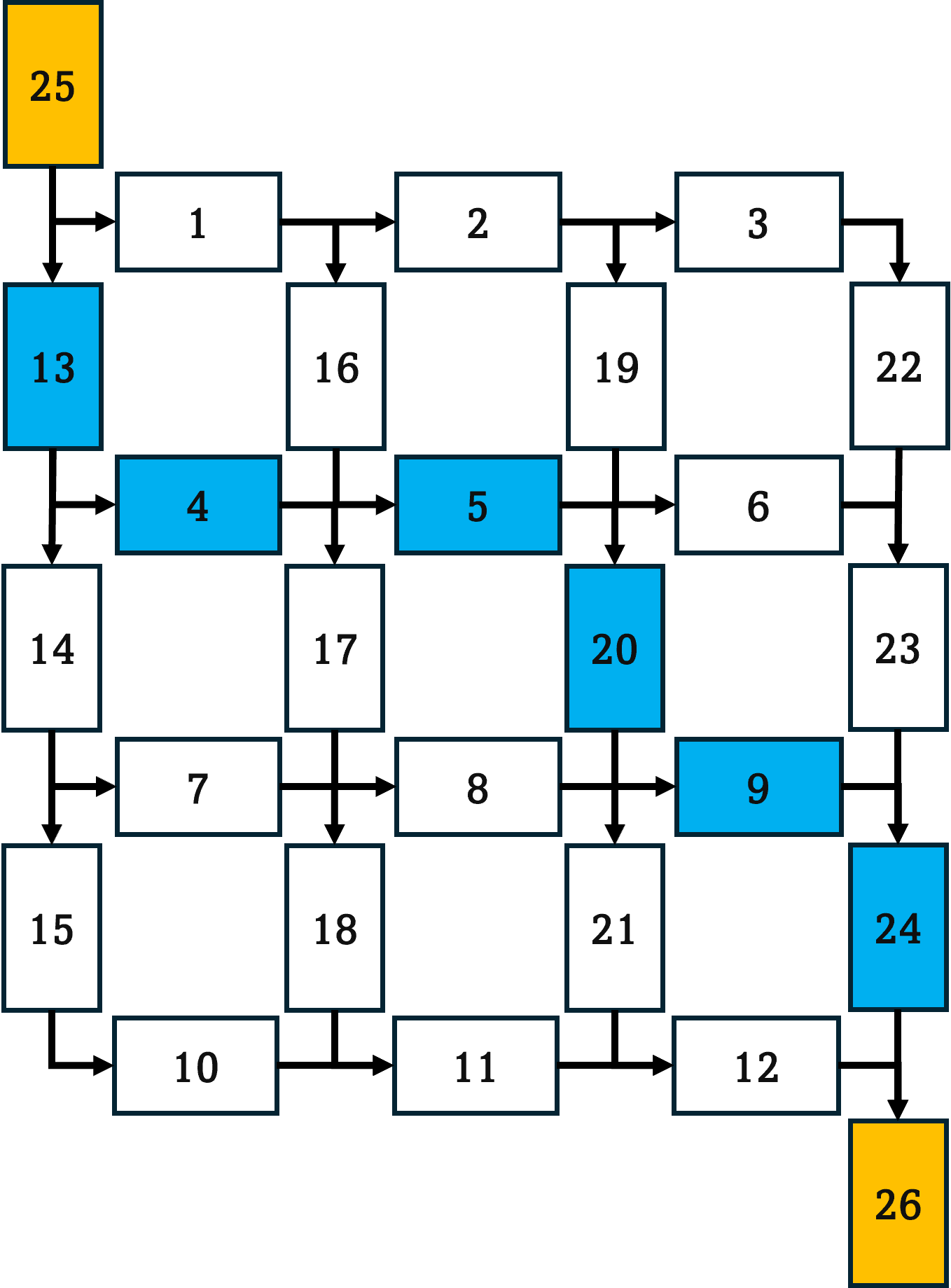}
        \caption{Vehicle 4}
        \label{fig:sub_toy_cover4}
    \end{subfigure}
    \begin{subfigure}[b]{0.25\textwidth}
        \includegraphics[width=\linewidth]{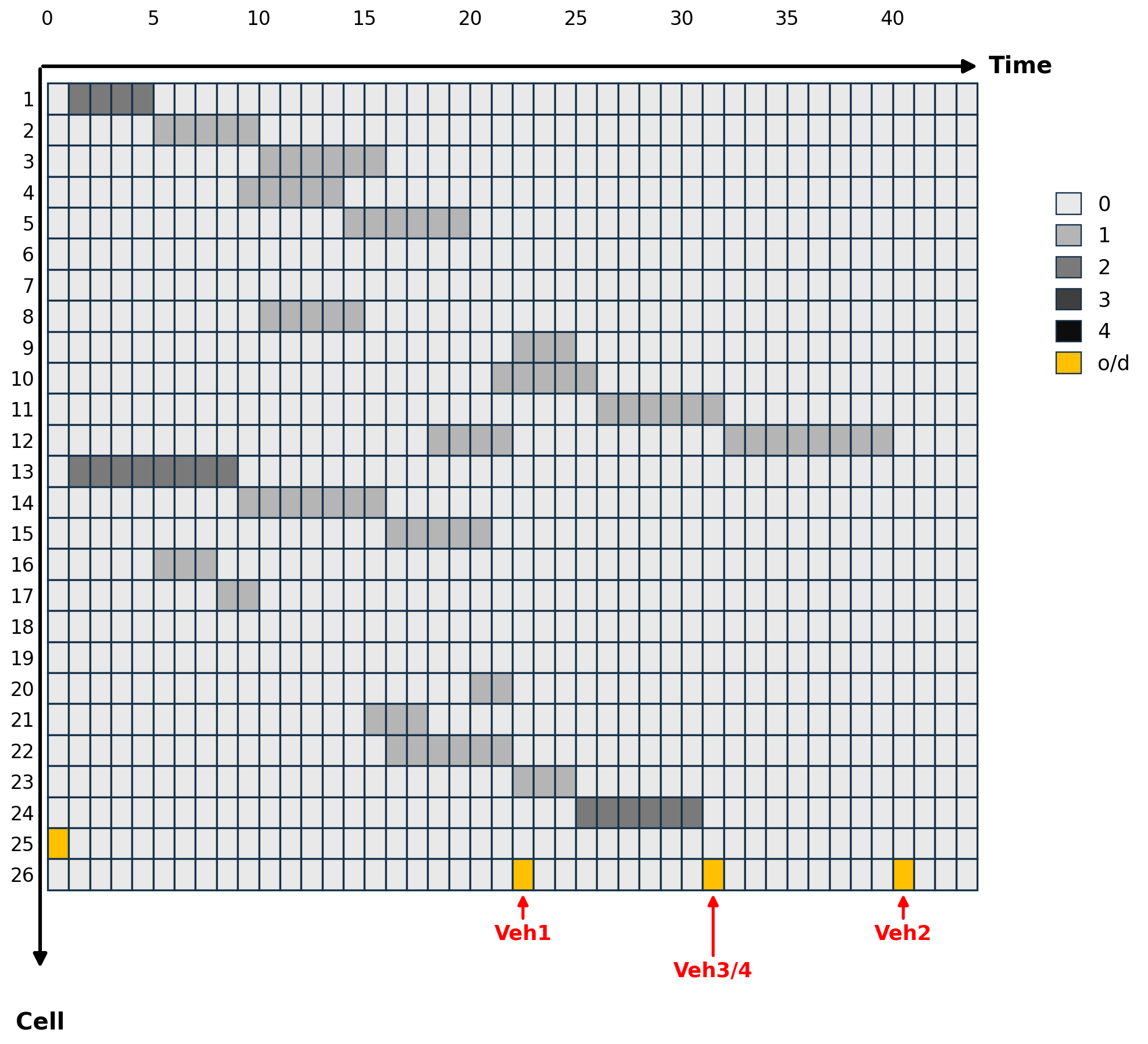}
        \caption{Spatiotemporal coverage}
        \label{fig:sub_toy_coverocc}
    \end{subfigure}

    \caption{Route choice and coverage under different weight parameters}
    \label{fig:toy_net_route_occ}
\end{figure}

\begin{figure}[pos=tbp]
    \centering
    \includegraphics[width=0.3\linewidth]{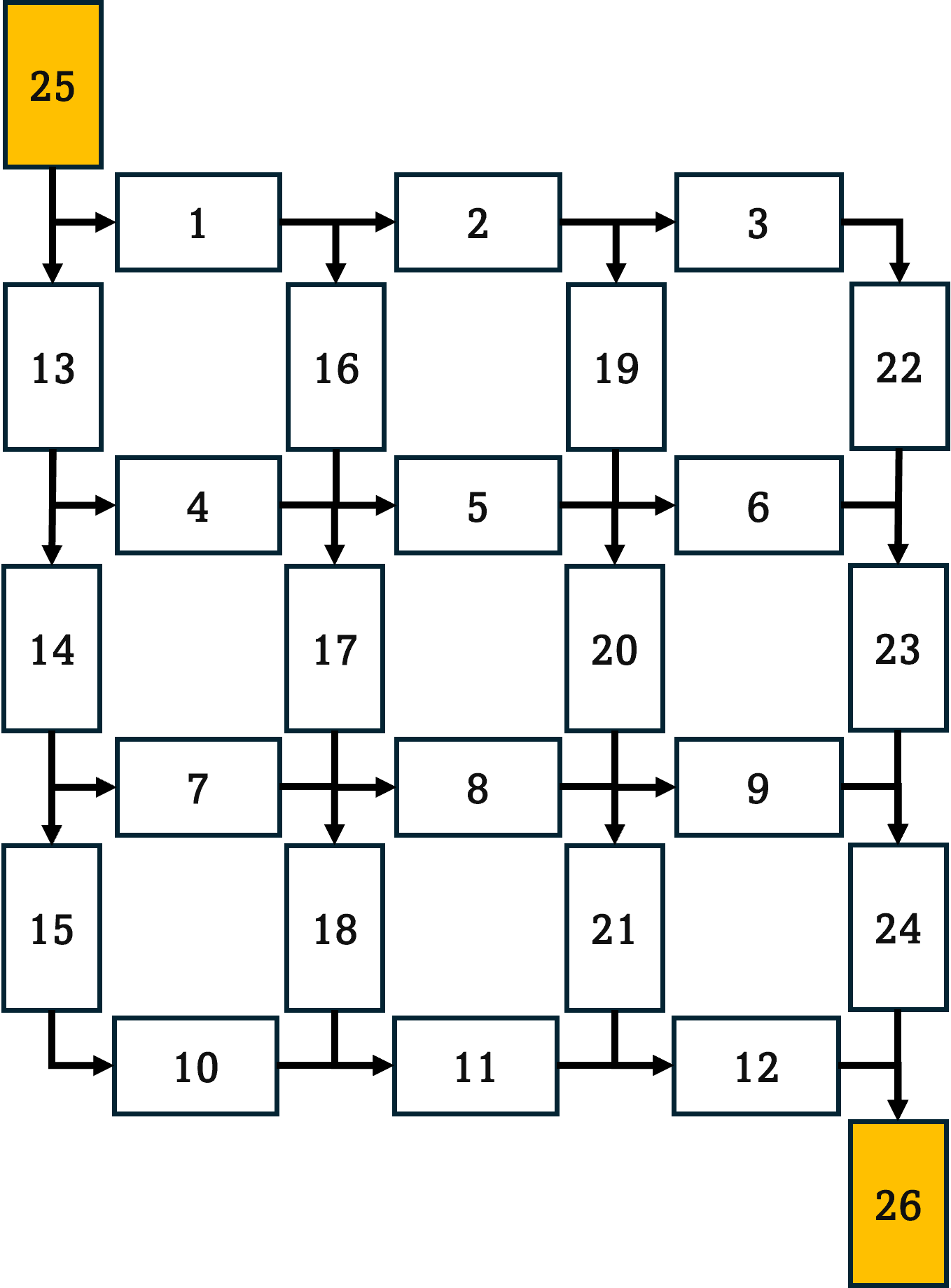}
    \caption{A toy network}
    \label{fig:toy_network}
\end{figure}

A toy network to test the dynamic vehicle routing formulation is shown in Figure \ref{fig:toy_network}. The toy network consists of 26 cells including one origin (cell 25) and one destination (cell 26). 
The time dependent travel cost of each cell is determined by Equation \eqref{eqa:toy_net-travel_cost}, where $i$ is the cell index and $t$ is the time step. Four vehicles start their trips at the same time (time 0) from cell 25. Two scenarios are examined. The first considers only the travel time cost ($\alpha_{1}:\alpha_{2} = 1:0$), while the second prioritizes the network coverage by assigning a substantially higher weight to $\alpha_2$ as ($\alpha_{1}:\alpha_{2} = 1:10^{6}$). The route choices in these two scenarios show different patterns, as shown in Figure \ref{fig:toy_net_route_occ}. The first row illustrates the routing decisions of the four vehicles and the corresponding cell-level spatiotemporal network coverage when only travel cost is considered, whereas the second row shows the same information when network coverage is emphasized. The densities in the spatiotemporal graphs indicates the number of vehicles in the same cell at the same time. In the first scenario, the four vehicles choose the same minimum travel time route, which results in limited spatiotemporal coverage in Figure \ref{fig:sub_toy_routeocc}. This indicates that all vehicles enter and exit the same cell at the same time and provide redundant traffic information. In the second scenario, the four vehicles choose different routes to maximize spatiotemporal coverage in Figure \ref{fig:sub_toy_coverocc}. It is obvious that routing strategies in the second scenario can greatly improve the efficiency and quality of traffic monitoring, given the same number of vehicles. However, the total travel time is increased as a trade-off, shown in the bottom row of the two figures.  

\begin{equation}
    c_{i, t} = (i + t)\%5 + 1
    \label{eqa:toy_net-travel_cost}
\end{equation}

\subsection{Grid Network Analysis}
\label{sec:grid_network_analysis}

\subsubsection{Experimental Setting}

\begin{figure}[pos=tbp]
    \centering
    \begin{subcaptionbox}{SUMO network\label{fig:sumo_network}}[0.45\linewidth]
        {\includegraphics[width=\linewidth]{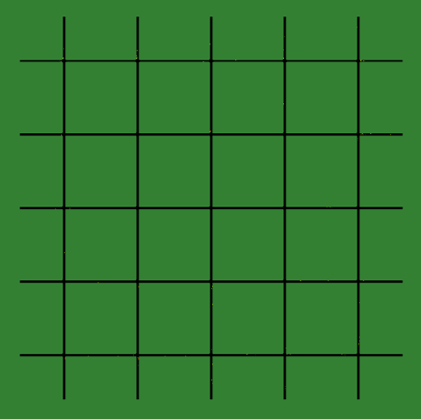}}
    \end{subcaptionbox}
    \begin{subcaptionbox}{CTM network\label{fig:ctm_network}}[0.45\linewidth]
        {\includegraphics[width=\linewidth]{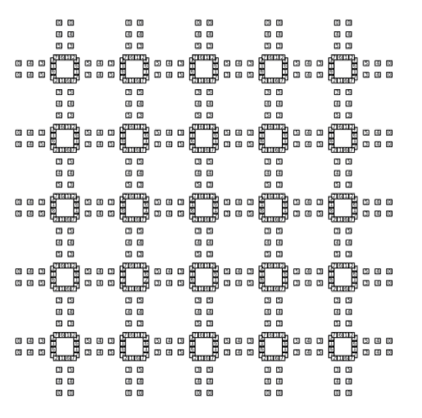}}
    \end{subcaptionbox}
    \hfill
    \caption{$5\times 5$ urban network for SUMO and CTM representation}
    \label{fig:sumo_ctm_network}
\end{figure}

\begin{figure}[pos=tbp]
    \centering
    \includegraphics[width=0.4\linewidth]{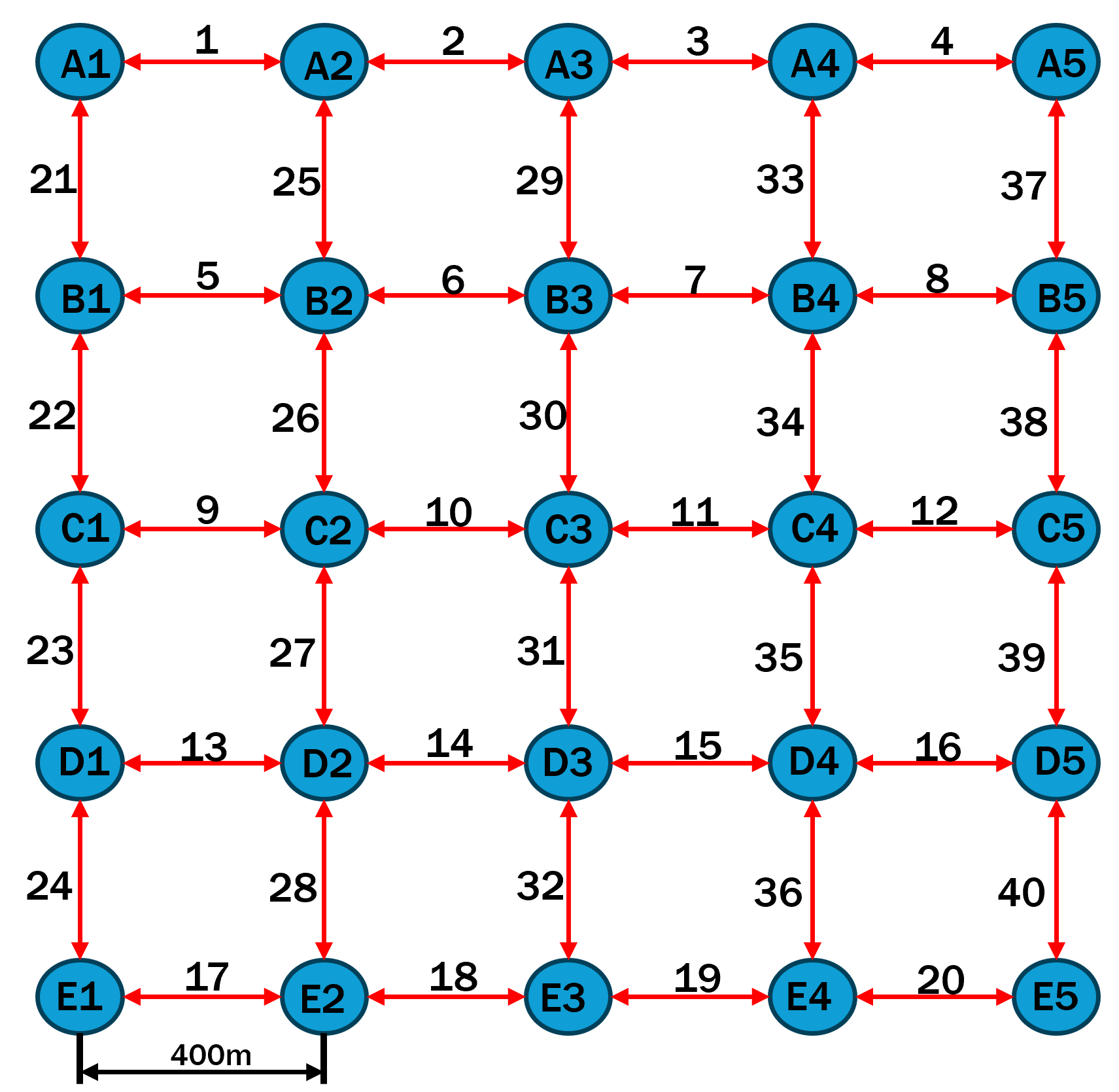}
    \caption{A $5\times 5$ grid network}
    \label{fig:grid_network}
\end{figure}

\subsubsubsection{SUMO configuration}
To conduct a comprehensive evaluation, a $5\times 5$ grid network, consisting of $40$ bidirectional two-lane links and $25$ four-approach intersections is built in SUMO, as shown in Figure \ref{fig:grid_network}. The network is further expanded with 20 entry and exit links for vehicle generation and departure, as shown in Figure \ref{fig:sumo_network}. The distance between intersections is set to $400$ meters. The entry and exit links are set to 240 meters. 
The corresponding CTM network is shown in Figure \ref{fig:ctm_network}. The cell length is set to 80 meters, calculated by the time resolution (i.e., 5s) and the free flow travel time in SUMO (i.e., 16 m/s). Each internal link contains 7 cells as shown in Figure \ref{fig:ctm_link} and each entry/exit link contains 3 cells, with a total number of 760 cells in the network.

\begin{figure}[pos=tbp]
    \centering
    \includegraphics[width=0.6\linewidth]{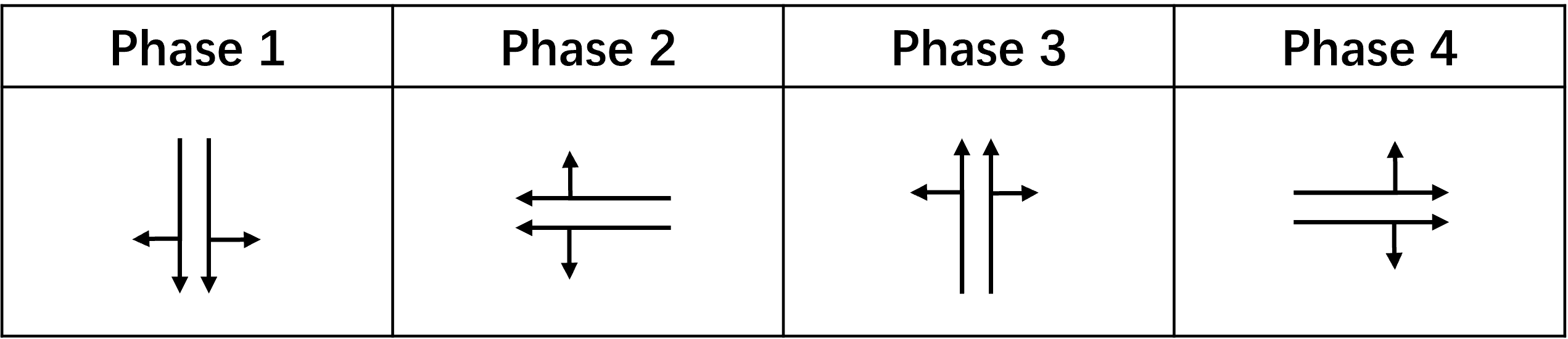}
    \caption{Fixed-time signal timing plan for intersections}
    \label{fig:phase_setting}
\end{figure}

Other settings in the SUMO simulation are listed below:
\begin{enumerate}
    \item \textbf{Traffic Demand Generation}: A total of $16$ origin–destination (OD) pairs are set in the grid network following a "diagonal pattern" that makes vehicles travel from one side of the network to the opposite side. In SUMO simulation, vehicles are generated at the entry links and demand is set to a fixed flow rate of $350 \,\mathrm{veh/h}$ for each link. All OD pairs and their corresponding traffic demands are shown in Table \ref{Table:od-demand}. The same flow rate is applied to the CTM source cells. 
    \item \textbf{Simulation Setting}: The total simulation duration is set to  $5400\,\mathrm{s}$, with the first $1800\,\mathrm{s}$ for warm up. The remaining one hour simulation ($3600\,\mathrm{s}$) time is used for data collection and performance evaluation.
    \item \textbf{Routing Strategy for Background Vehicles}: SUMO's dynamic user equilibrium (DUE) \citep{mehrabani2022proposing} algorithm is applied to generate routes for all background vehicles. Note that under different robotaxi MPRs, background vehicles' routes keep the same.
    \item \textbf{No Lane Changing Zone}: A no lane changing zone of 80 meters is set before the stop bar of each intersection. All robotaxis are not allowed to change lanes within the last $80\,\mathrm{m}$ of the approaching links, which is the same length as the intersection cell in the CTM network.
    \item  \textbf{Traffic Signal Setting}: Fixed-time traffic signal is applied at all $25$ intersections in the network. A four-phase split signal timing plan is adopted because of the shared lanes at the intersection, as shown in Figure \ref{fig:phase_setting}. The cycle length is set to 100s and evenly distributed to the four phases.
\end{enumerate}

As indicated in Figure \ref{fig:main_workflow}, different components in the proposed framework run at different time resolutions.  The SUMO network runs at a time resolution of $0.1\,\mathrm{s}$. The CTM model updates the traffic state every $5\,\mathrm{s}$ in simulation time. Every $100\,\mathrm{s}$ (i.e., every 20 CTM steps), the dynamic routing model is triggered to update the routes for all robotaxis in the network. In time steps when the dynamic routing algorithm is not triggered, the CTM only predicts one step ahead to keep up with the simulation environment. In time steps when the dynamic routing model is executed, CTM calculates traffic states for the next 300 steps, which is equivalent to the planning horizon $T$. Then the time dependent travel time can be calculated based on the predicted speed of each cell .

\subsubsubsection{Robotaxi De-routing Budget}
Although Equation \ref{eqa:single visit} prevents robotaxi subtour in the formulation, it may still be generated from later planning results because each route planning is executed independently and uses the robotaxis' current locations as the origins. Especially when coverage weight $\alpha_2$ is set to a large value, the optimization model may encourage robotaxis to de-route as much as possible and result in continuous subtours before reaching to the destination. To avoid this situation, a "de-routing budget" of each robotaxi is assigned.  To simplify the complexity of routing selection, robotaxis are only allowed to extend two more links (800m) beyond current remaining route length. Moreover, when the remaining travel distance is greater than six links (2400m), the de-routing budget is set to 0 to enforce it to choose a shortest distance route. Finally, when a robotaxi has traveled more than 12 links (4800m) from its origin node, the de-routing budget is also set to 0. Based on the de-routing rules, a set of candidate routes are generated for each robotaxi before solving the dynamic routing problem. This will also greatly reduce the feasible space for the routing decision variable $x$ and $z$. 

\begin{table}
\centering
\caption{Origin-destination (OD) demand for each node in the grid network}
\label{Table:od-demand}
\begin{tabular}{ccl @{\hspace{2cm}} ccl} 
\hline\hline
\textbf{Origin} & \textbf{Destination} & \textbf{Flow Rate} & \textbf{Origin} & \textbf{Destination} & \textbf{Flow Rate} \\
\hline
$A1$ & $E5$ & $700 \ \mathrm{veh/h}$ & $E3$ & $A3$ & $350 \ \mathrm{veh/h}$ \\
$A2$ & $E4$ & $350 \ \mathrm{veh/h}$ & $E2$ & $A4$ & $350 \ \mathrm{veh/h}$ \\
$A3$ & $E3$ & $350 \ \mathrm{veh/h}$ & $E1$ & $A5$ & $700 \ \mathrm{veh/h}$ \\
$A4$ & $E2$ & $350 \ \mathrm{veh/h}$ & $D1$ & $B5$ & $350 \ \mathrm{veh/h}$ \\
$A5$ & $E1$ & $700 \ \mathrm{veh/h}$ & $C1$ & $C5$ & $350 \ \mathrm{veh/h}$ \\
$B5$ & $D1$ & $350 \ \mathrm{veh/h}$ & $B1$ & $D5$ & $350 \ \mathrm{veh/h}$ \\
$C5$ & $C1$ & $350 \ \mathrm{veh/h}$ & $E4$ & $A2$ & $350 \ \mathrm{veh/h}$ \\
$D5$ & $B1$ & $350 \ \mathrm{veh/h}$ & $E5$ & $A1$ & $700 \ \mathrm{veh/h}$ \\
\hline\hline
\end{tabular}
\end{table}


\subsubsection{CTM Calibration and Evaluation}

\begin{figure}[pos=tbp]
    \centering
    \includegraphics[width=0.5\linewidth]{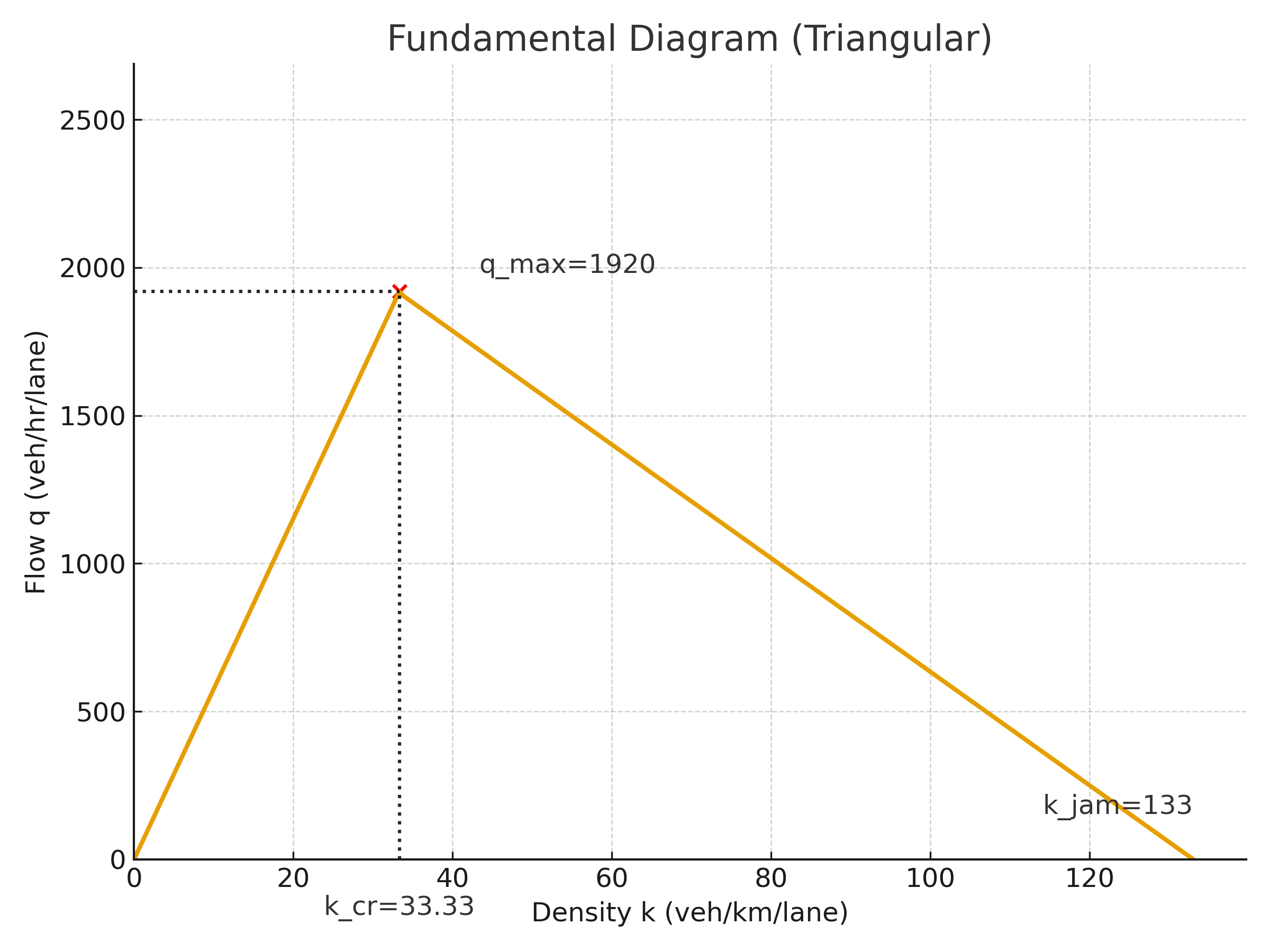}
    \caption{Calibrated fundamental diagram for cell transmission model}
    \label{fig:ctm_fd}
\end{figure}

\begin{figure}[pos=tbp]
    \centering
    \begin{subfigure}[b]{0.45\textwidth}
        \includegraphics[width=\linewidth]{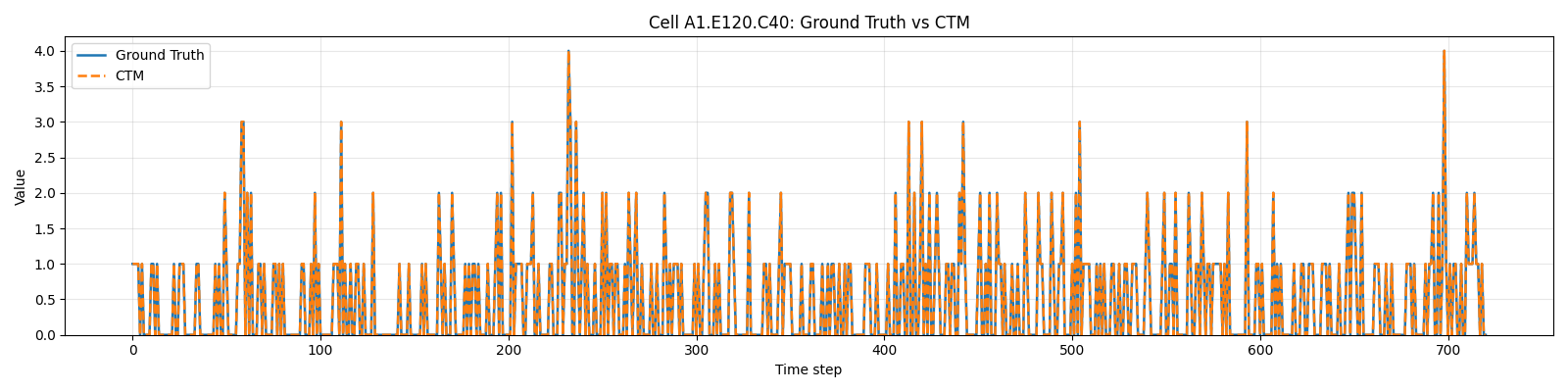}
        \caption{Source cell}
        \label{fig:sub_ctm_e103c40}
    \end{subfigure}
    \begin{subfigure}[b]{0.45\textwidth}
        \includegraphics[width=\linewidth]{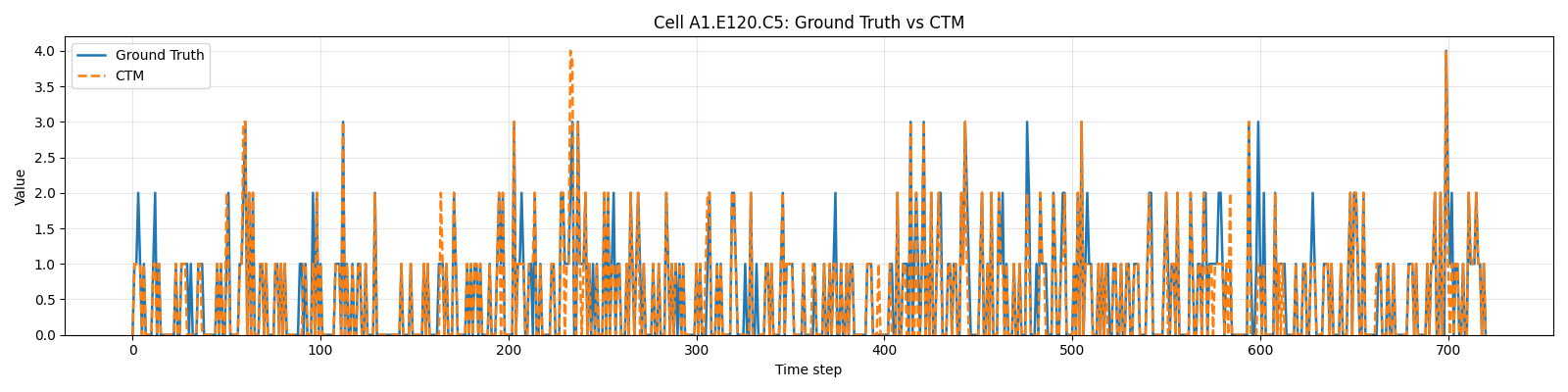}
        \caption{Cell $C5$}
        \label{fig:sub_ctm_e103c5}
    \end{subfigure}
    
    \vspace{1em}

    \begin{subfigure}[b]{0.45\textwidth}
        \includegraphics[width=\linewidth]{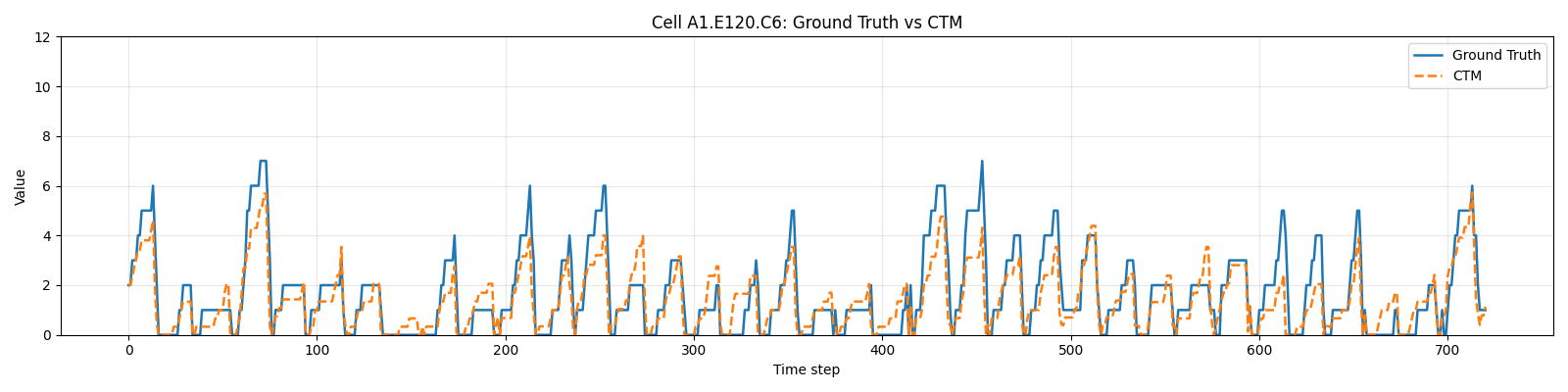}
        \caption{Cell $C6$}
        \label{fig:sub_ctm_e103c6}
    \end{subfigure}
    \begin{subfigure}[b]{0.45\textwidth}
        \includegraphics[width=\linewidth]{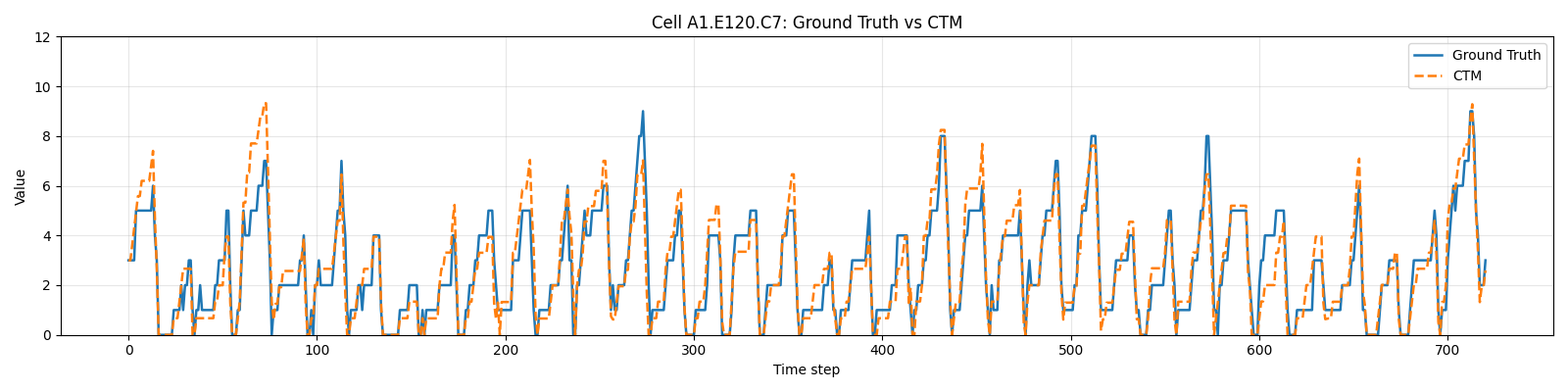}
        \caption{Cell $C7$}
        \label{fig:sub_ctm_e103c7}
    \end{subfigure}

    \caption{Comparison between CTM result and ground truth for link $E120$}
    \label{fig:ctm_comparison}
\end{figure}

CTM plays an important role by providing the predicted travel time in the routing model, its parameters (i.e., the fundamental diagram) need to be calibrated to reflect accurate traffic flow dynamics in the network. Given the settings of free flow speed as 16 m/s and jam spacing of 7.5 m/veh (equivalent to a jam density of 133 veh/km) in SUMO, we vary the values of the maximum flow rate ($q_{max}$) and the upper critical density ($k_2$) in the trapezoid fundamental diagram, to minimize the Mean-Average Error (MAE) in terms of the number of vehicles between the CTM cells and the ground truth in a one-hour simulation period. Note that in order to evaluate the CTM prediction accuracy, we load the exact number of vehicles entering the SUMO network, instead of the hourly average, to the CTM source cells at each time step. The calibrated fundamental diagram happens to be triangular with $q_{max}=1920 veh/h$ and $k_1=k_2=33.33 veh/km$, as shown in Figure \ref{fig:ctm_fd}. 
The MAE between the CTM and ground truth is $0.38 \ veh/(cell\cdot step)$ for entry links, which is an acceptable result. 
The main source of error comes from the fixed turning ratios in the CTM setting, which may not be consistent with the real turning ratios for each signal cycle.
Figure \ref{fig:ctm_comparison} shows the number of vehicles between the ground truth (blue line) and CTM prediction (yellow line) for one of the entry links that contains four cells. 

\subsubsection{Solving Time Analysis}

\begin{figure}[pos=tbp]
    \centering
    \includegraphics[width=0.7\linewidth]{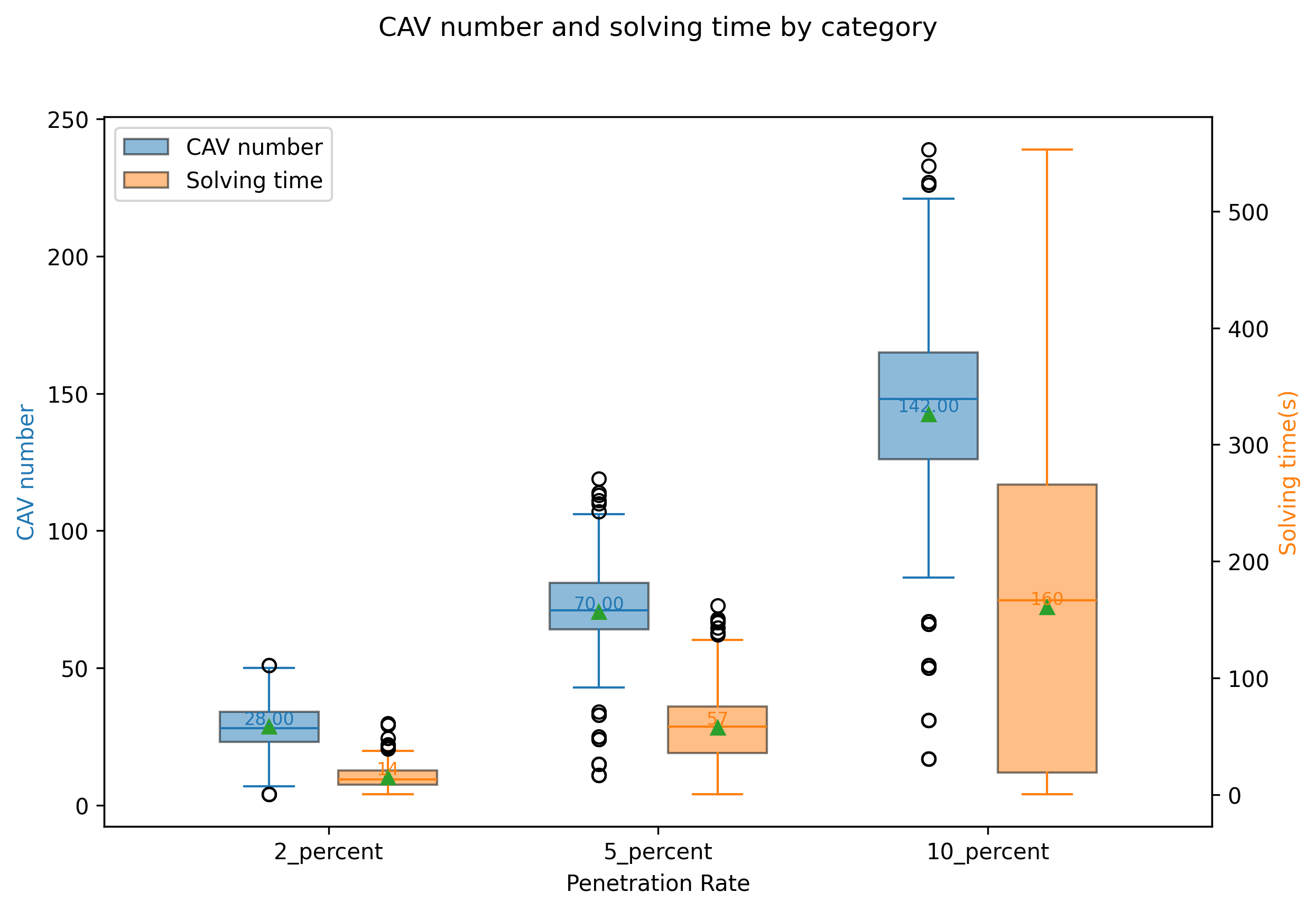}
    \caption{Solving time distribution under different MPRs}
    \label{fig:solvingtime_box}
\end{figure}

An evaluation of the dynamic routing algorithm solving time is presented in Figure \ref{fig:solvingtime_box}, which illustrates the distribution of robotaxi counts and corresponding computation time at each decision step under different MPRs. As shown in the figure, when the MPR reaches 10\%, the average solving time reaches 160 seconds, with roughly 142 vehicles being routed at each iteration. The computation time increases consistently with the number of robotaxis.
Even in the 10\% MPR scenario, which represents a relatively high robotaxi density, the average solving time remains below 200 seconds.

\subsubsection{Result and Discussion}
\label{sec:result and discussion}

\begin{table}[H]
\caption{Experiment results under different robotaxi MPRs and objective weights}
\label{Table:350_result}
\begin{center}
\resizebox{\textwidth}{!}{ 
\begin{tabular}{lccccccccc} 
\hline\hline
\textbf{Case($\alpha_1$:$\alpha_2$)} & 
\makecell{\textbf{Penetration} \\ \textbf{Rate}} & 
\makecell{\textbf{Robotaxi Avg.} \\ \textbf{Travel Time (s)}} & 
\makecell{\textbf{Robotaxi Avg.} \\ \textbf{Travel Distance (m)}} & 
\makecell{\textbf{Robotaxi Avg.} \\ \textbf{Speed (m/s)}} &
\makecell{\textbf{BV Avg.} \\ \textbf{Travel Time (s)}} & 
\makecell{\textbf{CTM MAE}} &
\textbf{Network Coverage} &
\makecell{\textbf{Vehicle} \\ \textbf{Coverage}} \\
$1:10$     & $2\%$  & 767.58  & 3026.48 & 4.16 & 825.54 & 1.2508 & 3.29\% & 10.25\% \\
$1:100$    & $2\%$  & 776.08  & 3064.15 & 4.13 & 827.55 & 1.2319 & 3.32\% & 10.16\% \\
$1:1000$   & $2\%$  & 778.39  & 3074.90 & 4.14 & 820.80 & 1.2118 & 3.37\% & 10.55\% \\
$1:1500$   & $2\%$  & 790.23  & 3089.95 & 4.20 & 822.72 & 1.1548 & 3.43\% & 10.63\% \\
$1:2000$   & $2\%$  & 775.08  & 3085.89 & 4.19 & 822.95 & 1.1961 & 3.36\% & 10.35\% \\
$1:2500$   & $2\%$  & 770.88  & 3041.93 & 4.16 & 822.11 & 1.1573 & 3.40\% & 10.57\% \\
$1:3000$   & $2\%$  & 794.30  & 3043.79 & 4.08 & 829.81 & 1.1975 & 3.44\% & 10.65\% \\
$1:5000$   & $2\%$  & 755.62  & 3072.55 & 4.25 & 817.65 & 1.1494 & 3.43\% & 10.58\% \\
$1:10000$  & $2\%$  & 1077.18 & 4049.58 & 4.02 & 829.39 & 1.0977 & 4.54\% & 14.07\% \\
$1:100000$ & $2\%$  & 1103.99 & 4034.90 & 3.89 & 843.86 & 1.0712 & 4.58\% & 14.33\% \\
\hline
$1:10$     & $5\%$  & 775.57  & 3100.63 & 4.16 & 795.35 & 1.1776 & 8.03\% & 24.65\% \\
$1:100$    & $5\%$  & 765.75  & 3132.75 & 4.25 & 786.69 & 1.1288 & 7.90\% & 24.27\% \\
$1:1000$   & $5\%$  & 773.00  & 3152.45 & 4.27 & 785.39 & 1.0893 & 8.26\% & 25.22\% \\
$1:1500$   & $5\%$  & 786.30  & 3179.35 & 4.21 & 787.82 & 1.0615 & 8.57\% & 26.16\% \\
$1:2000$   & $5\%$  & 815.31  & 3320.80 & 4.23 & 790.57 & 0.9879 & 8.83\% & 26.67\% \\
$1:2500$   & $5\%$  & 912.40  & 3759.65 & 4.28 & 785.30 & 0.9140 & 9.67\% & 28.85\% \\
$1:3000$   & $5\%$  & 1033.81 & 4092.76 & 4.14 & 817.00 & 0.8959 & 10.61\% & 31.49\% \\
$1:5000$   & $5\%$  & 1034.10 & 4064.09 & 4.08 & 813.68 & 0.8778 & 10.64\% & 31.74\% \\
$1:10000$  & $5\%$  & 1080.94 & 4197.27 & 4.09 & 808.37 & 0.8820 & 10.76\% & 32.10\% \\
$1:100000$ & $5\%$  & 1064.74 & 4073.07 & 4.03 & 821.57 & 0.8780 & 10.75\% & 32.45\% \\
\hline
$1:10$     & $10\%$ & 796.66  & 3083.08 & 4.02 & 791.89 & 0.9692 & 14.21\% & 41.81\% \\
$1:100$    & $10\%$ & 791.04  & 3115.05 & 4.09 & 784.14 & 0.9220 & 14.48\% & 42.80\% \\
$1:1000$   & $10\%$ & 820.95  & 3461.67 & 4.34 & 766.34 & 0.7239 & 16.79\% & 47.63\% \\
$1:1500$   & $10\%$ & 907.72  & 3794.28 & 4.31 & 784.59 & 0.6763 & 18.11\% & 50.19\% \\
$1:2000$   & $10\%$ & 942.91  & 3854.46 & 4.22 & 789.88 & 0.6809 & 18.35\% & 51.19\% \\
$1:2500$   & $10\%$ & 962.54  & 3910.00 & 4.21 & 789.68 & 0.6715 & 18.47\% & 51.85\% \\
$1:3000$   & $10\%$ & 982.61  & 3986.90 & 4.20 & 791.63 & 0.6595 & 18.91\% & 52.49\% \\
$1:5000$   & $10\%$ & 988.06  & 4021.07 & 4.19 & 793.46 & 0.6617 & 18.76\% & 52.44\% \\
$1:10000$  & $10\%$ & 1007.29 & 4046.73 & 4.16 & 798.65 & 0.6516 & 19.08\% & 52.84\% \\
$1:100000$ & $10\%$ & 1023.41 & 4106.12 & 4.15 & 799.90 & 0.6586 & 19.08\% & 53.00\% \\
\hline

$1:10$ & $10\% \ \text{no budget}$ & 763.94 & 2939.37 & 4.01 & 784.09 & 0.9918 & 13.69\% & 41.50\% \\
$1:100$ & $10\% \ \text{no budget}$ & 762.78 & 2941.44 & 4.02 & 777.76 & 0.9552 & 13.84\% & 41.76\% \\
$1:1000$ & $10\% \ \text{no budget}$ & 725.85 & 2938.10 & 4.20 & 763.33 & 0.7856 & 15.03\% & 45.48\% \\
$1:1500$ & $10\% \ \text{no budget}$ & 723.84 & 2947.76 & 4.24 & 759.09 & 0.7561 & 15.00\% & 45.99\% \\
$1:2000$ & $10\% \ \text{no budget}$ & 730.21 & 2939.76 & 4.18 & 765.42 & 0.7607 & 15.12\% & 45.80\% \\
$1:2500$ & $10\% \ \text{no budget}$ & 732.63 & 2947.76 & 4.18 & 758.38 & 0.7431 & 15.07\% & 46.26\% \\
$1:3000$ & $10\% \ \text{no budget}$ & 733.56 & 2943.39 & 4.19 & 756.58 & 0.7521 & 15.18\% & 46.42\% \\
$1:5000$ & $10\% \ \text{no budget}$ & 736.87 & 2938.55 & 4.16 & 765.42 & 0.7560 & 15.27\% & 46.47\% \\
$1:10000$ & $10\% \ \text{no budget}$ & 737.55 & 2939.76 & 4.14 & 763.46 & 0.7461 & 15.09\% & 46.51\% \\
$1:100000$ & $10\% \ \text{no budget}$ & 743.15 & 2942.92 & 4.14 & 763.73 & 0.7489 & 15.22\% & 46.81\% \\

\hline\hline
\end{tabular}
}
\end{center}
\end{table}

\begin{figure}[pos=tbp]
    \centering
    \includegraphics[width=\textwidth]{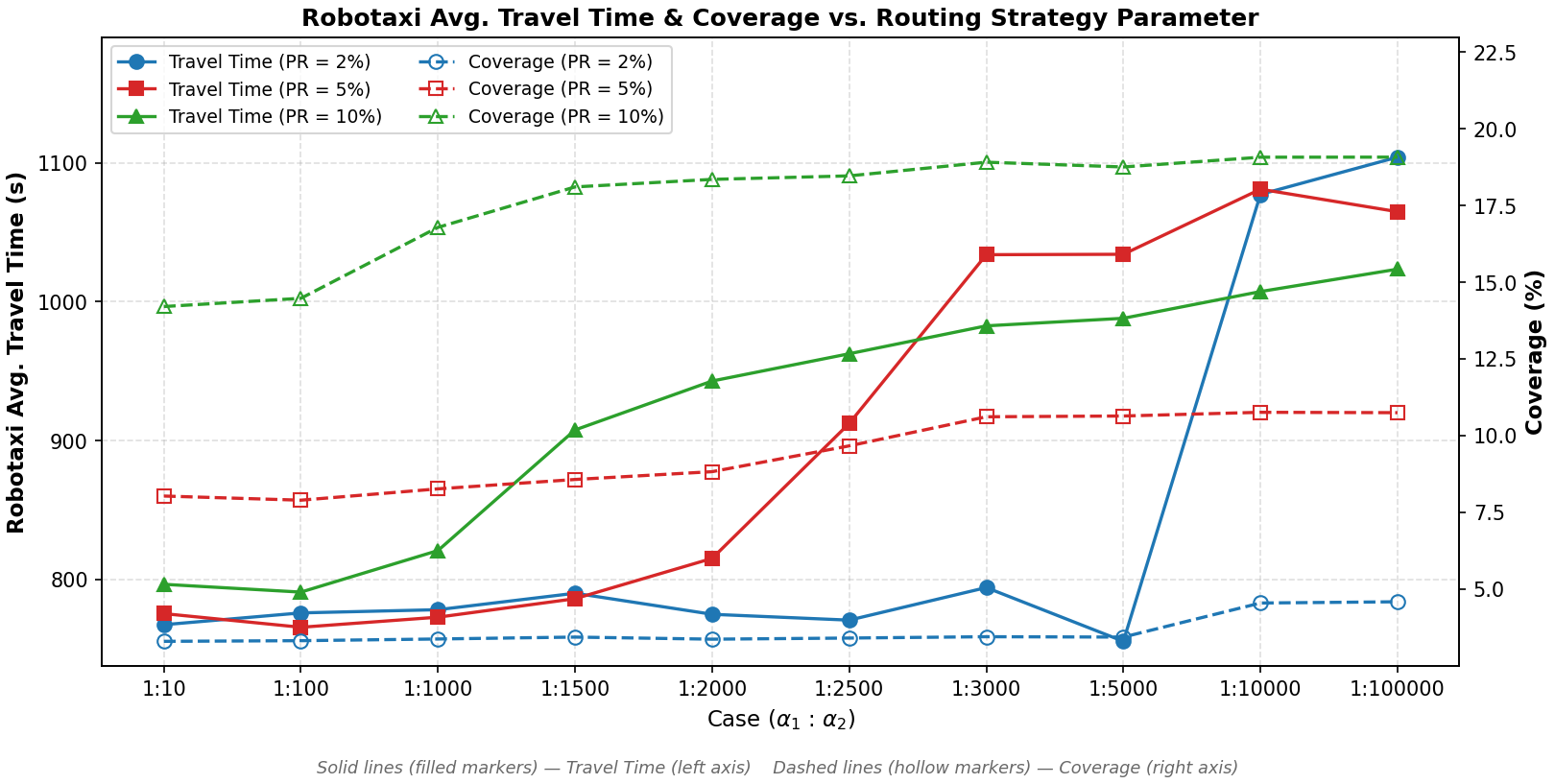}
    \caption{Robotaxi average travel time (left axis, solid lines with filled markers)
and network coverage (right axis, dashed lines with hollow markers) under different
routing strategies $\alpha_1 : \alpha_2$ and MPRs. Blue, red, and green
correspond to MPRs of $2\%$, $5\%$, and $10\%$, respectively.}
    \label{fig:350_tt_coverage}
\end{figure}

Simulation experiments are conducted under three robotaxi MPRs (2\%, 5\%, and 10\%) with a set of weight combinations of $\alpha_1$ and $\alpha_2$ (from $(1:10)$ to $(1:100000)$). Considering the scaling differences between two objectives,the $(1:10)$ weight combination means that robotaxis only try to minimize travel time and higher $\alpha_2$ indicates the increasing priority of robotaxis considering the monitoring objective. Another set of simulation experiments are conducted under 10\% robotaxi MPR without de-routing budget, 
meaning that robotaxis can only choose shortest distance paths. Experiment results are summarized in Table \ref{Table:350_result}. We report both network coverage and vehicle coverage metrics to evaluate monitoring performance. The network coverage represents the percentage of spatiotemporal cells that are monitored while vehicle coverage means the percentage of vehicles in the network are observed. 
Generally, we obtain the following observations:

\begin{enumerate}

    \item \textbf{Coverage scales with $\alpha_2$.} Increasing network coverage weight $\alpha_2$ raises spatiotemporal coverage under all MPRs, achieved through a mix of longer routes and more diverse route combinations. 

    \item \textbf{A win-win situation emerges without de-routing budget.} Higher coverage reduces MAEs in CTM, which leads to more accurate time-dependent travel time prediction. When the de-routing budget is removed, more accurate prediction could effectively reduce robotaxi travel time. This is contradiction with the intuition that travel-time minimization and coverage maximization are always competing objectives.

    \item \textbf{Coverage is not a sufficient proxy for sensing power.}
    Although spatiotemporal coverage and sensing power \citep{okeeffe2019sensing} are positively correlated, the spatial distribution of observations and the realized fleet speed also play decisive roles in determining the sensing power.  

    
    \item \textbf{Routing strategy adapts to fleet size.} The proposed framework adjusts the routing strategy over different robotaxi MPRs. At low MPRs, diversifying route combinations across a limited fleet size yields little coverage gain relative to extending individual travel distances. When the fleet size gets larger, choosing diversified routes becomes a competitive strategy.
    
\end{enumerate}
Detailed analysis of each point is provided in the subsections below.



\subsubsubsection{Robotaxi Mobility and Spatiotemporal Coverage}
\label{sec:robotaxi mobility and spatiotemporal coverage}


Table~\ref{Table:350_result} shows spatiotemporal coverage keeps increasing with the increase of $\alpha_2$, under all robotaxi MPRs and de-routing budgets. This indicates the effectiveness of our algorithm in improving the spatiotemporal coverage by changing robotaxis' routing choices. 
Additionally, across all MPRs, the increase of both travel time and coverage stops at a certain value of  $\alpha_2$, beyond which further weighting yields no additional gain. As shown in Figure~\ref{fig:350_tt_coverage}, we observe two primary findings: 1) The robotaxi average travel time converges to a similar level (about 1000--1100\,s) across MPRs as $\alpha_2$ increases, while coverage converges to a penetration-rate-dependent ceiling, indicating that the robotaxi fleet has reached its maximum achievable coverage performance. 2) The value of $\alpha_2$ when this ceiling is reached differs under different MPRs (10,000 for 2\% MPR, 3,000 for 5\% MPR, and 1,500 for 10\% MPR). This can be attributed to the fact that higher MPRs yield larger sizes of robotaxi fleet, thereby giving the routing algorithm greater flexibility in selecting vehicles for re-routing. With this added flexibility, the system can either extend travel distances or identify better routing combinations. Conversely, when the robotaxi fleet is small, the coordination of multiple vehicles for better routing combinations is highly constrained. In such cases, the objective function mainly drives vehicles to increase their travel distance once $\alpha_2$ exceeds a threshold. Moreover, the growth rate of travel time decreases as the MPR increases, which also shows that increased fleet number diversifies the path combinations in the routing algorithm. 


With de-routing budget, the average travel time of the robotaxi also relates to travel distance, making it difficult to isolate the impact of traffic state prediction. Therefore, we conducted the "no budget" cases under 10\% MPR, where all other factors are eliminated by setting the total de-routing budget to be zero. With the increase of $\alpha_2$, the improved network coverage reduces the CTM MAE from 0.99 to 0.74, while the robotaxi average travel distance remains almost constant around $\sim 2940$\,m. Meanwhile, the average travel time decreases from 763.94\,s at $\alpha_2 = 10$ to 723.84\,s at $\alpha_2 = 1500$, demonstrating that improved traffic state estimation alone leads to average travel time reduction. When $\alpha_2$ is further increased, the average travel time starts to increase again, due to the dominance of the coverage objective. This demonstrates a virtuous cycle between coverage and average travel time objectives: enhanced coverage improves estimation quality (lower CTM MAE), which enables more efficient route selection and further reduces travel time, without requiring any detour from the shortest paths. This results show that the two objectives can be improved at the same time under proper weighting parameters, creating a "win-win" condition. This observation has important implications in real-world deployment of robotaxi fleets as drive-by sensors, which will be discussed in Section \ref{sec:implication}.

\subsubsubsection{Sensing Power Evaluation}
\label{sec:sensing power evaluation}

To better align the proposed monitoring objective with existing literature, we calculate the \textbf{\textit{sensing power}} of the robotaxi fleet. 
Following ~\citep{okeeffe2019sensing}, the sensing performance of a vehicle fleet can be quantified by its sensing power, defined as the fraction of road segments that are observed within a given time period $T$. 
As shown in \eqref{eqa:sensing_power}, let $N_S$ denote the number of road segments, $p_i$ is the empirical popularity of segment $i$ (i.e., ratio between number of times segment $i$ being sensed and total number of segments being sensed during the time period), and $\langle B \rangle$ is the average number of segments traversed by a vehicle during $T$. Using this formulation, the expected sensing power of a vehicle fleet with size $N_V$ is given by the following expression~\citep{okeeffe2019sensing}:

\begin{equation}
    \langle C \rangle_{N_V}
    \approx 
    1 - \frac{1}{N_S} \sum_{i=1}^{N_S} 
    (1 - p_i)^{\, \langle B \rangle N_V},
    \label{eqa:sensing_power}
\end{equation}


Given a road network and the number of robotaxi, the values of $N_s$ and $N_V$ are fixed numbers, so the shape of the $\langle C \rangle_{N_V}$ curve is governed by two parameters: $\langle B \rangle$ and $\{p_i\}$. Firstly, $\langle B \rangle$ is equivalent to the average travel distance per unit time, which is directly proportional to robotaxi average speed. Higher vehicle speeds yield larger $\langle B \rangle$ values, elevating the entire sensing power curve, with the exponent term $(1 - p_i)^{\langle B \rangle N_V}$ in Eq.~\eqref{eqa:sensing_power}. 
Second, the segment popularity distribution $\{p_i\}$ influences sensing power through both its uniformity. By the Cauchy-Schwarz inequality\citep{cauchy1821cours}, the term $\sum_{i=1}^{N_S} (1-p_i)^{\langle B \rangle N_V}$ is minimized when $p_i$ values are uniformly distributed across all segments, which maximizes $\langle C \rangle_{N_V}$. 
Based on this model, a routing strategy that generates higher sensing power should satisfy the following conditions: 
(1) number of visits of each road segment (or cell) are distributed more uniformly across the network (i.e., more uniform $p_i$ distribution); and (2) higher average speed of the fleet vehicles. 

\begin{figure}[pos=tbp]
    \centering
    
    \begin{subfigure}[t]{0.48\textwidth}
        \includegraphics[width=\textwidth]{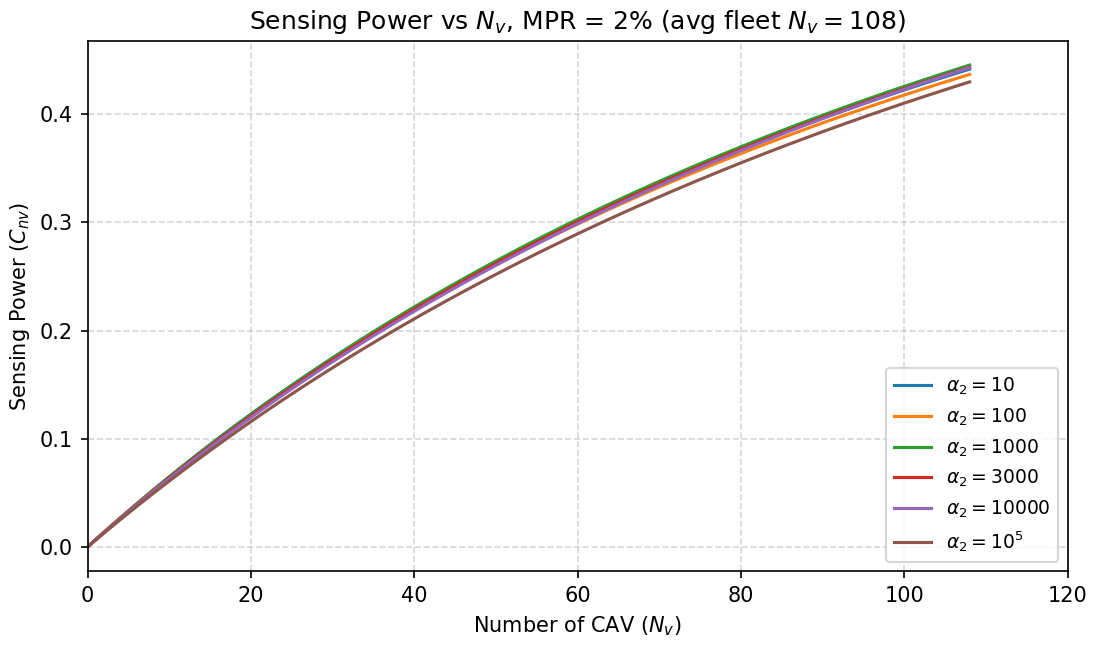}
        \subcaption*{}
        \label{Fig:sensing_power_pr2}
    \end{subfigure}
    \hfill
    \begin{subfigure}[t]{0.48\textwidth}
        \includegraphics[width=\textwidth]{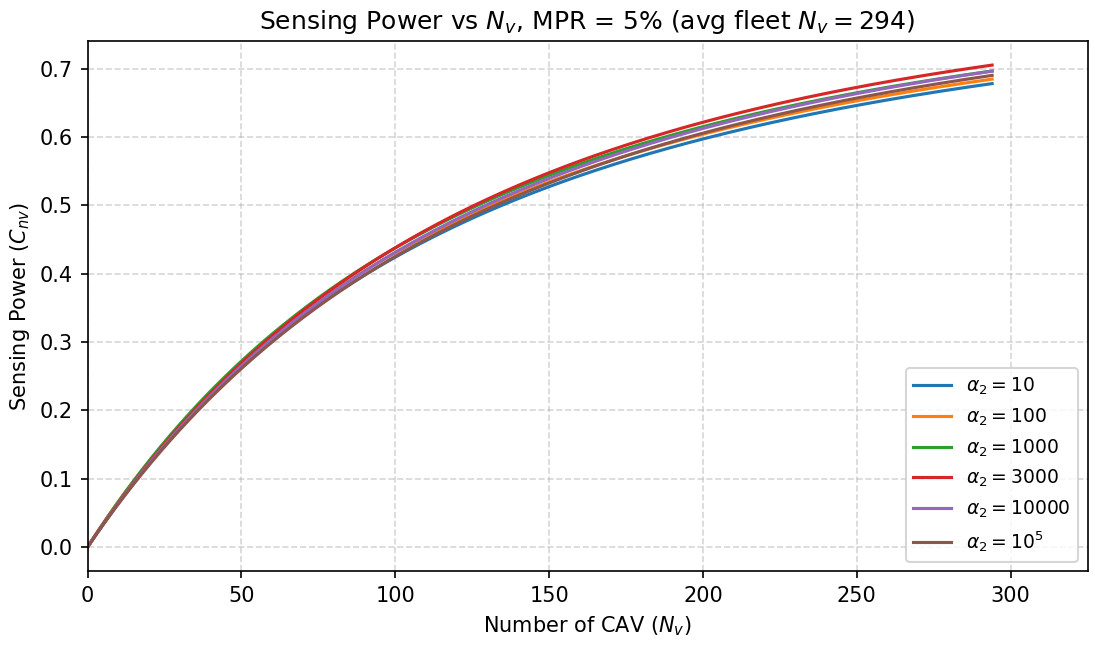}
        \subcaption*{}
        \label{Fig:sensing_power_pr5}
    \end{subfigure}
    
    \vspace{0.8em}
    
    \begin{subfigure}[t]{0.48\textwidth}
        \includegraphics[width=\textwidth]{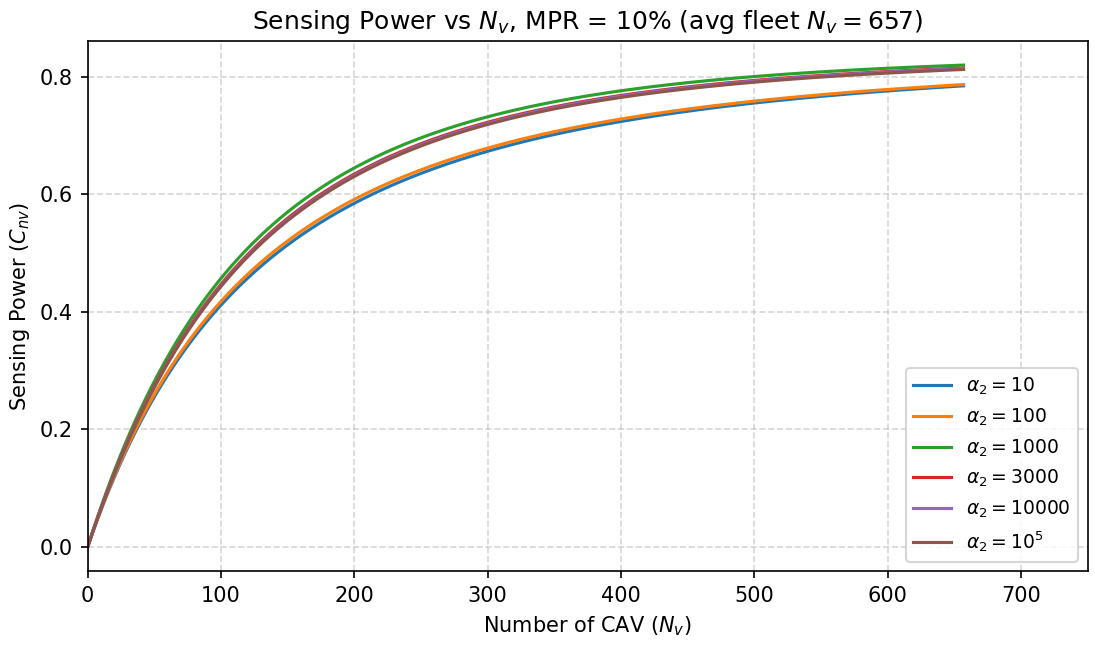}
        \subcaption*{}
        \label{Fig:sensing_power_pr10}
    \end{subfigure}
    \hfill
    
    \caption{Sensing power curves over the coverage weighting parameter $\alpha_2$ under four different settings. Top-left: MPR $= 2\%$; top-right: MPR $= 5\%$; bottom: MPR $= 10\%$.} 
    \label{fig:sensing_power}
\end{figure}

The sensing power under different robotaxi MPRs during the one hour evaluation time are plotted in Figure \ref{fig:sensing_power}. The curve shows an exponential pattern of the sensing power over the fleet size $N_v$. Higher curves indicate that the routing strategy can more effectively maximize the robotaxi fleet's drive-by sensing abilities. In general, increasing robotaxi MPR raises the upper bound of sensing power as more drive-by sensors are assigned. Under each MPR, the sensing power generally increases with $\alpha_2$. However, the highest sensing power does not appear under the highest $\alpha_2$ value, but peaks at the intermediate values vary from MPR ($\alpha_2=10000$ at 2\%, $\alpha_2=3000$ at 5\% and $\alpha_2=1000$ at 10\%). 

\begin{figure}[pos=tbp]
    \centering
    \captionsetup[subfigure]{skip=2pt}

    \begin{subfigure}[b]{0.15\textwidth}
        \includegraphics[width=\linewidth]{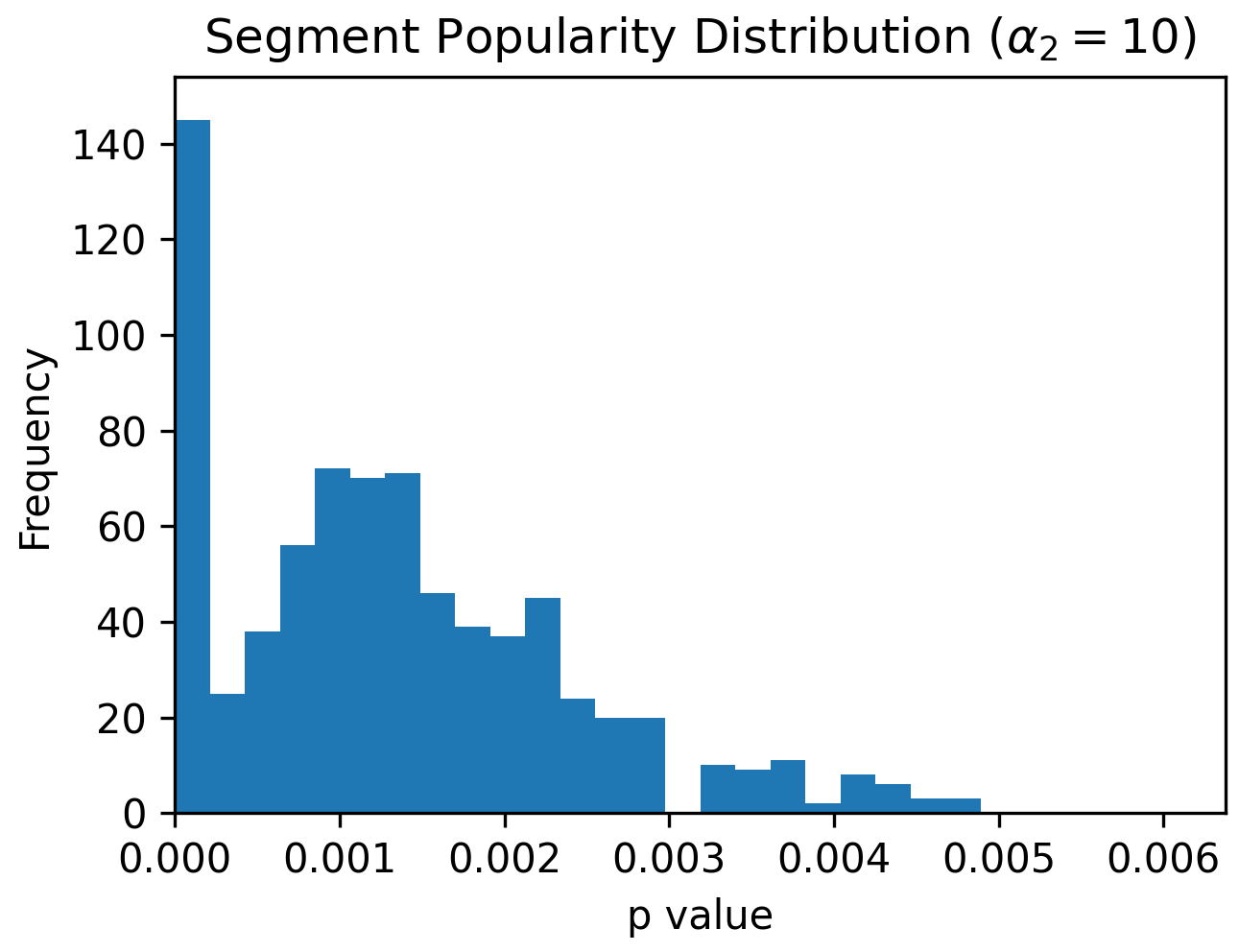}
        \caption{}\label{fig:setting1_col1}
    \end{subfigure}\hfill
    \begin{subfigure}[b]{0.15\textwidth}
        \includegraphics[width=\linewidth]{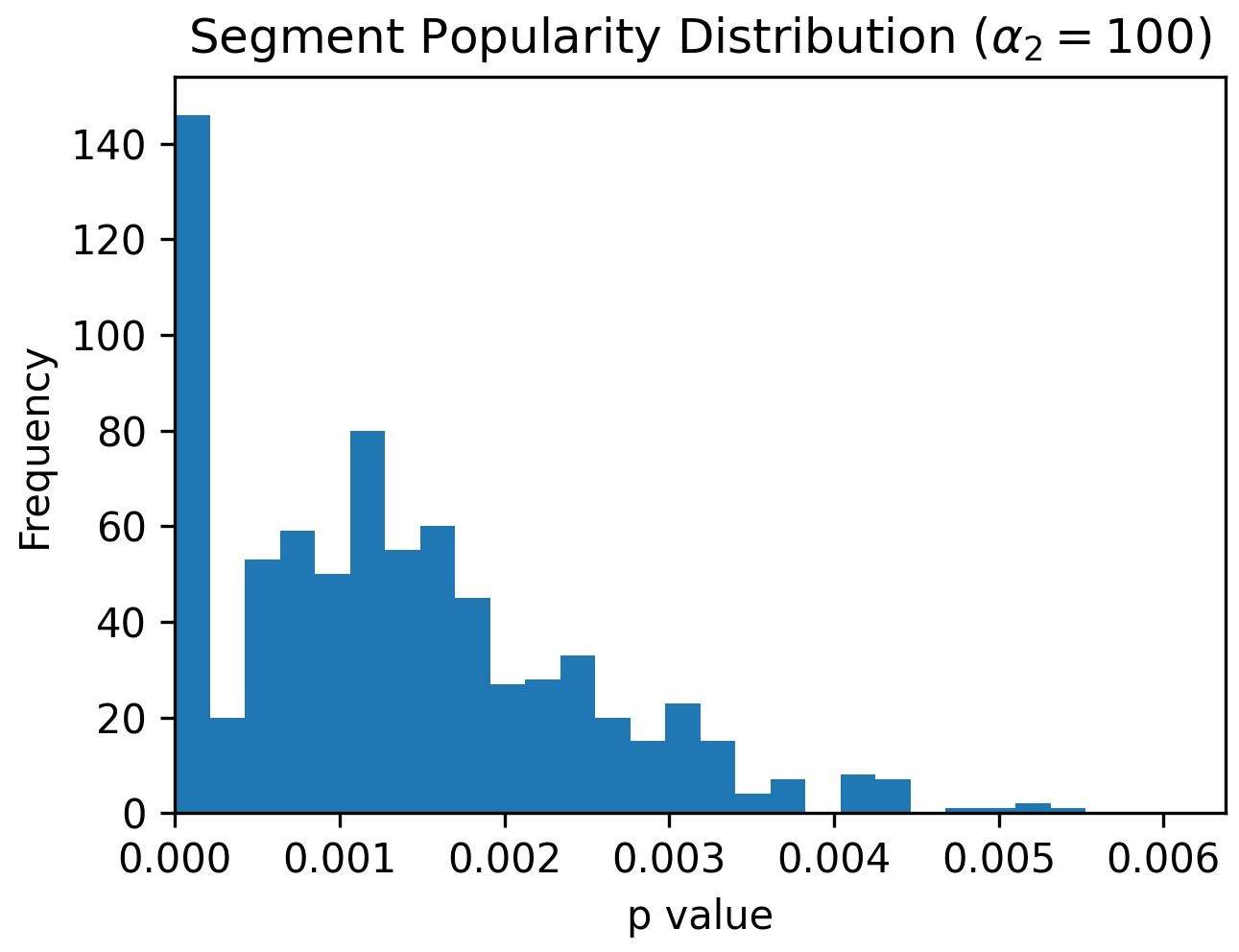}
        \caption{}\label{fig:setting1_col2}
    \end{subfigure}\hfill
    \begin{subfigure}[b]{0.15\textwidth}
        \includegraphics[width=\linewidth]{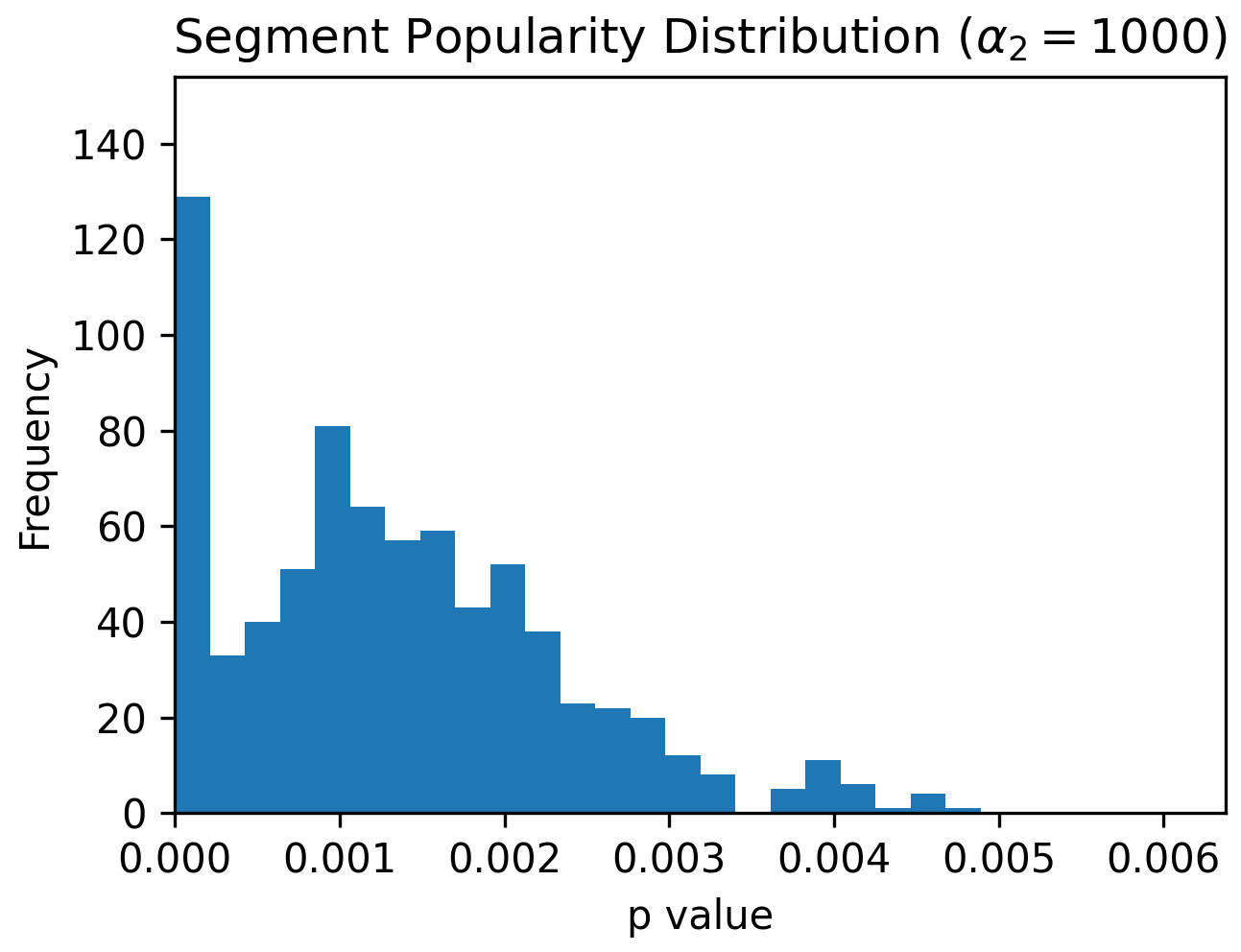}
        \caption{}\label{fig:setting1_col3}
    \end{subfigure}\hfill
    \begin{subfigure}[b]{0.15\textwidth}
        \includegraphics[width=\linewidth]{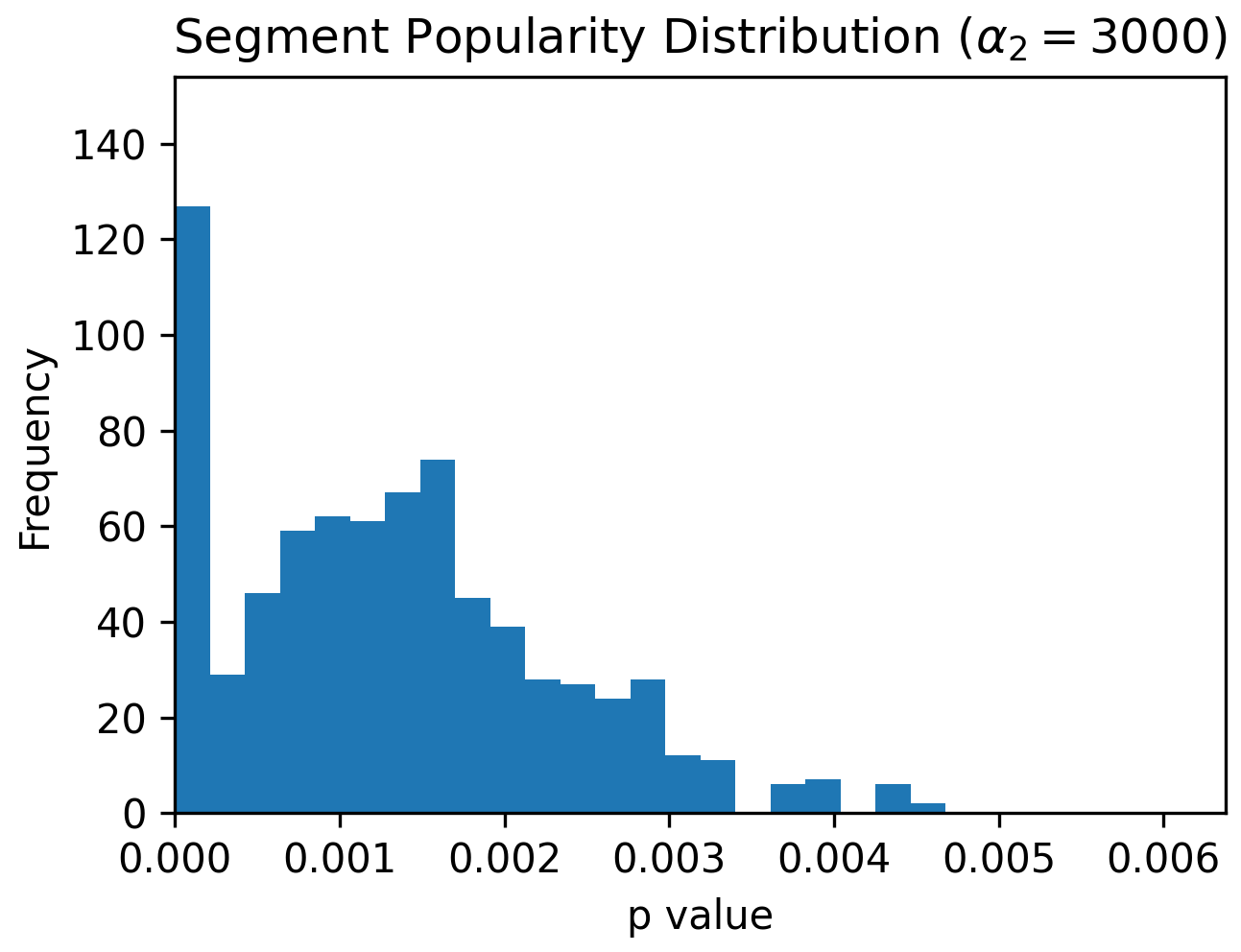}
        \caption{}\label{fig:setting1_col4}
    \end{subfigure}\hfill
    \begin{subfigure}[b]{0.15\textwidth}
        \includegraphics[width=\linewidth]{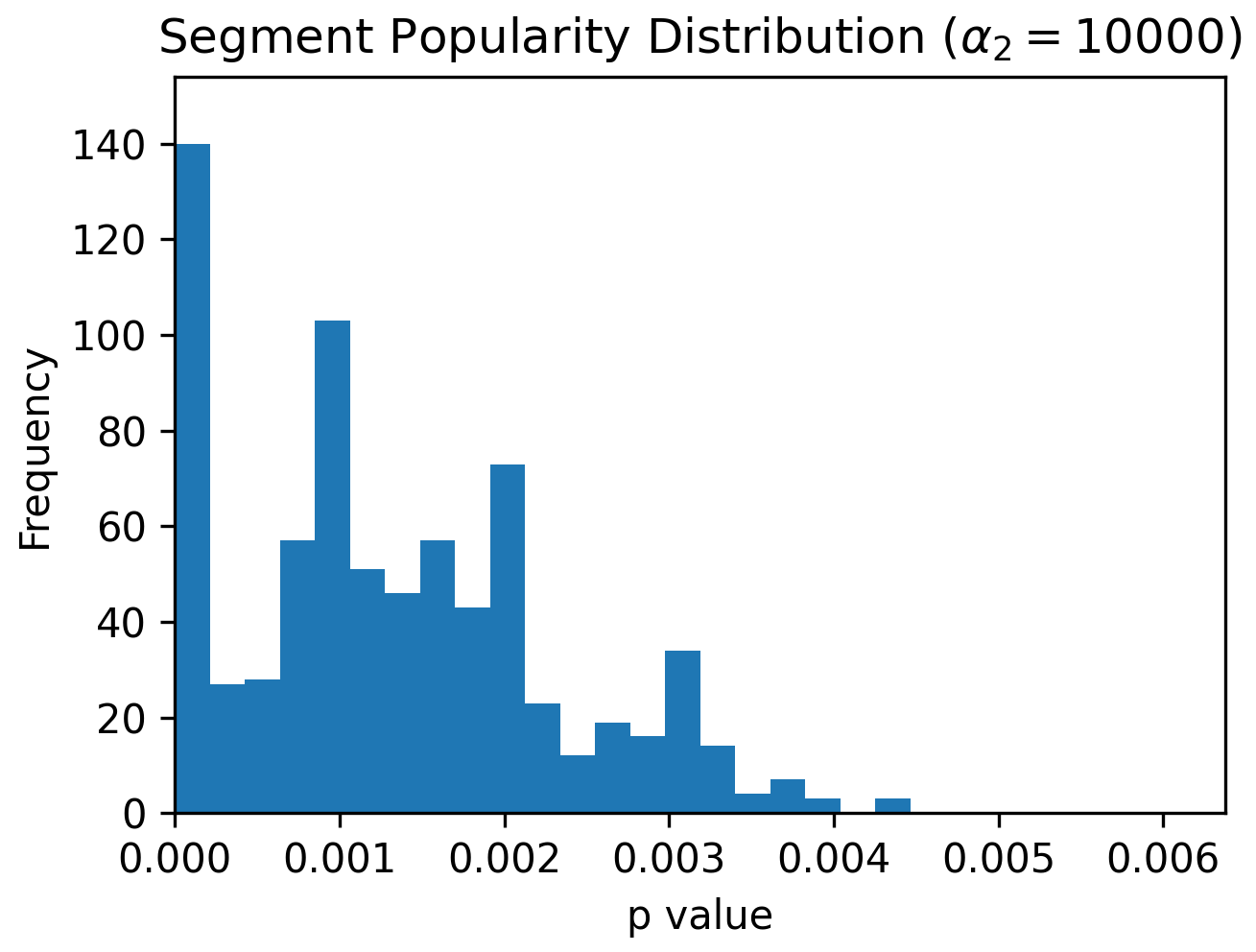}
        \caption{}\label{fig:setting1_col5}
    \end{subfigure}\hfill
    \begin{subfigure}[b]{0.15\textwidth}
        \centering
        \includegraphics[width=\linewidth]{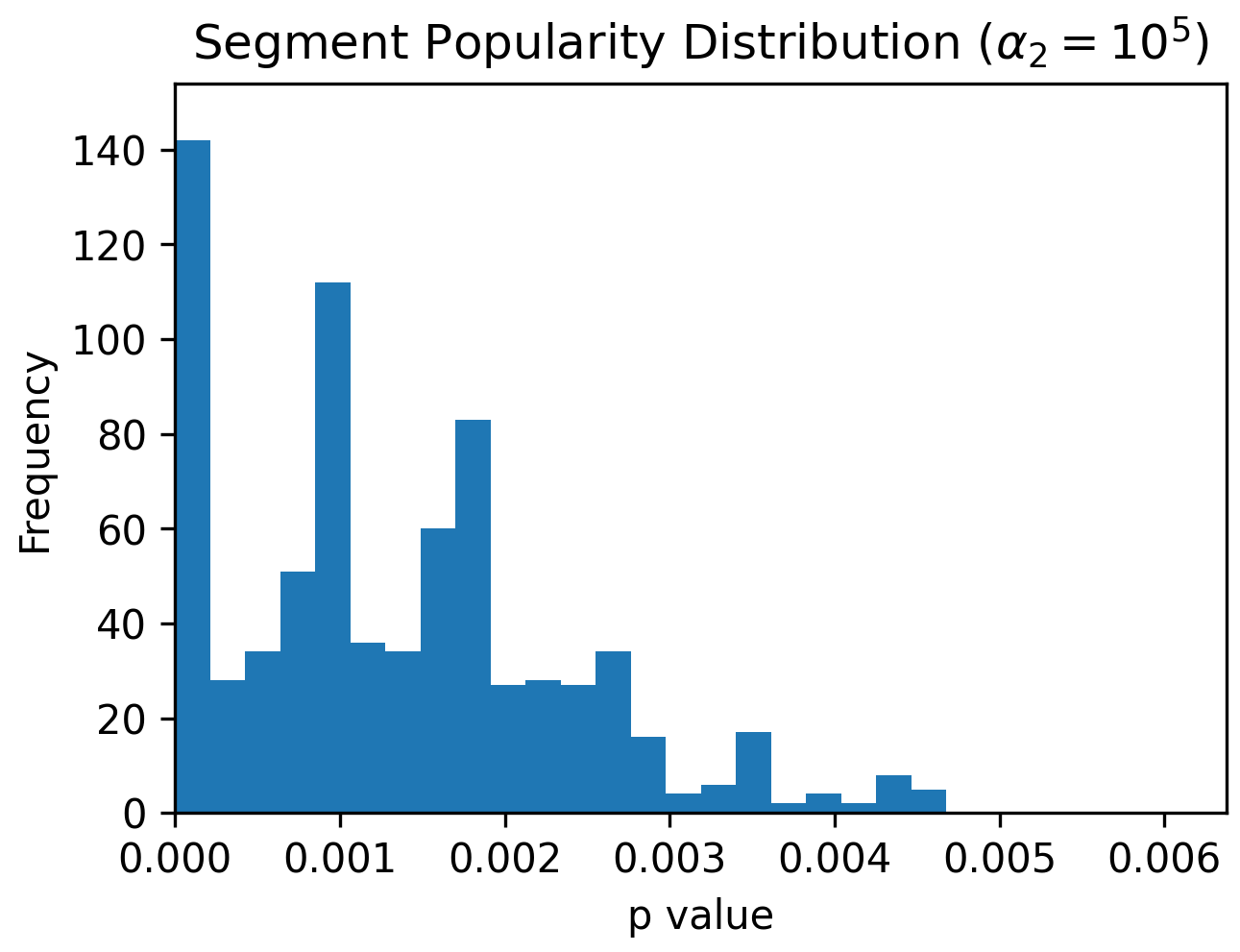}
        \caption{}\label{fig:setting1_col6}
    \end{subfigure}

    \vspace{0.3em}

    \begin{subfigure}[b]{0.15\textwidth}
        \includegraphics[width=\linewidth]{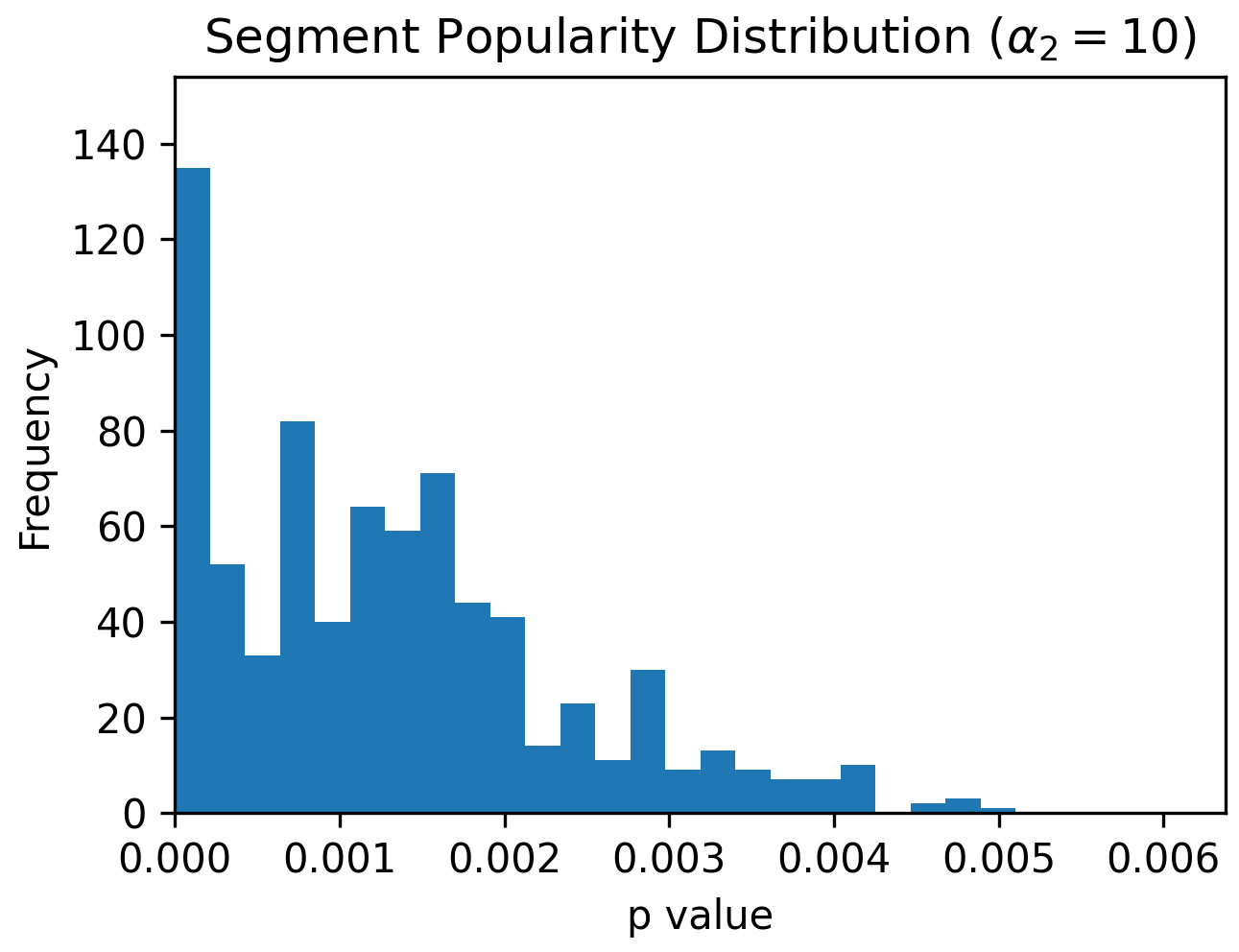}
        \caption{}\label{fig:setting2_col1}
    \end{subfigure}\hfill
    \begin{subfigure}[b]{0.15\textwidth}
        \includegraphics[width=\linewidth]{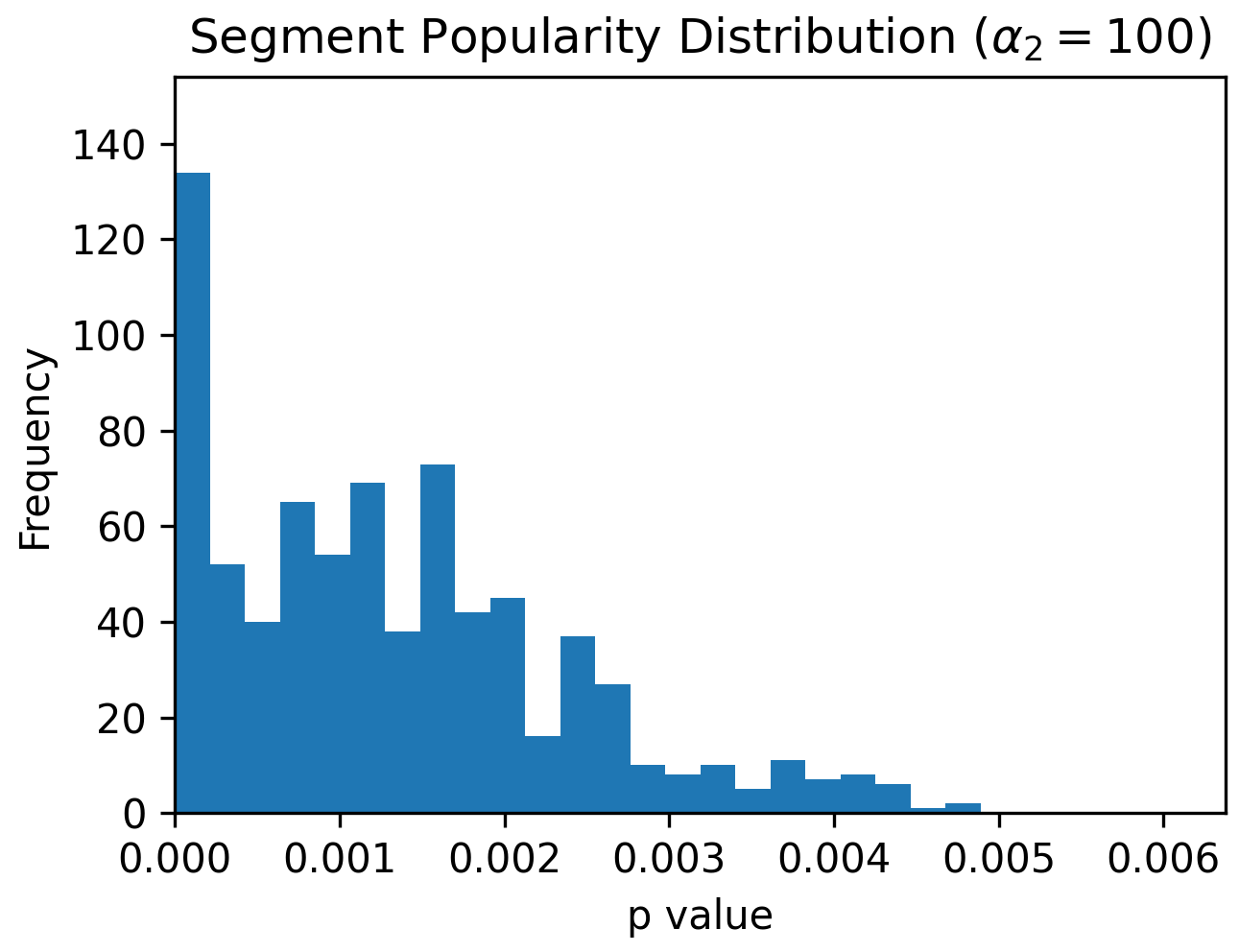}
        \caption{}\label{fig:setting2_col2}
    \end{subfigure}\hfill
    \begin{subfigure}[b]{0.15\textwidth}
        \includegraphics[width=\linewidth]{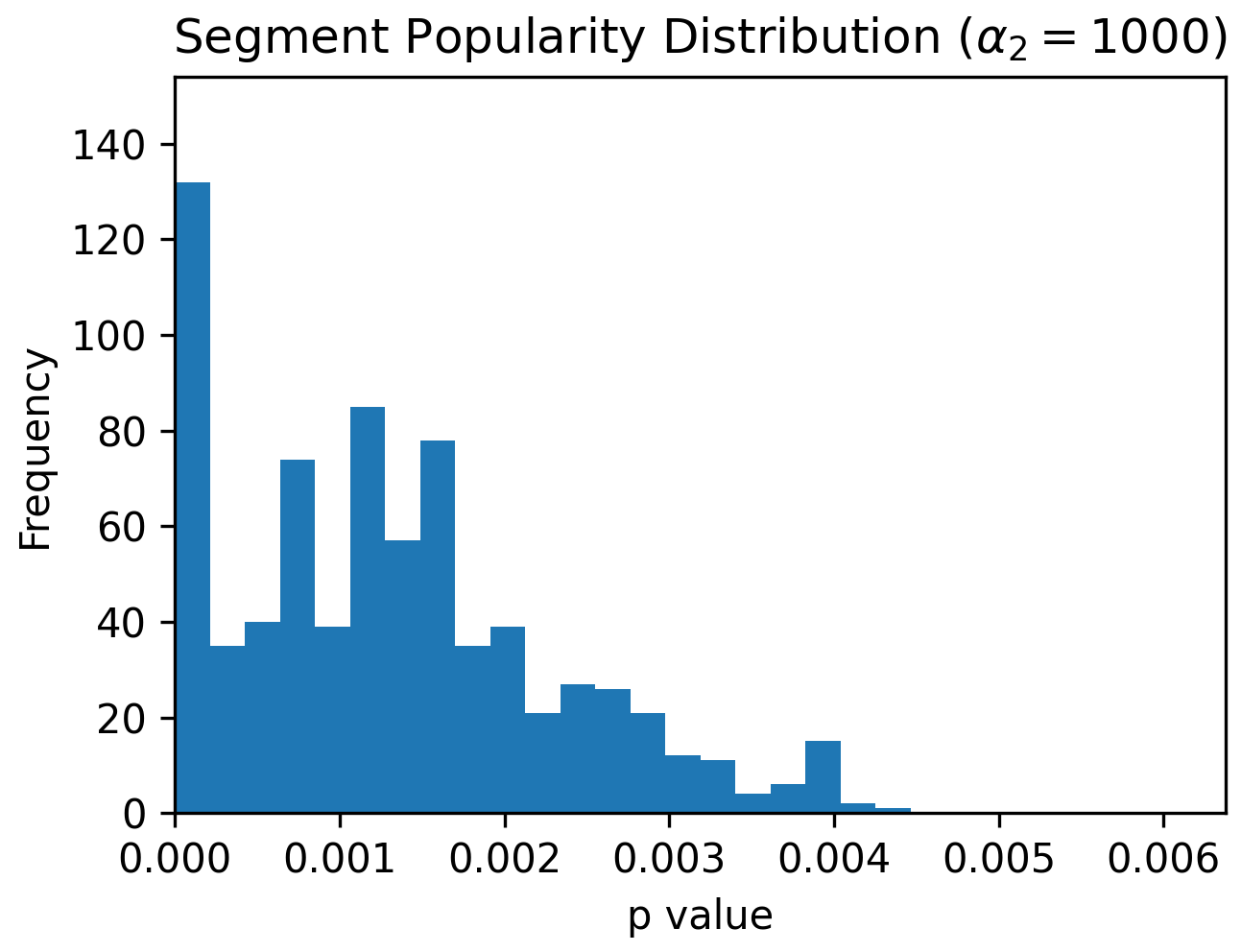}
        \caption{}\label{fig:setting2_col3}
    \end{subfigure}\hfill
    \begin{subfigure}[b]{0.15\textwidth}
        \includegraphics[width=\linewidth]{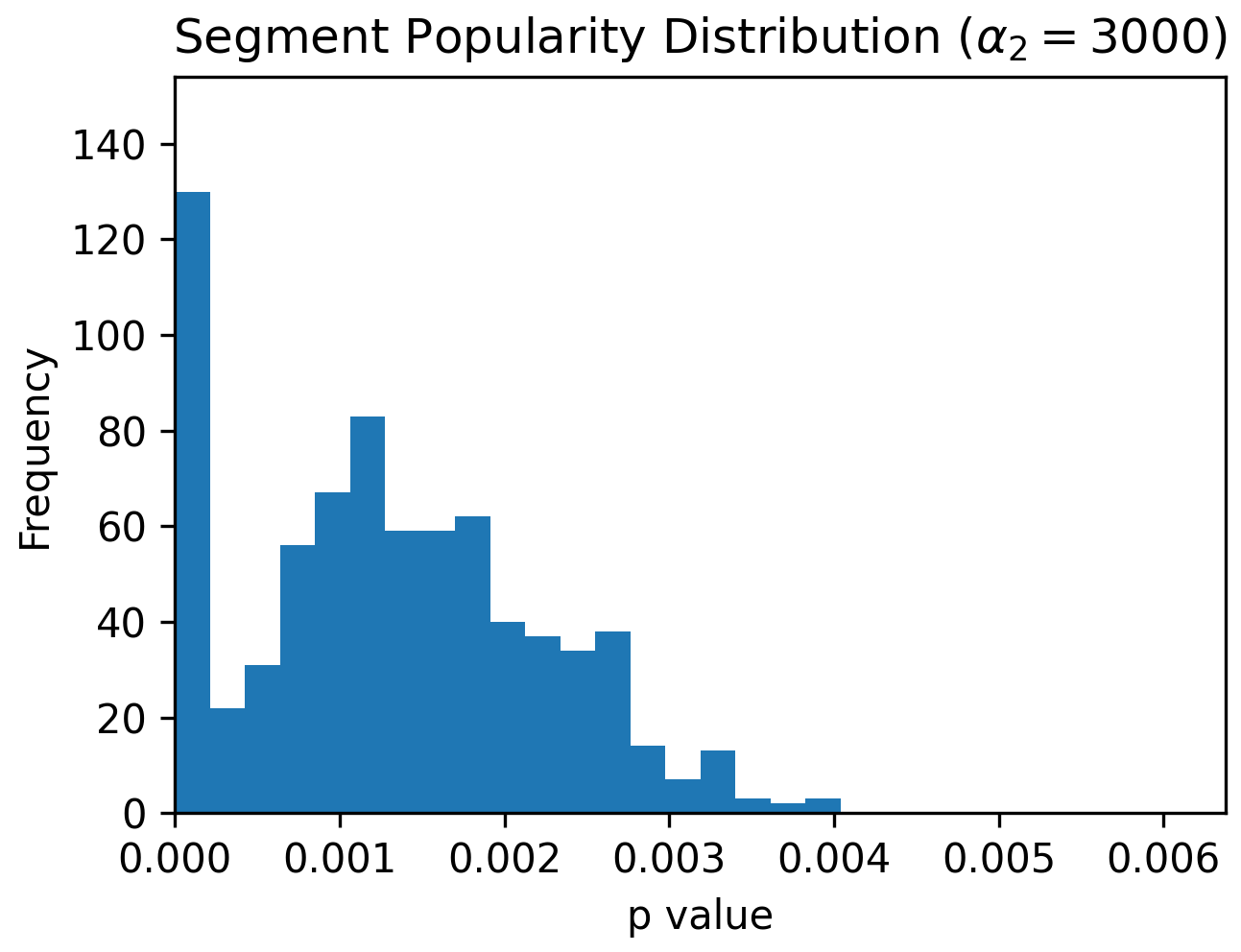}
        \caption{}\label{fig:setting2_col4}
    \end{subfigure}\hfill
    \begin{subfigure}[b]{0.15\textwidth}
        \includegraphics[width=\linewidth]{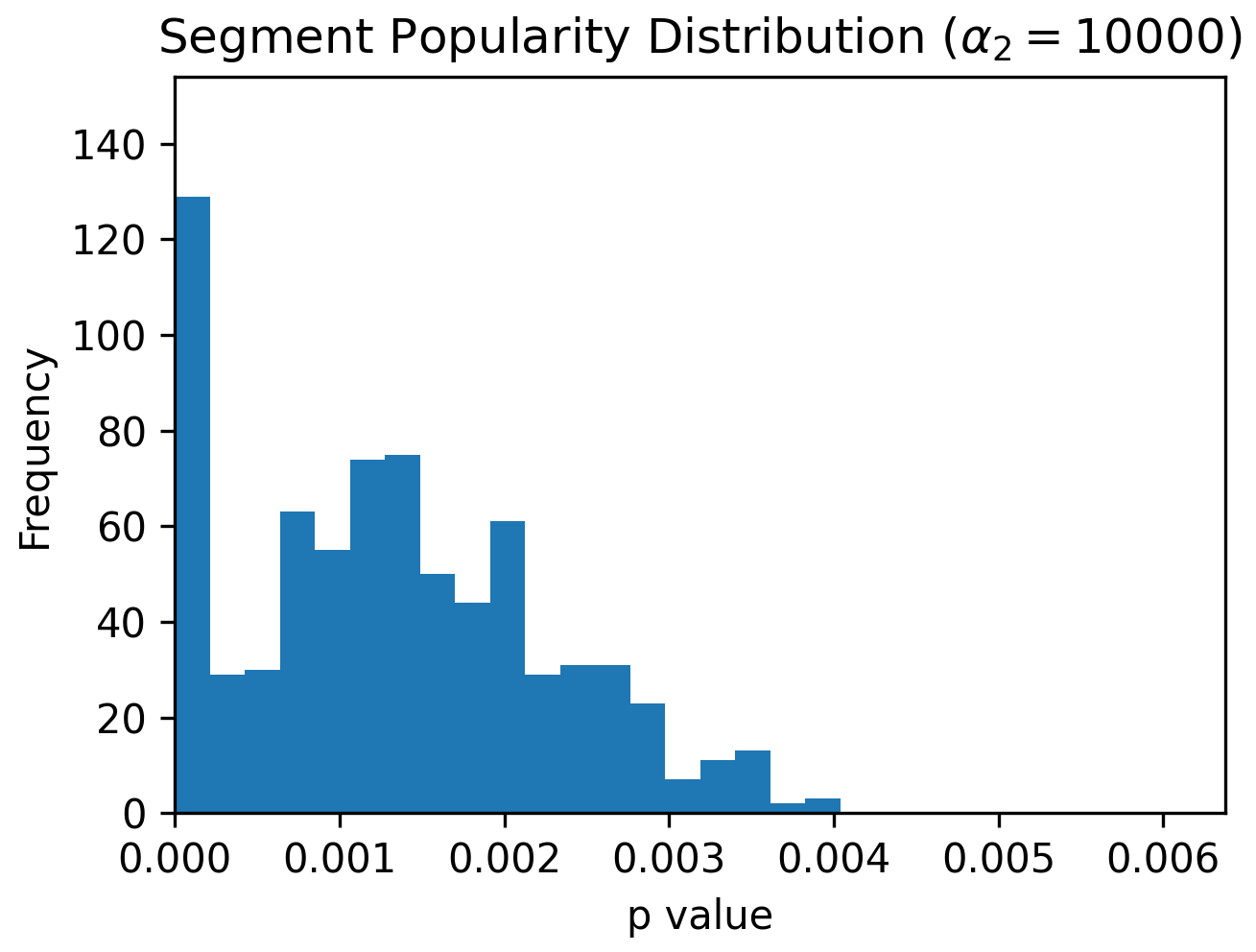}
        \caption{}\label{fig:setting2_col5}
    \end{subfigure}\hfill
    \begin{subfigure}[b]{0.15\textwidth}
        \includegraphics[width=\linewidth]{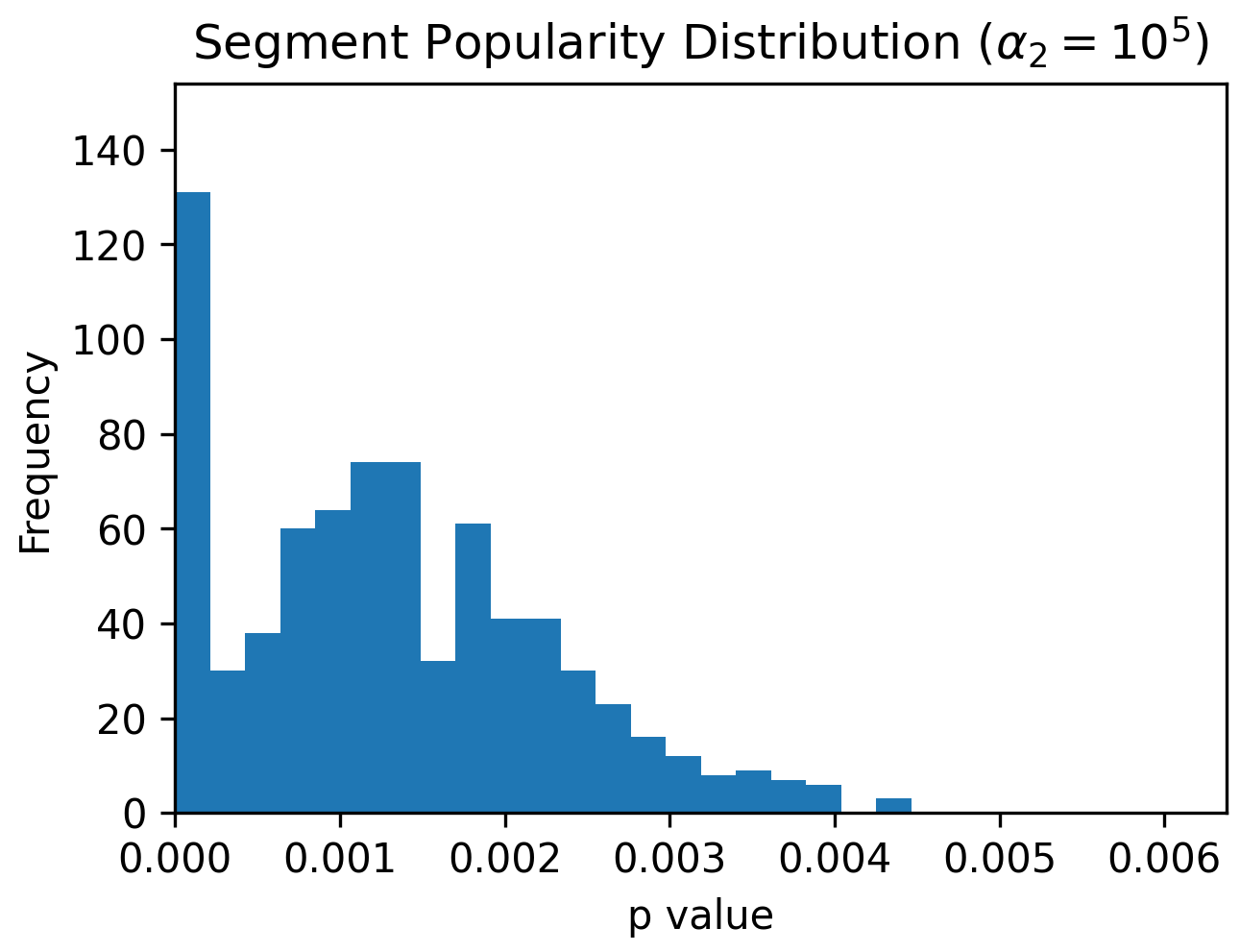}
        \caption{}\label{fig:setting2_col6}
    \end{subfigure}

    \vspace{0.3em}

    \begin{subfigure}[b]{0.15\textwidth}
        \includegraphics[width=\linewidth]{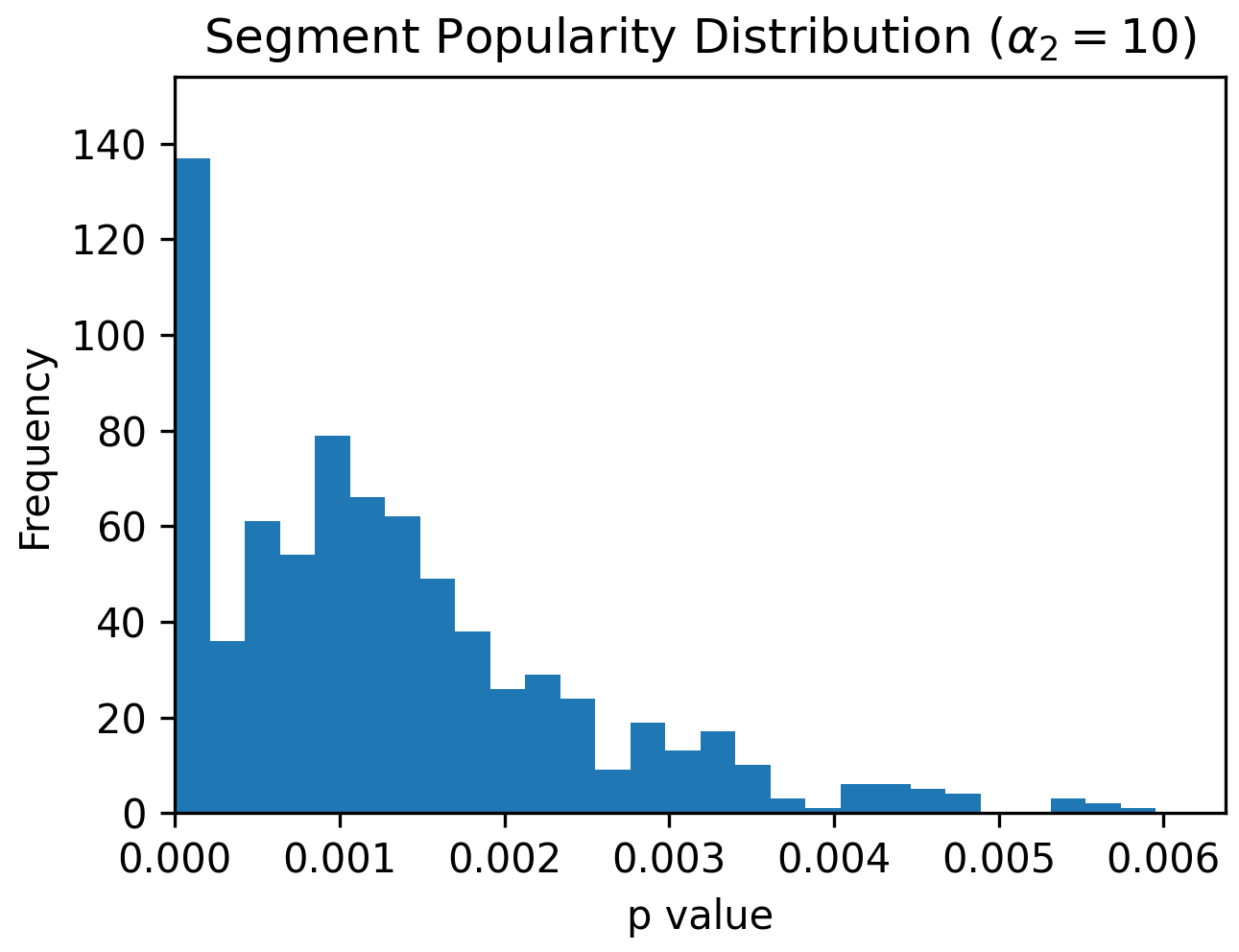}
        \caption{}\label{fig:setting3_col1}
    \end{subfigure}\hfill
    \begin{subfigure}[b]{0.15\textwidth}
        \includegraphics[width=\linewidth]{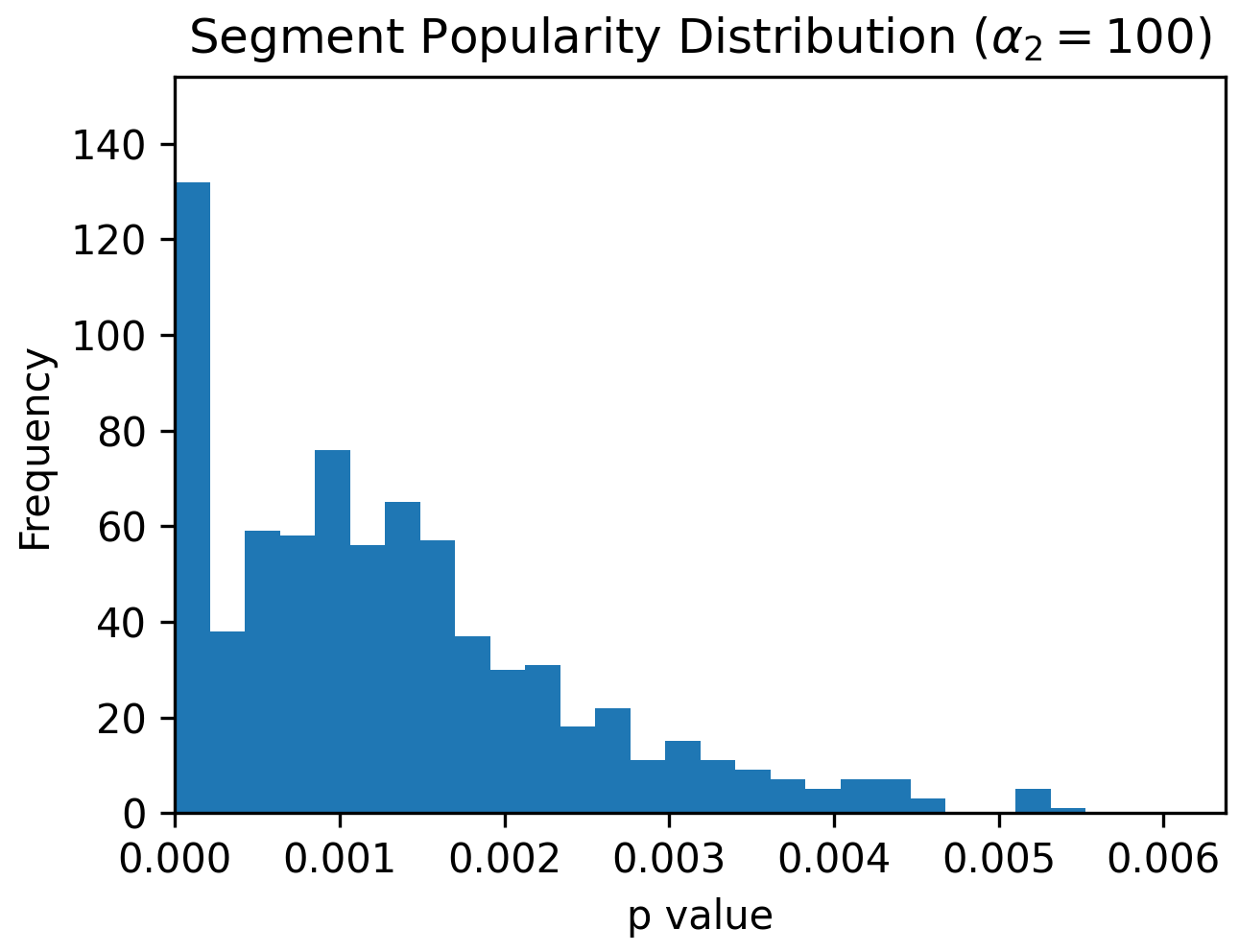}
        \caption{}\label{fig:setting3_col2}
    \end{subfigure}\hfill
    \begin{subfigure}[b]{0.15\textwidth}
        \includegraphics[width=\linewidth]{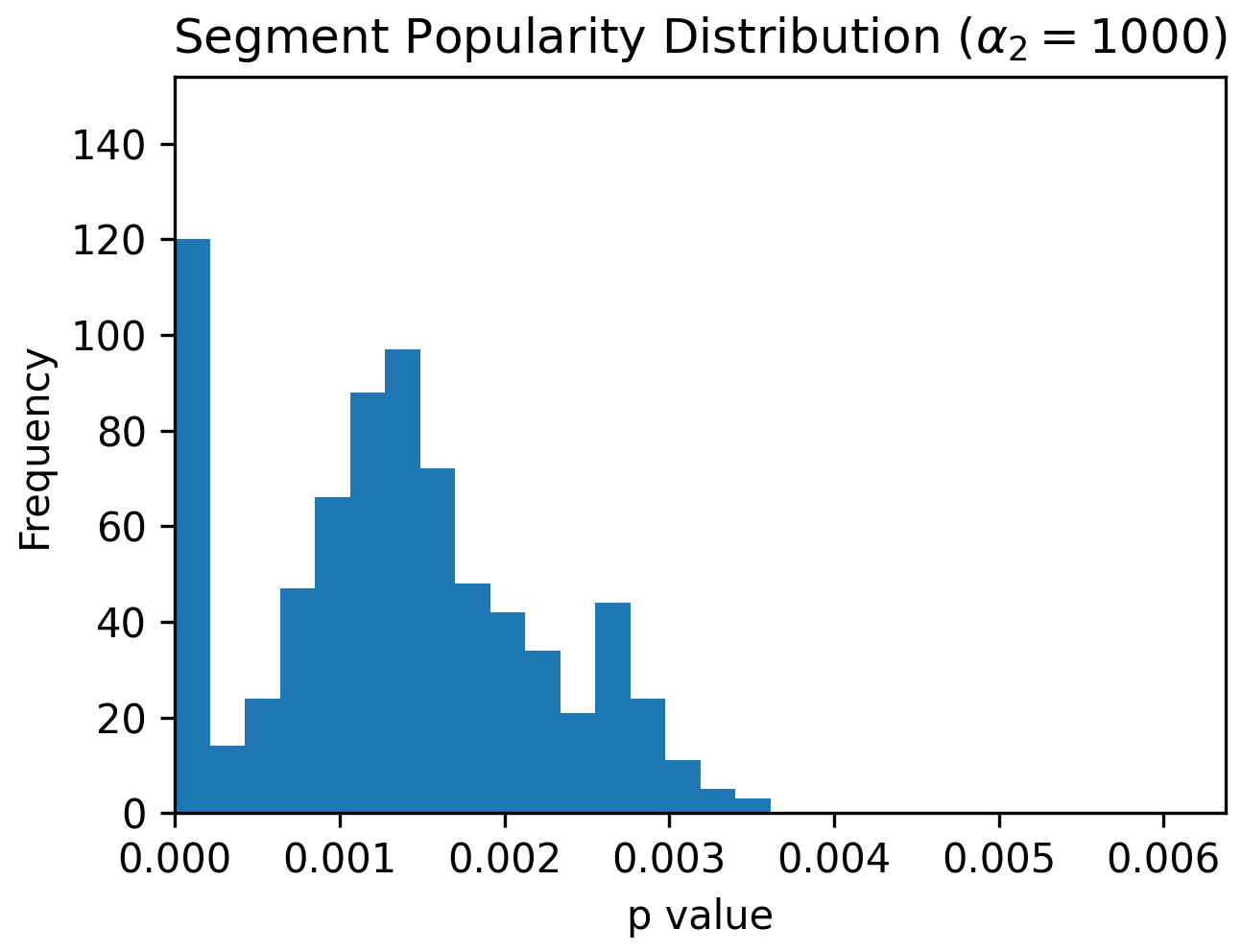}
        \caption{}\label{fig:setting3_col3}
    \end{subfigure}\hfill
    \begin{subfigure}[b]{0.15\textwidth}
        \includegraphics[width=\linewidth]{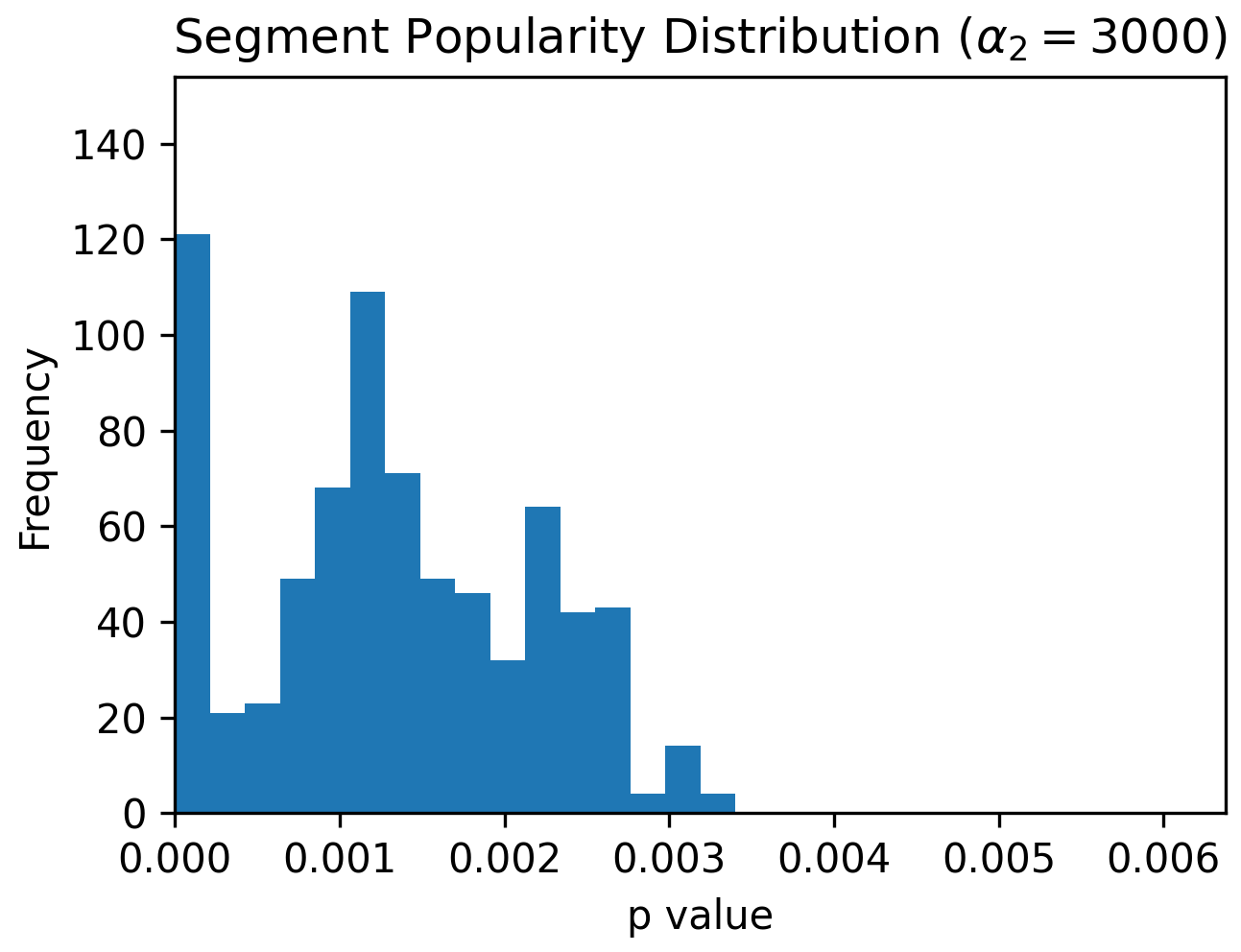}
        \caption{}\label{fig:setting3_col4}
    \end{subfigure}\hfill
    \begin{subfigure}[b]{0.15\textwidth}
        \includegraphics[width=\linewidth]{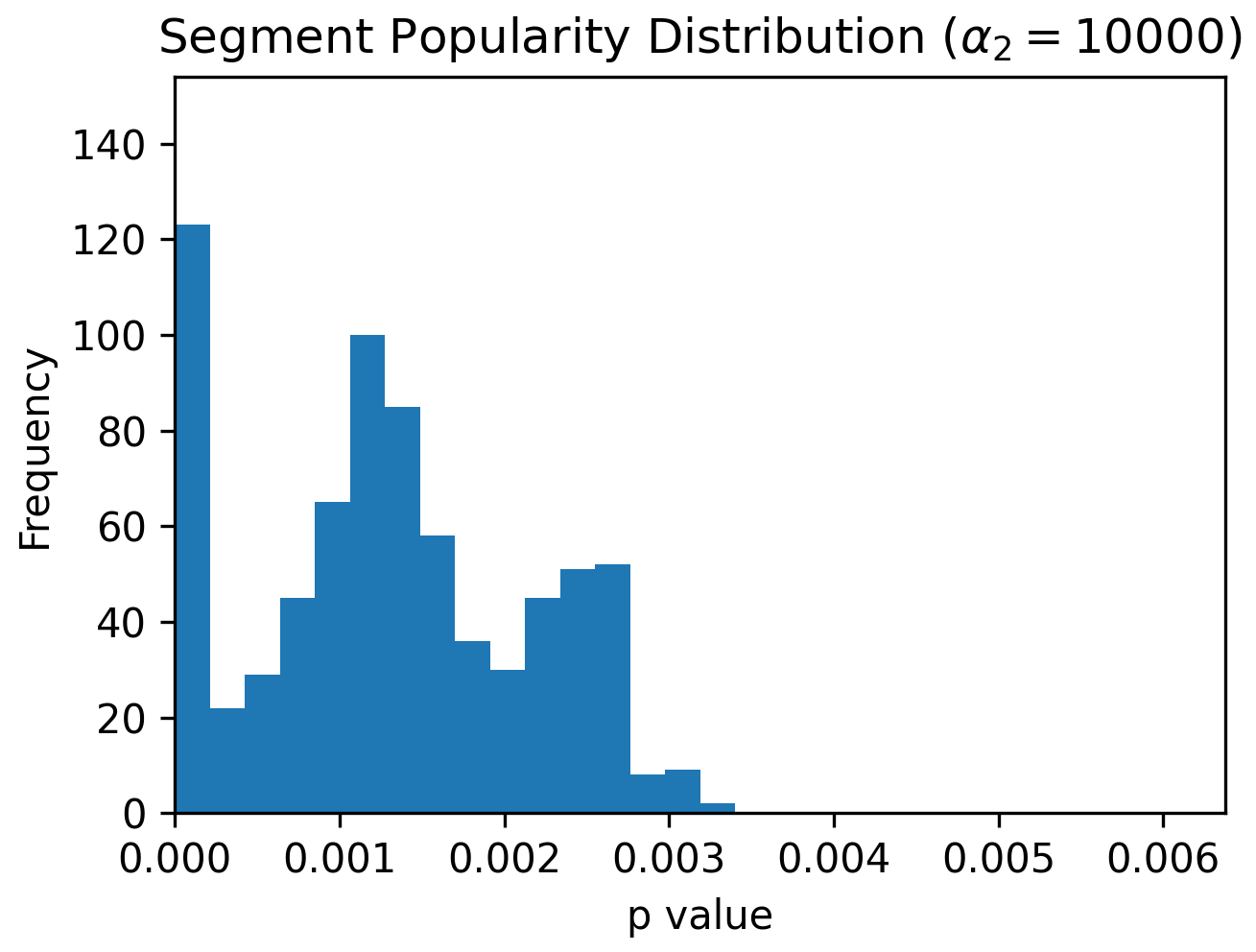}
        \caption{}\label{fig:setting3_col5}
    \end{subfigure}\hfill
    \begin{subfigure}[b]{0.15\textwidth}
        \includegraphics[width=\linewidth]{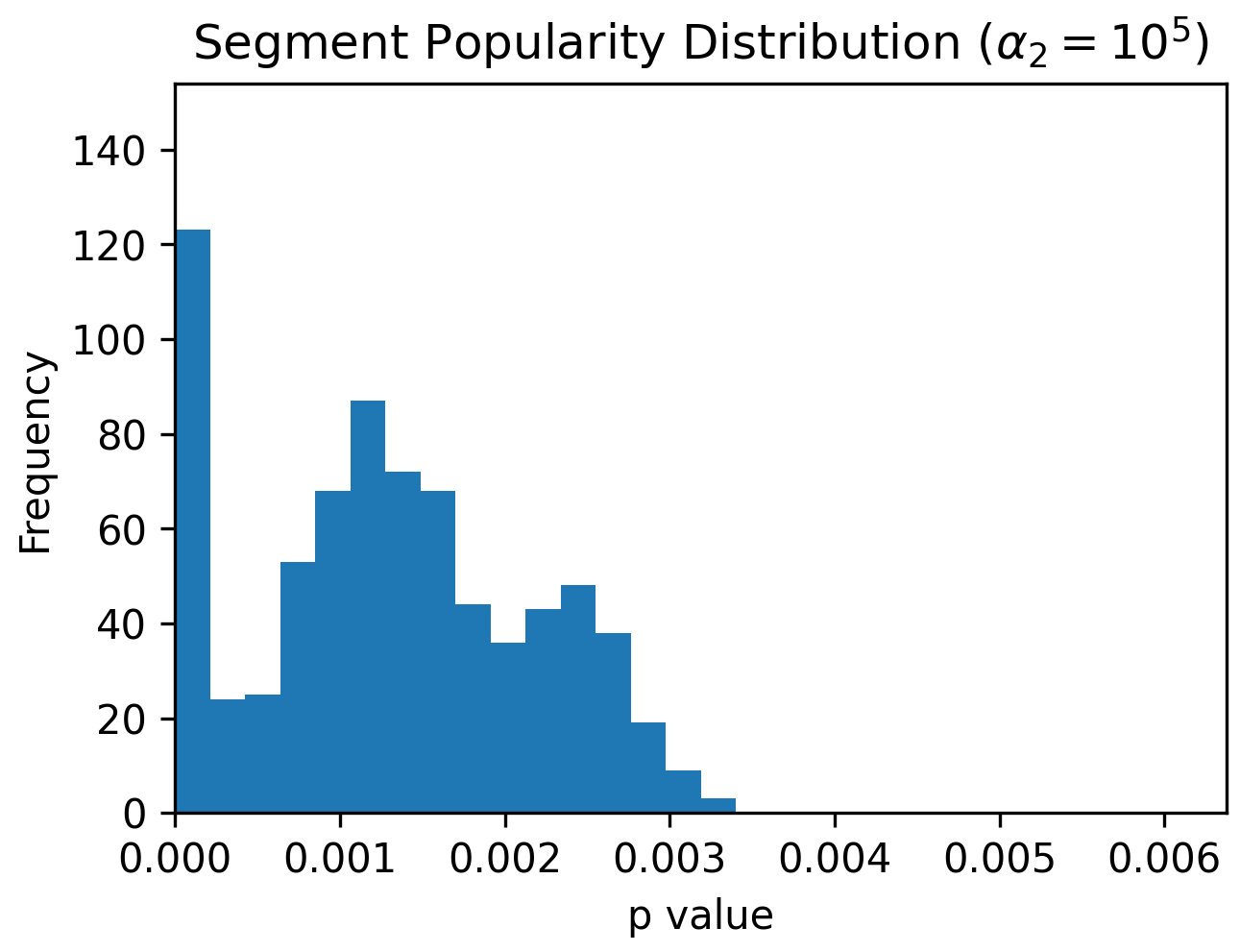}
        \caption{}\label{fig:setting3_col6}
    \end{subfigure}

    \vspace{0.4em}

    \noindent
    \begin{minipage}[t]{0.15\textwidth}\centering\footnotesize $\alpha_2 = 10$\end{minipage}\hfill
    \begin{minipage}[t]{0.15\textwidth}\centering\footnotesize $\alpha_2 = 100$\end{minipage}\hfill
    \begin{minipage}[t]{0.15\textwidth}\centering\footnotesize $\alpha_2 = 1000$\end{minipage}\hfill
    \begin{minipage}[t]{0.15\textwidth}\centering\footnotesize $\alpha_2 = 3000$\end{minipage}\hfill
    \begin{minipage}[t]{0.15\textwidth}\centering\footnotesize $\alpha_2 = 10000$\end{minipage}\hfill
    \begin{minipage}[t]{0.15\textwidth}\centering\footnotesize $\alpha_2 = 10^{5}$\end{minipage}

    \caption{Segment popularity ($p_i$) under different coverage weights across MPRs. (a)--(f): 2\% MPR; (g)--(l): 5\% MPR; (m)--(r): 10\% MPR. Columns from left to right correspond to $\alpha_2 = 10, 100, 1000, 3000, 10000, 10^5$.}

    \label{fig:Pi_3settings_grid}
\end{figure}

The explanation of this observation reveals an interesting patterns. From Table~\ref{Table:350_result} and Figure~\ref{fig:Pi_3settings_grid}, it is noticed that the average travel speed is firstly increasing and then decreasing. Meanwhile, the uniformity of segment popularity increases with $\alpha_2$ until the robotaxis use up their de-routing budget. Afterwards, increasing $\alpha_2$ values makes the algorithm choose a routing plan that have longer travel time but similar (maximum) total travel distance, in order to further improve the spatiotemporal coverage objective. This routing strategy results in reduced average travel speeds with similar segment popularity, and thus reduces the sensing power, according to Equation \ref{eqa:sensing_power}.

The decrease of the critical values of $\alpha_2$ under different MPRs also reveal increasing robotaxi fleet size makes it easier for the routing algorithm achieves higher segment popularity across the entire network, which aligns with Figure~\ref{fig:Pi_3settings_grid}.

Observations from the sensing power analysis validates the design of the proposed routing for monitoring framework. Higher weights in the monitoring objective increase spatiotemporal coverage, but not always improve the fleet's sensing power. Thus, there exists an optimal strategy to balance the spatiotemporal coverage and mobility in order to achieve the highest sensing power. This leads to a more detailed discussion on routing strategy presented in the next section.

\subsubsubsection{Routing Strategy Analysis}
\label{sec:routing strategies over MRP}

\begin{figure}[pos=tbp]
    \centering
    \captionsetup[subfigure]{skip=2pt}
    \begin{subfigure}[b]{0.16\linewidth}
        \includegraphics[width=\linewidth]{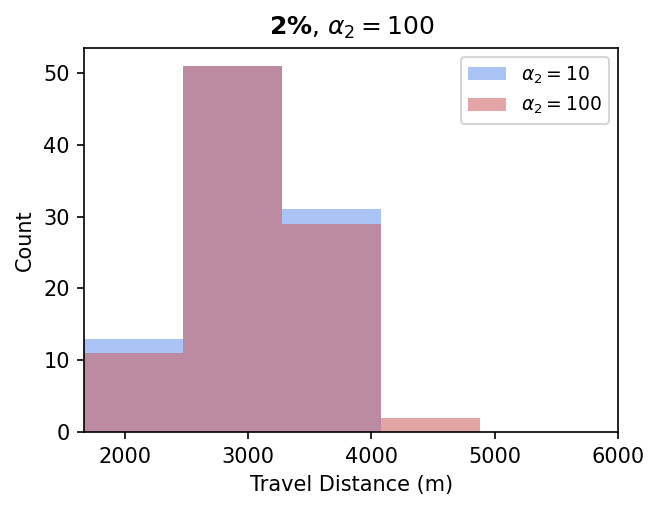}
        \caption{}\label{fig:dist_2p_a100}
    \end{subfigure}\hfill
    \begin{subfigure}[b]{0.16\linewidth}
        \includegraphics[width=\linewidth]{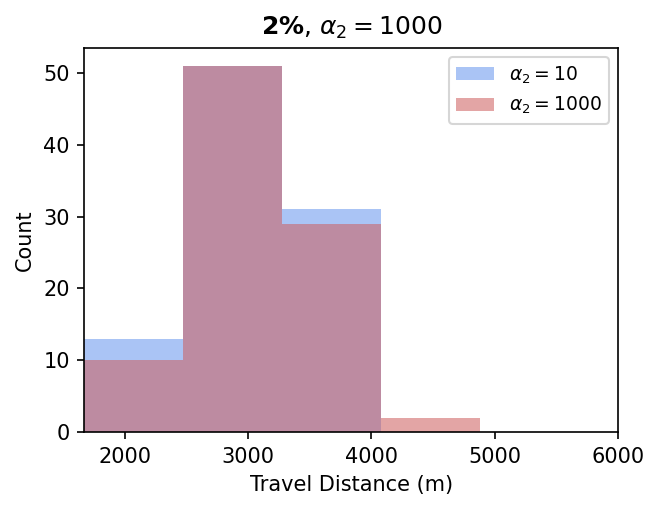}
        \caption{}\label{fig:dist_2p_a1000}
    \end{subfigure}\hfill
    \begin{subfigure}[b]{0.16\linewidth}
        \includegraphics[width=\linewidth]{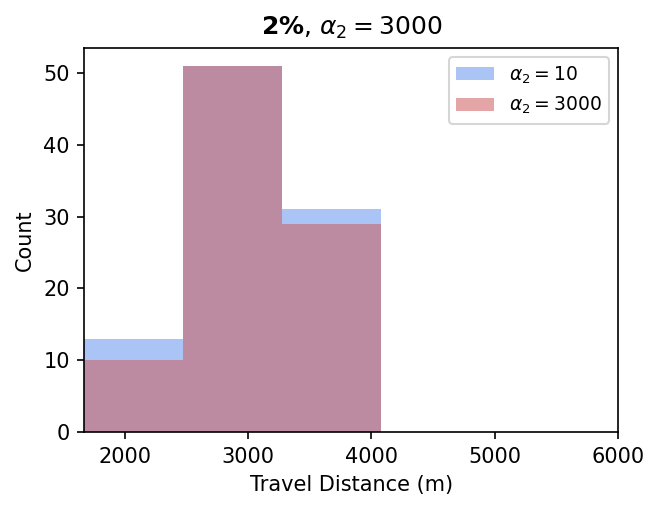}
        \caption{}\label{fig:dist_2p_a3000}
    \end{subfigure}\hfill
    \begin{subfigure}[b]{0.16\linewidth}
        \includegraphics[width=\linewidth]{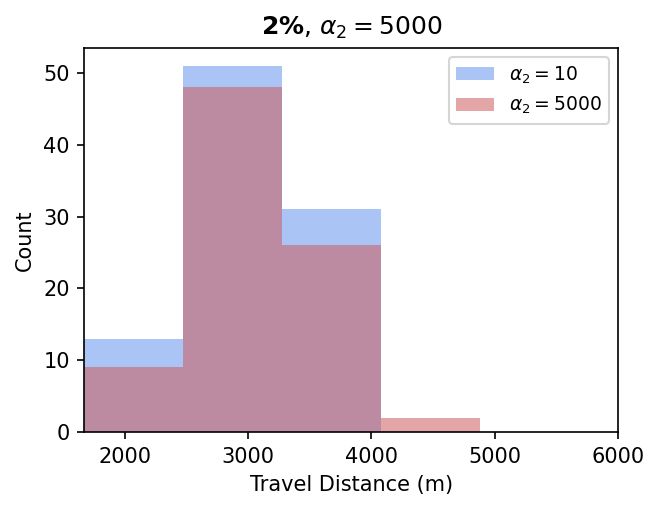}
        \caption{}\label{fig:dist_2p_a5000}
    \end{subfigure}\hfill
    \begin{subfigure}[b]{0.16\linewidth}
        \includegraphics[width=\linewidth]{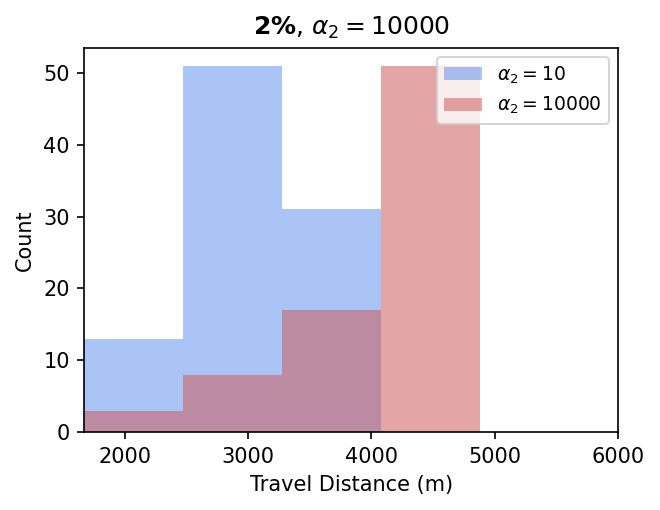}
        \caption{}\label{fig:dist_2p_a10000}
    \end{subfigure}\hfill
    \begin{subfigure}[b]{0.16\linewidth}
        \includegraphics[width=\linewidth]{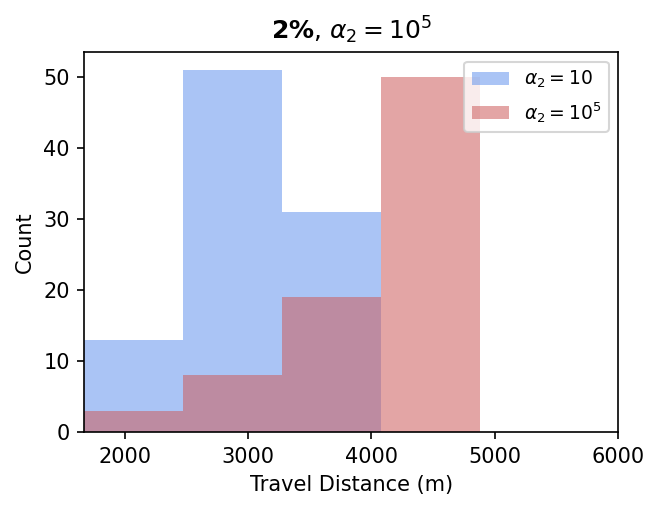}
        \caption{}\label{fig:dist_2p_a1e5}
    \end{subfigure}

    \vspace{0.3em}
    \begin{subfigure}[b]{0.16\linewidth}
        \includegraphics[width=\linewidth]{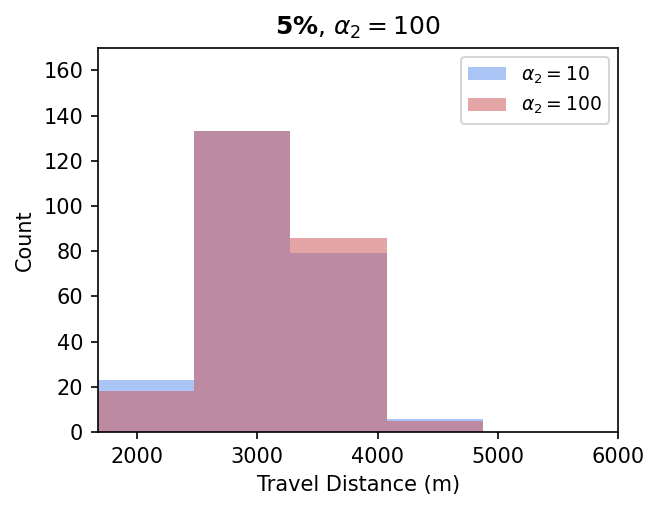}
        \caption{}\label{fig:dist_5p_a100}
    \end{subfigure}\hfill
    \begin{subfigure}[b]{0.16\linewidth}
        \includegraphics[width=\linewidth]{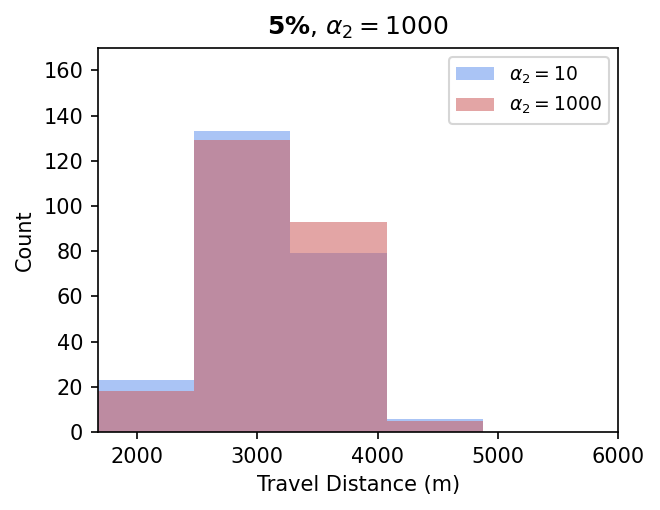}
        \caption{}\label{fig:dist_5p_a1000}
    \end{subfigure}\hfill
    \begin{subfigure}[b]{0.16\linewidth}
        \includegraphics[width=\linewidth]{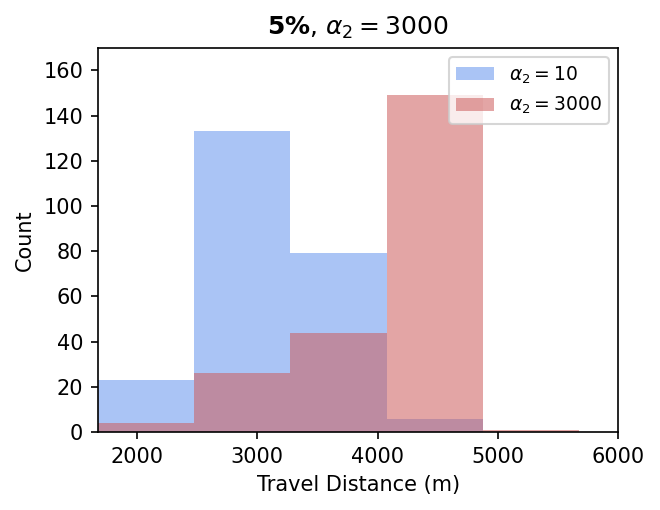}
        \caption{}\label{fig:dist_5p_a3000}
    \end{subfigure}\hfill
    \begin{subfigure}[b]{0.16\linewidth}
        \includegraphics[width=\linewidth]{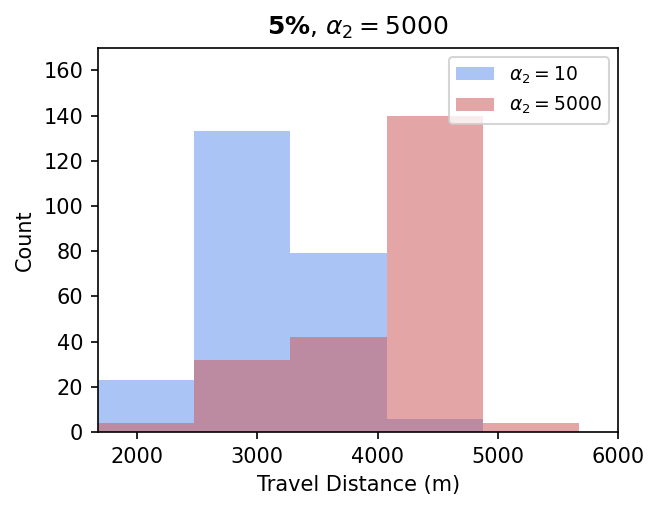}
        \caption{}\label{fig:dist_5p_a5000}
    \end{subfigure}\hfill
    \begin{subfigure}[b]{0.16\linewidth}
        \includegraphics[width=\linewidth]{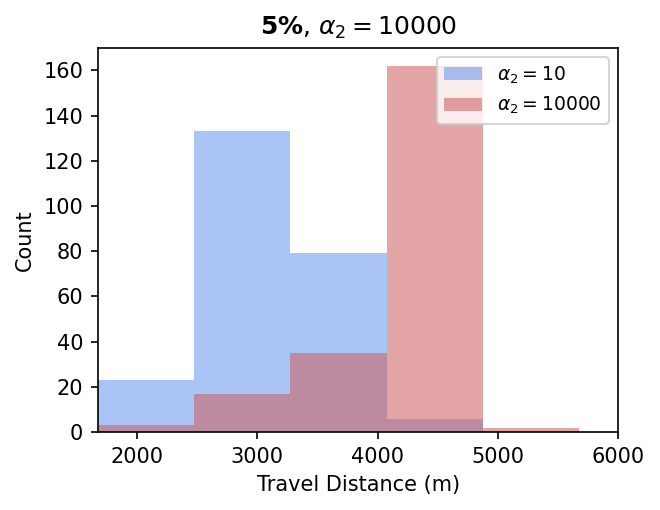}
        \caption{}\label{fig:dist_5p_a10000}
    \end{subfigure}\hfill
    \begin{subfigure}[b]{0.16\linewidth}
        \includegraphics[width=\linewidth]{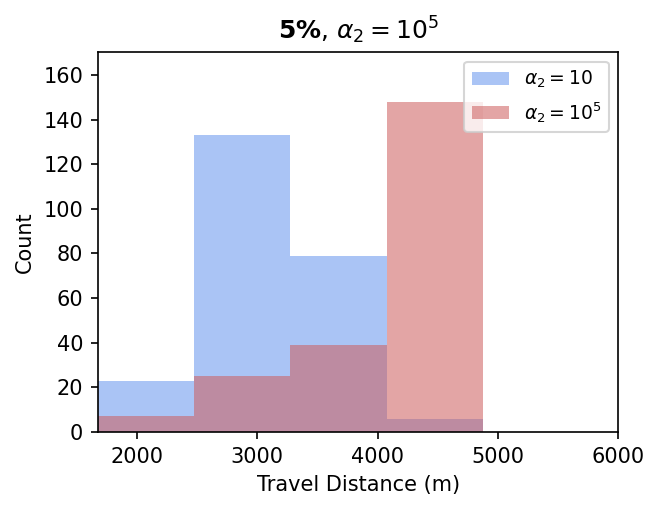}
        \caption{}\label{fig:dist_5p_a1e5}
    \end{subfigure}

    \vspace{0.3em}
    \begin{subfigure}[b]{0.16\linewidth}
        \includegraphics[width=\linewidth]{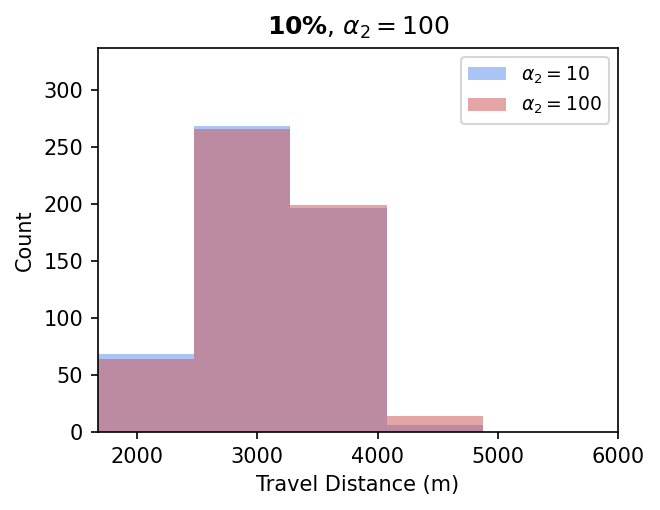}
        \caption{}\label{fig:dist_10p_a100}
    \end{subfigure}\hfill
    \begin{subfigure}[b]{0.16\linewidth}
        \includegraphics[width=\linewidth]{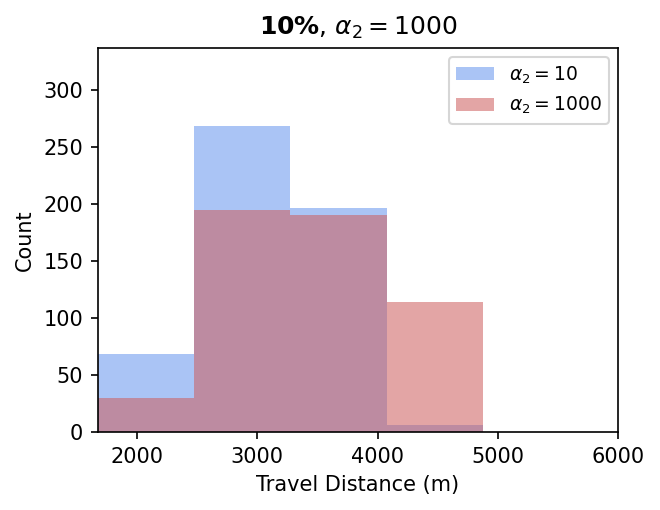}
        \caption{}\label{fig:dist_10p_a1000}
    \end{subfigure}\hfill
    \begin{subfigure}[b]{0.16\linewidth}
        \includegraphics[width=\linewidth]{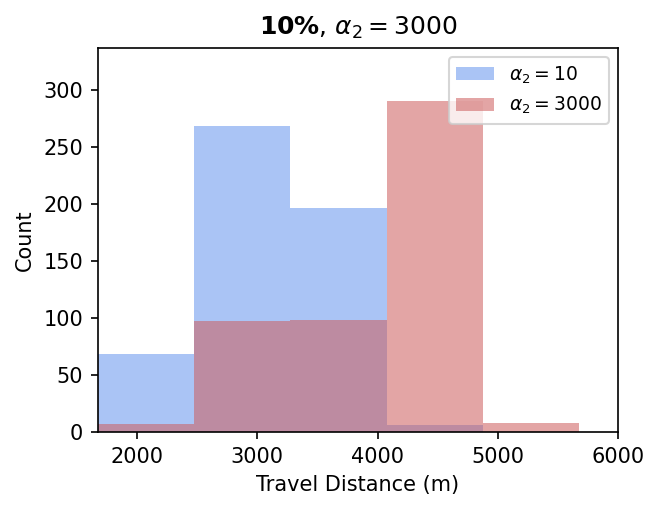}
        \caption{}\label{fig:dist_10p_a3000}
    \end{subfigure}\hfill
    \begin{subfigure}[b]{0.16\linewidth}
        \includegraphics[width=\linewidth]{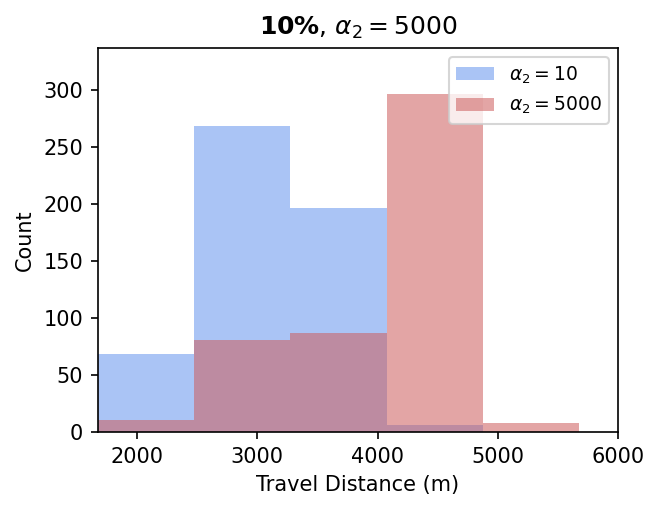}
        \caption{}\label{fig:dist_10p_a5000}
    \end{subfigure}\hfill
    \begin{subfigure}[b]{0.16\linewidth}
        \includegraphics[width=\linewidth]{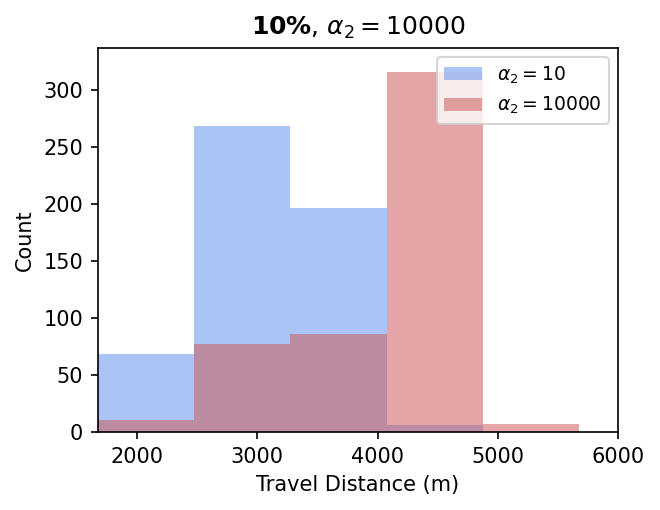}
        \caption{}\label{fig:dist_10p_a10000}
    \end{subfigure}\hfill
    \begin{subfigure}[b]{0.16\linewidth}
        \includegraphics[width=\linewidth]{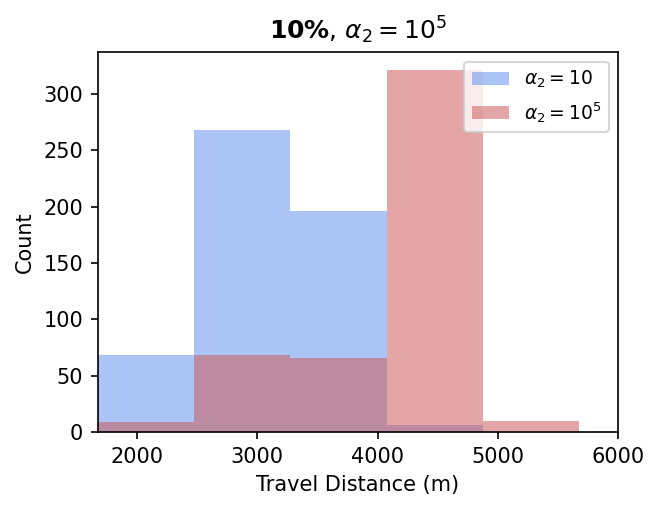}
        \caption{}\label{fig:dist_10p_a1e5}
    \end{subfigure}

    \vspace{0.3em}
    \begin{subfigure}[b]{0.16\linewidth}
        \includegraphics[width=\linewidth]{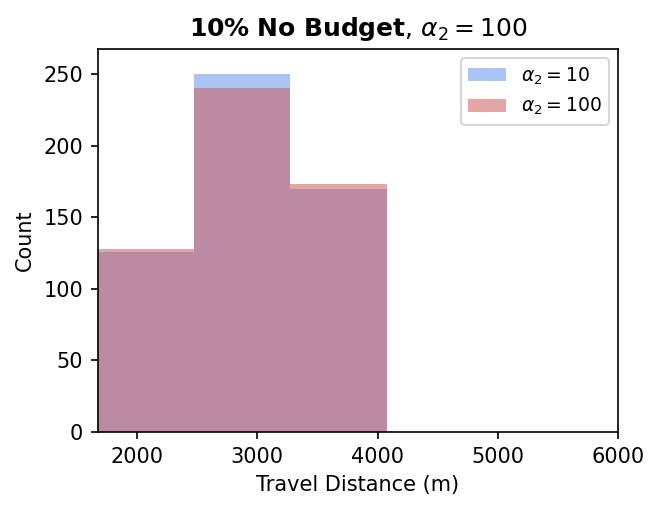}
        \caption{}\label{fig:dist_10p_nobgt_a100}
    \end{subfigure}\hfill
    \begin{subfigure}[b]{0.16\linewidth}
        \includegraphics[width=\linewidth]{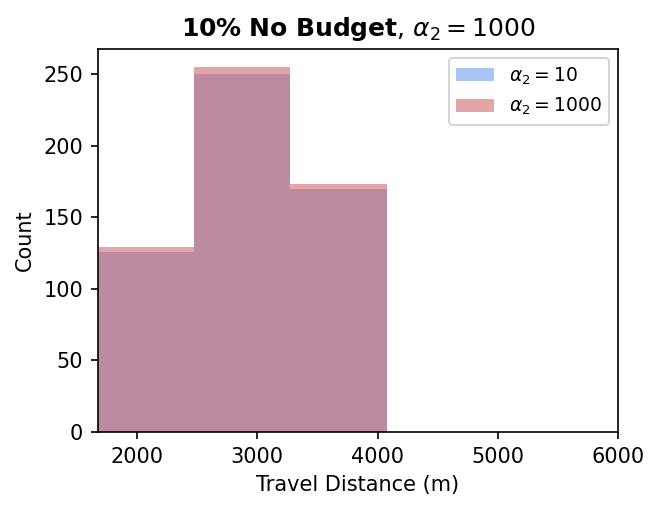}
        \caption{}\label{fig:dist_10p_nobgt_a1000}
    \end{subfigure}\hfill
    \begin{subfigure}[b]{0.16\linewidth}
        \includegraphics[width=\linewidth]{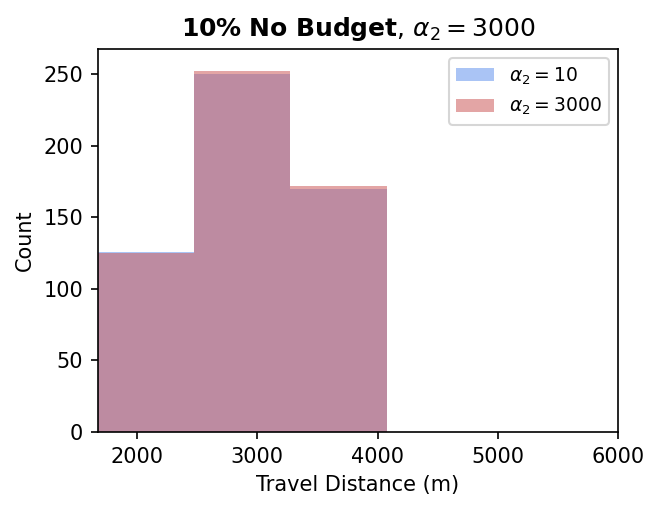}
        \caption{}\label{fig:dist_10p_nobgt_a3000}
    \end{subfigure}\hfill
    \begin{subfigure}[b]{0.16\linewidth}
        \includegraphics[width=\linewidth]{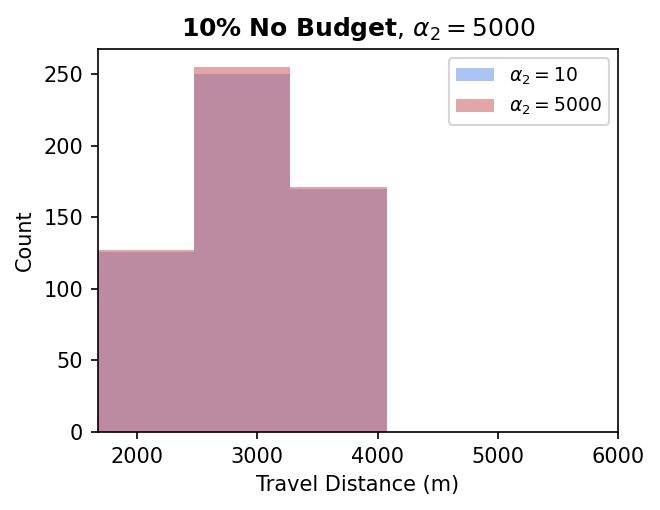}
        \caption{}\label{fig:dist_10p_nobgt_a5000}
    \end{subfigure}\hfill
    \begin{subfigure}[b]{0.16\linewidth}
        \includegraphics[width=\linewidth]{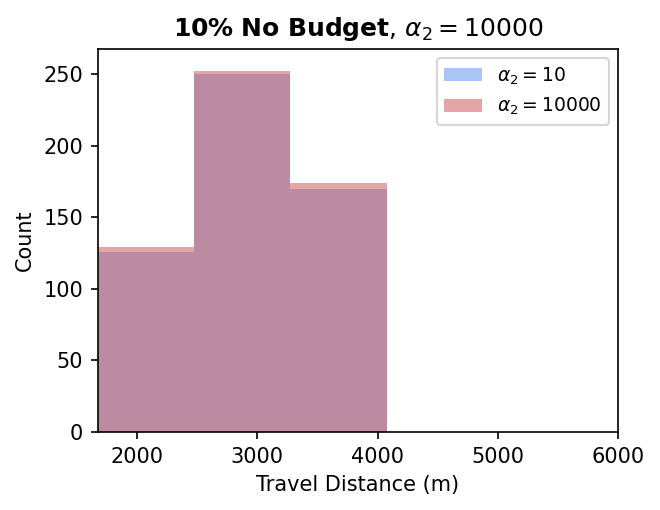}
        \caption{}\label{fig:dist_10p_nobgt_a10000}
    \end{subfigure}\hfill
    \begin{subfigure}[b]{0.16\linewidth}
        \includegraphics[width=\linewidth]{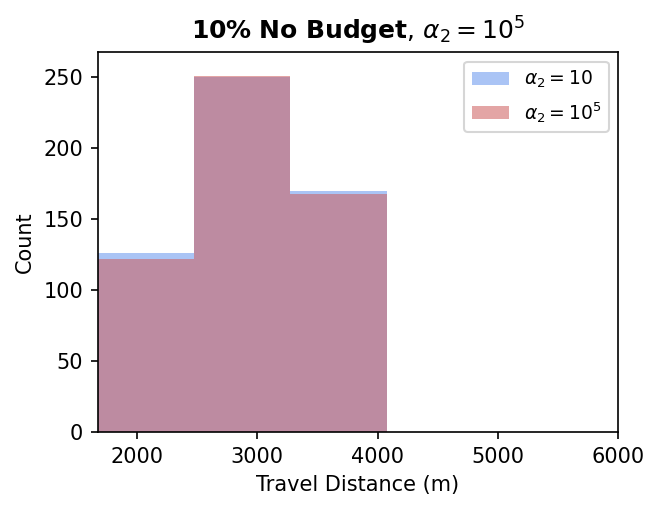}
        \caption{}\label{fig:dist_10p_nobgt_a1e5}
    \end{subfigure}

    \vspace{0.4em}
    \noindent
    \begin{minipage}[t]{0.16\textwidth}\centering\footnotesize $\alpha_2 = 100$\end{minipage}\hfill
    \begin{minipage}[t]{0.16\textwidth}\centering\footnotesize $\alpha_2 = 1000$\end{minipage}\hfill
    \begin{minipage}[t]{0.16\textwidth}\centering\footnotesize $\alpha_2 = 3000$\end{minipage}\hfill
    \begin{minipage}[t]{0.16\textwidth}\centering\footnotesize $\alpha_2 = 5000$\end{minipage}\hfill
    \begin{minipage}[t]{0.16\textwidth}\centering\footnotesize $\alpha_2 = 10000$\end{minipage}\hfill
    \begin{minipage}[t]{0.16\textwidth}\centering\footnotesize $\alpha_2 = 10^{5}$\end{minipage}

    \caption{Robotaxi travel distance distributions under different coverage weights across MPRs. Each subplot compares the distribution at a given $\alpha_2$ (red) against the most travel-time-oriented case $\alpha_2=10$ (blue). (a)--(f): 2\% MPR; (g)--(l): 5\% MPR; (m)--(r): 10\% MPR; (s)--(x): 10\% MPR without de-routing budget. Columns from left to right correspond to $\alpha_2 = 100, 1000, 3000, 5000, 10000, 10^5$.}
    \label{fig:distance_grid}
\end{figure}

\begin{figure}[pos=tbp]
    \centering
    \captionsetup[subfigure]{skip=2pt}
    \begin{subfigure}[b]{0.16\linewidth}
        \includegraphics[width=\linewidth]{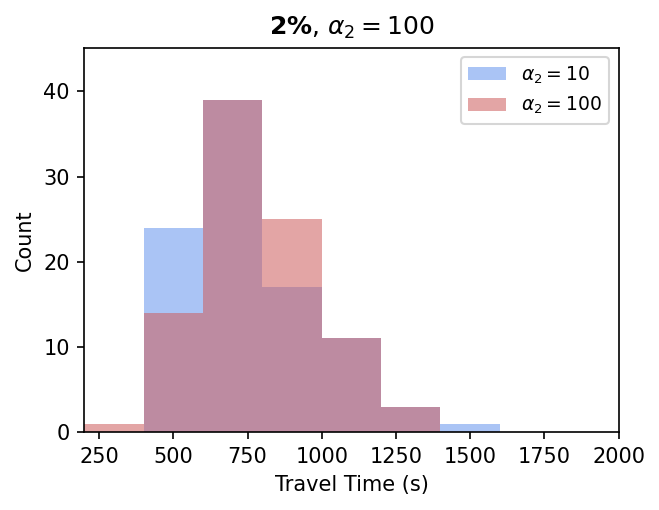}
        \caption{}\label{fig:time_2p_a100}
    \end{subfigure}\hfill
    \begin{subfigure}[b]{0.16\linewidth}
        \includegraphics[width=\linewidth]{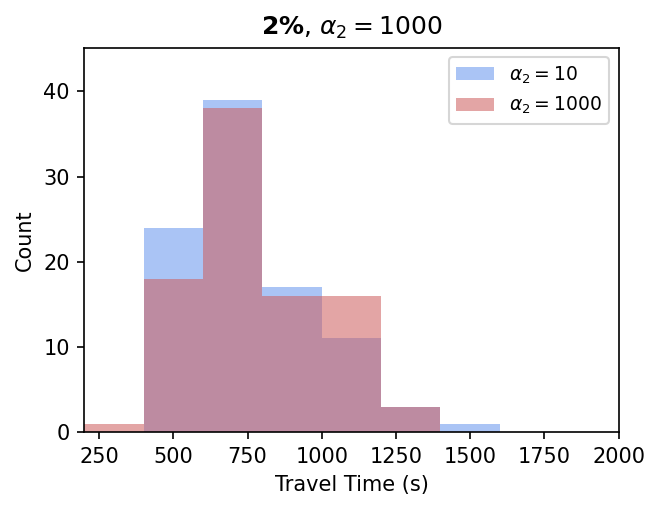}
        \caption{}\label{fig:time_2p_a1000}
    \end{subfigure}\hfill
    \begin{subfigure}[b]{0.16\linewidth}
        \includegraphics[width=\linewidth]{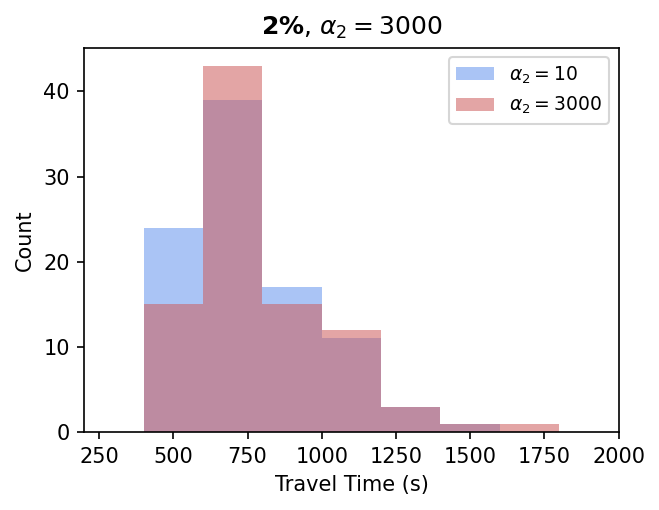}
        \caption{}\label{fig:time_2p_a3000}
    \end{subfigure}\hfill
    \begin{subfigure}[b]{0.16\linewidth}
        \includegraphics[width=\linewidth]{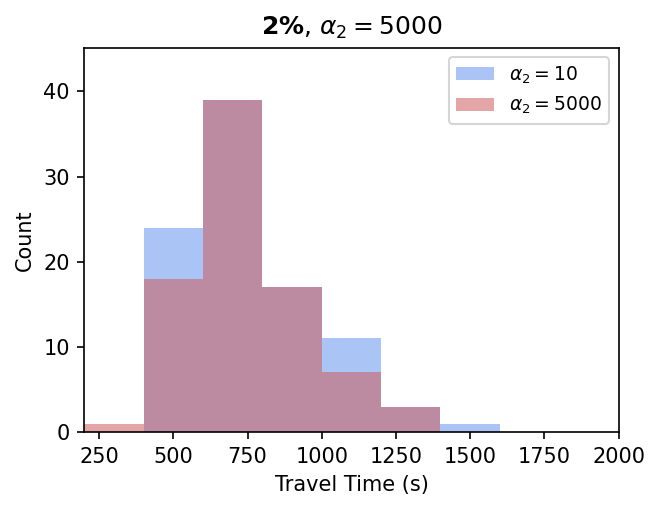}
        \caption{}\label{fig:time_2p_a5000}
    \end{subfigure}\hfill
    \begin{subfigure}[b]{0.16\linewidth}
        \includegraphics[width=\linewidth]{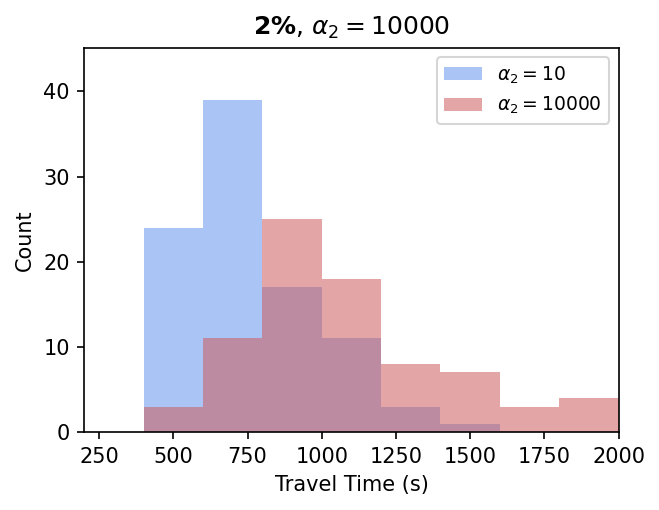}
        \caption{}\label{fig:time_2p_a10000}
    \end{subfigure}\hfill
    \begin{subfigure}[b]{0.16\linewidth}
        \includegraphics[width=\linewidth]{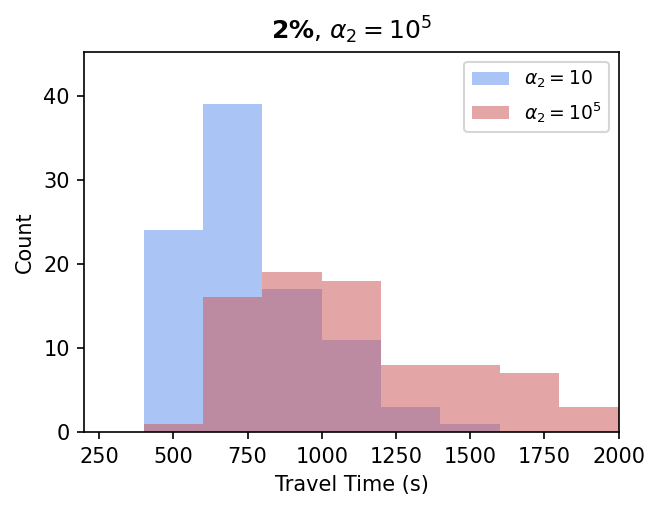}
        \caption{}\label{fig:time_2p_a1e5}
    \end{subfigure}

    \vspace{0.3em}
    \begin{subfigure}[b]{0.16\linewidth}
        \includegraphics[width=\linewidth]{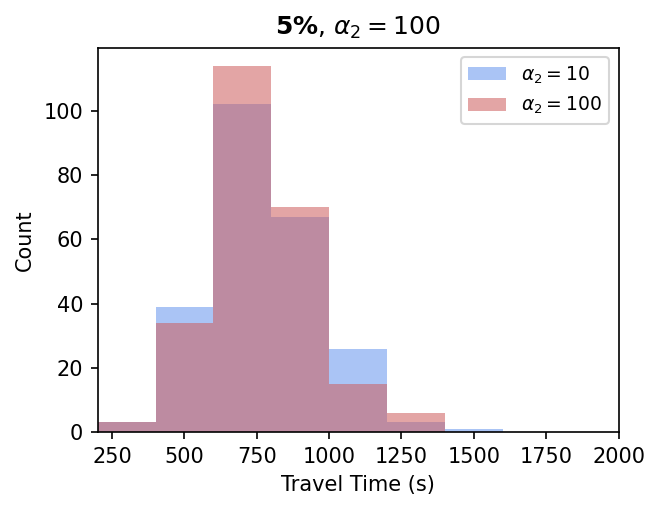}
        \caption{}\label{fig:time_5p_a100}
    \end{subfigure}\hfill
    \begin{subfigure}[b]{0.16\linewidth}
        \includegraphics[width=\linewidth]{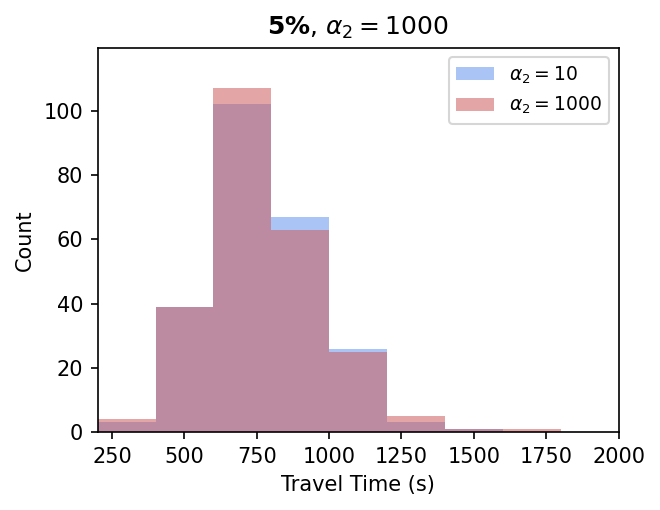}
        \caption{}\label{fig:time_5p_a1000}
    \end{subfigure}\hfill
    \begin{subfigure}[b]{0.16\linewidth}
        \includegraphics[width=\linewidth]{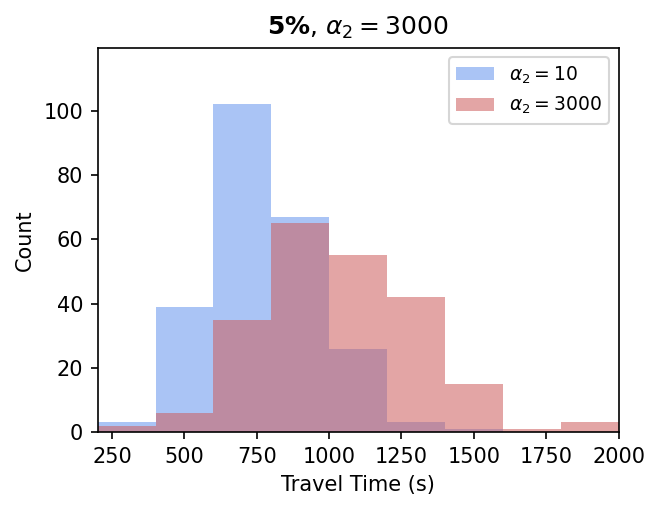}
        \caption{}\label{fig:time_5p_a3000}
    \end{subfigure}\hfill
    \begin{subfigure}[b]{0.16\linewidth}
        \includegraphics[width=\linewidth]{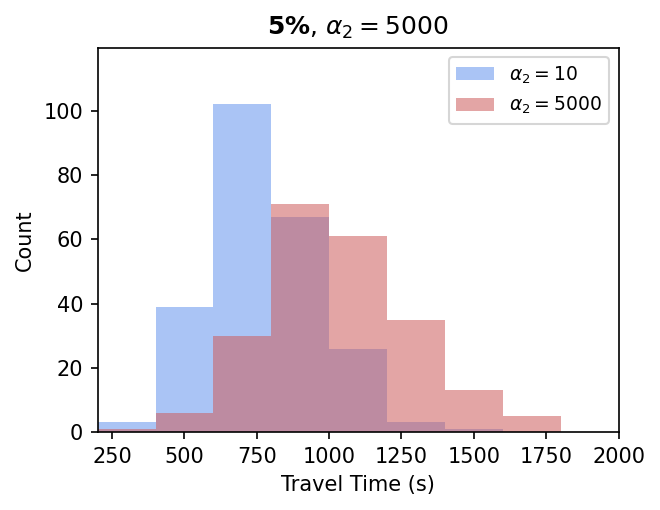}
        \caption{}\label{fig:time_5p_a5000}
    \end{subfigure}\hfill
    \begin{subfigure}[b]{0.16\linewidth}
        \includegraphics[width=\linewidth]{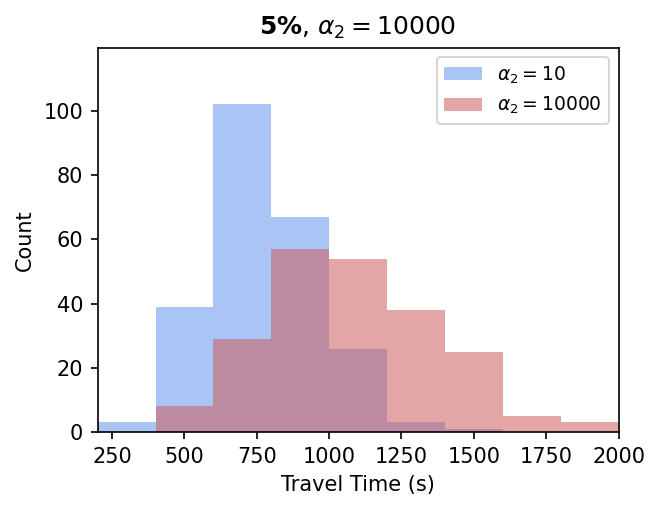}
        \caption{}\label{fig:time_5p_a10000}
    \end{subfigure}\hfill
    \begin{subfigure}[b]{0.16\linewidth}
        \includegraphics[width=\linewidth]{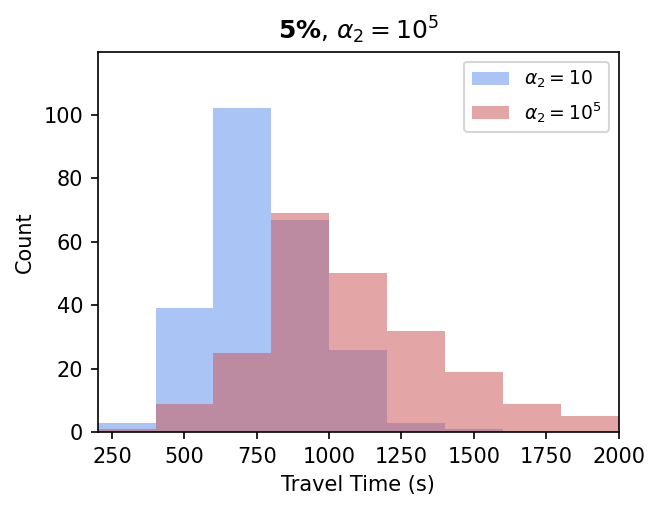}
        \caption{}\label{fig:time_5p_a1e5}
    \end{subfigure}

    \vspace{0.3em}
    \begin{subfigure}[b]{0.16\linewidth}
        \includegraphics[width=\linewidth]{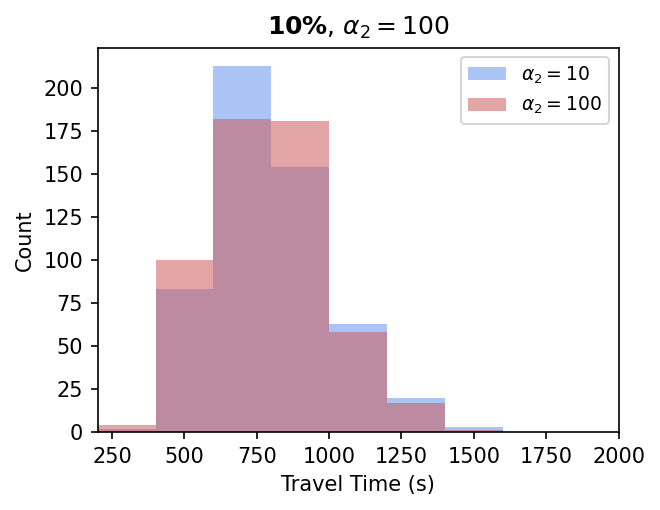}
        \caption{}\label{fig:time_10p_a100}
    \end{subfigure}\hfill
    \begin{subfigure}[b]{0.16\linewidth}
        \includegraphics[width=\linewidth]{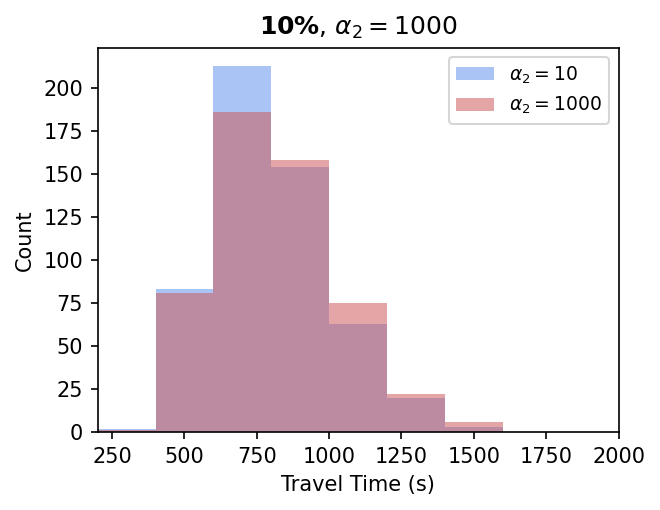}
        \caption{}\label{fig:time_10p_a1000}
    \end{subfigure}\hfill
    \begin{subfigure}[b]{0.16\linewidth}
        \includegraphics[width=\linewidth]{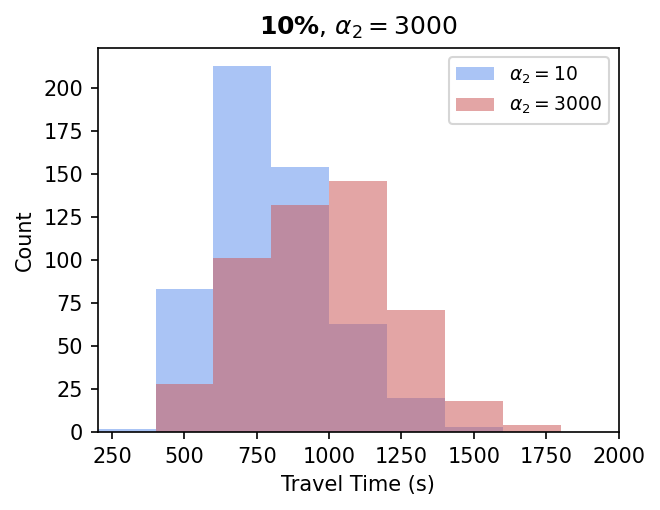}
        \caption{}\label{fig:time_10p_a3000}
    \end{subfigure}\hfill
    \begin{subfigure}[b]{0.16\linewidth}
        \includegraphics[width=\linewidth]{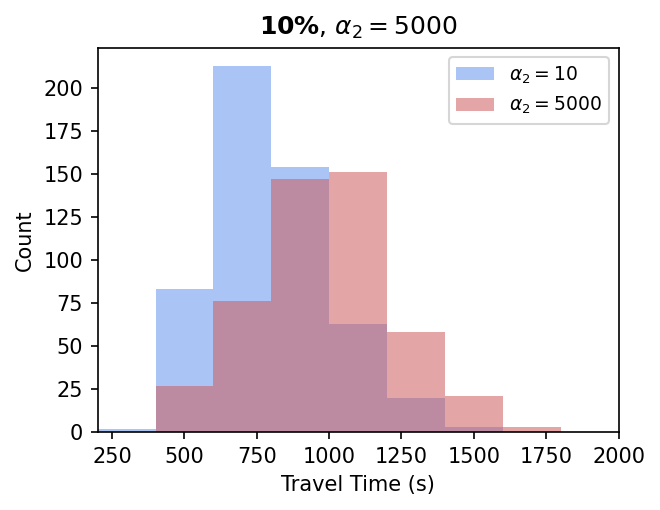}
        \caption{}\label{fig:time_10p_a5000}
    \end{subfigure}\hfill
    \begin{subfigure}[b]{0.16\linewidth}
        \includegraphics[width=\linewidth]{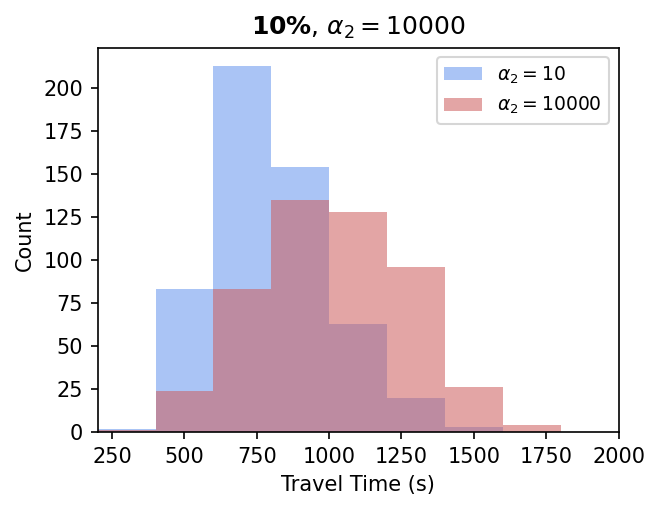}
        \caption{}\label{fig:time_10p_a10000}
    \end{subfigure}\hfill
    \begin{subfigure}[b]{0.16\linewidth}
        \includegraphics[width=\linewidth]{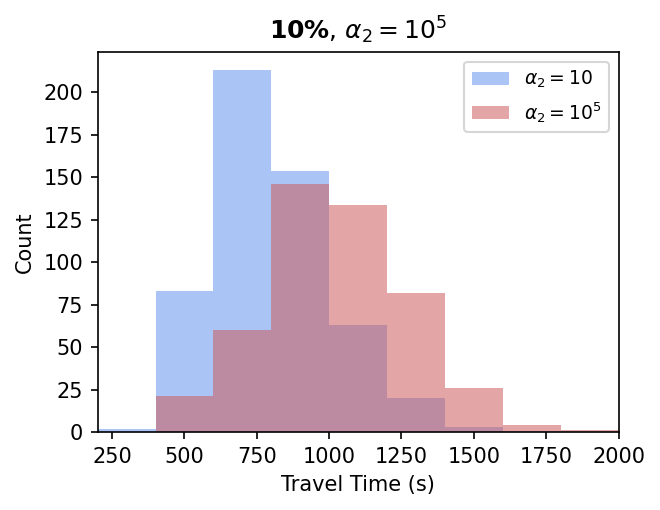}
        \caption{}\label{fig:time_10p_a1e5}
    \end{subfigure}

    \vspace{0.3em}
    \begin{subfigure}[b]{0.16\linewidth}
    \includegraphics[width=\linewidth]{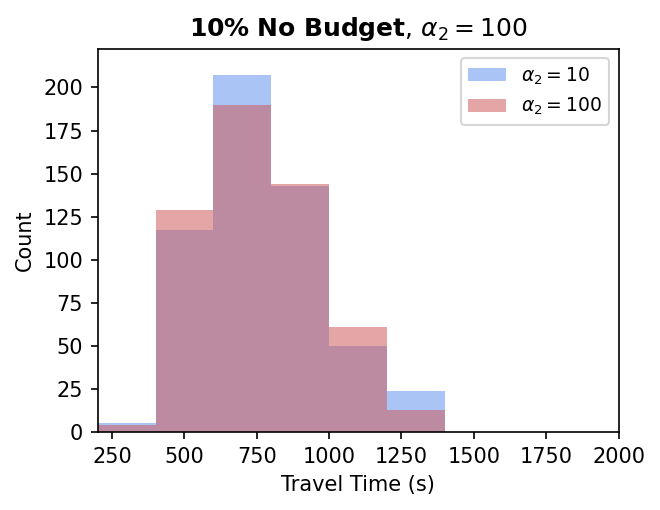}
        \caption{}\label{fig:time_10p_nobgt_a100}
    \end{subfigure}\hfill
    \begin{subfigure}[b]{0.16\linewidth}
    \includegraphics[width=\linewidth]{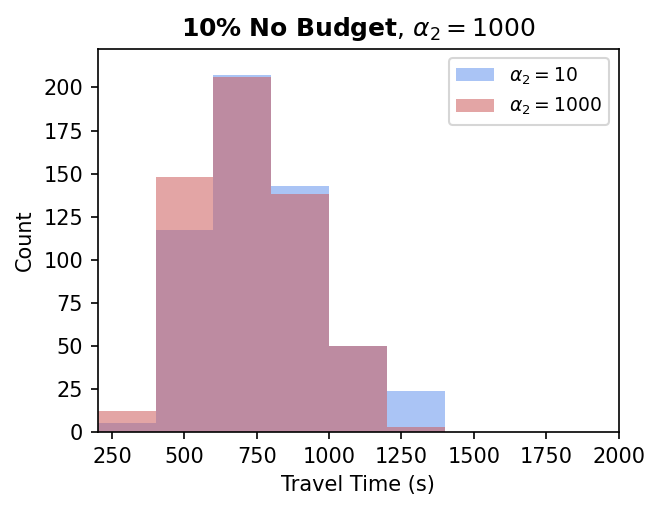}
        \caption{}\label{fig:time_10p_nobgt_a1000}
    \end{subfigure}\hfill
    \begin{subfigure}[b]{0.16\linewidth}
    \includegraphics[width=\linewidth]{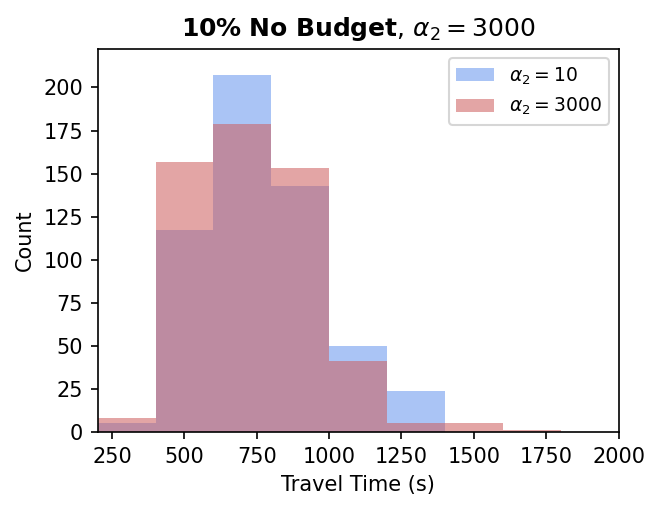}
        \caption{}\label{fig:time_10p_nobgt_a3000}
    \end{subfigure}\hfill
    \begin{subfigure}[b]{0.16\linewidth}
    \includegraphics[width=\linewidth]{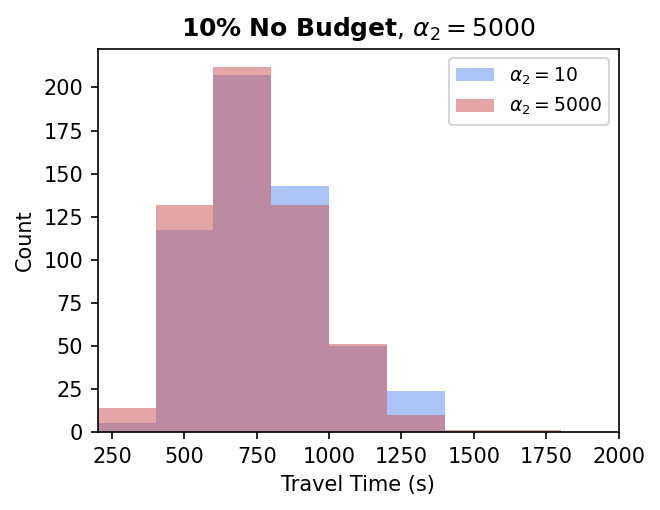}
        \caption{}\label{fig:time_10p_nobgt_a5000}
    \end{subfigure}\hfill
        \begin{subfigure}[b]{0.16\linewidth}
    \includegraphics[width=\linewidth]{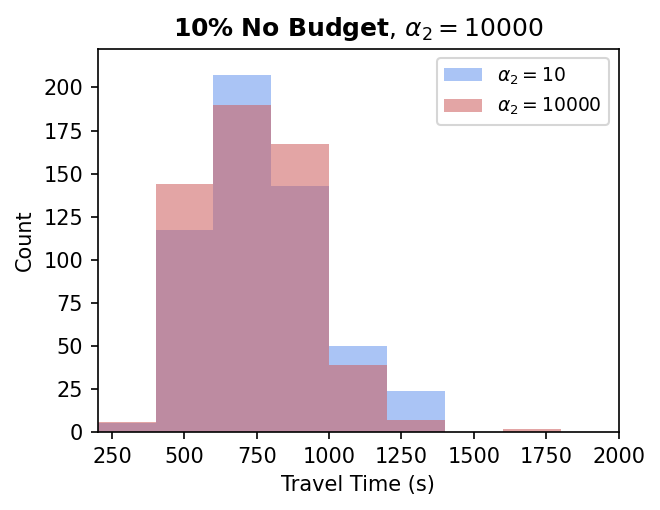}
        \caption{}\label{fig:time_10p_nobgt_a10000}
    \end{subfigure}\hfill
    \begin{subfigure}[b]{0.16\linewidth}
        \includegraphics[width=\linewidth]{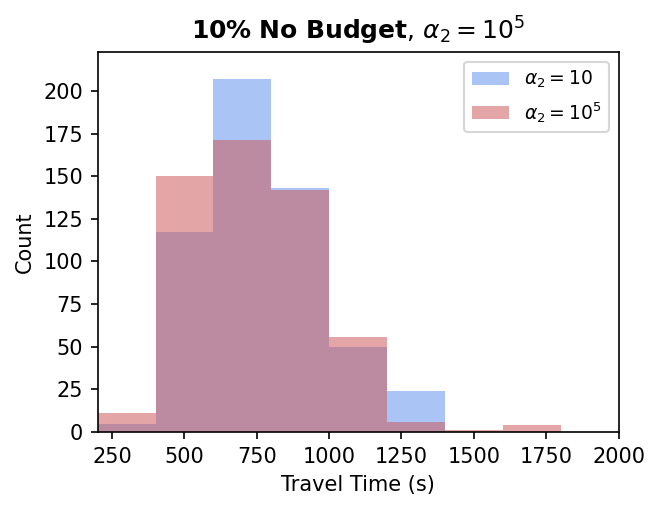}
        \caption{}\label{fig:time_10p_nobgt_a1e5}
    \end{subfigure}

    \vspace{0.4em}
    \noindent
    \begin{minipage}[t]{0.16\textwidth}\centering\footnotesize $\alpha_2 = 100$\end{minipage}\hfill
    \begin{minipage}[t]{0.16\textwidth}\centering\footnotesize $\alpha_2 = 1000$\end{minipage}\hfill
    \begin{minipage}[t]{0.16\textwidth}\centering\footnotesize $\alpha_2 = 3000$\end{minipage}\hfill
    \begin{minipage}[t]{0.16\textwidth}\centering\footnotesize $\alpha_2 = 5000$\end{minipage}\hfill
    \begin{minipage}[t]{0.16\textwidth}\centering\footnotesize $\alpha_2 = 10000$\end{minipage}\hfill
    \begin{minipage}[t]{0.16\textwidth}\centering\footnotesize $\alpha_2 = 10^{5}$\end{minipage}

    \caption{Robotaxi travel time distributions under different coverage weights across MPRs. The layout matches Figure~\ref{fig:distance_grid}. (a)--(f): 2\% MPR; (g)--(l): 5\% MPR; (m)--(r): 10\% MPR; (s)--(t): 10\% MPR without de-routing budget. The rightward shift of travel time distributions reflects both extended travel distances and reduced travel speeds at high $\alpha_2$ values.}
    \label{fig:traveltime_grid}
\end{figure}

As $\alpha_2$ increases, the routing algorithm experiences three main stages in sequence: (i)~route diversification, (ii)~distance extension, and (iii)~saturation. At low-to-moderate $\alpha_2$, the algorithm improves coverage by diversifying routes across the fleet without extending individual travel distances. With the increase of $\alpha_2$, longer travel distances become the major reason for improved coverage. When the second objective dominates (i.e., high $\alpha_2$ values), both coverage and travel distance converge to upper bounds due to the de-routing budget constraint. Figure~\ref{fig:distance_grid} and Figure~\ref{fig:traveltime_grid} illustrate these transitions through travel distance and travel time distributions at each MPR.

\begin{figure}[pos=tbp]
    \centering
    \includegraphics[width=\textwidth]{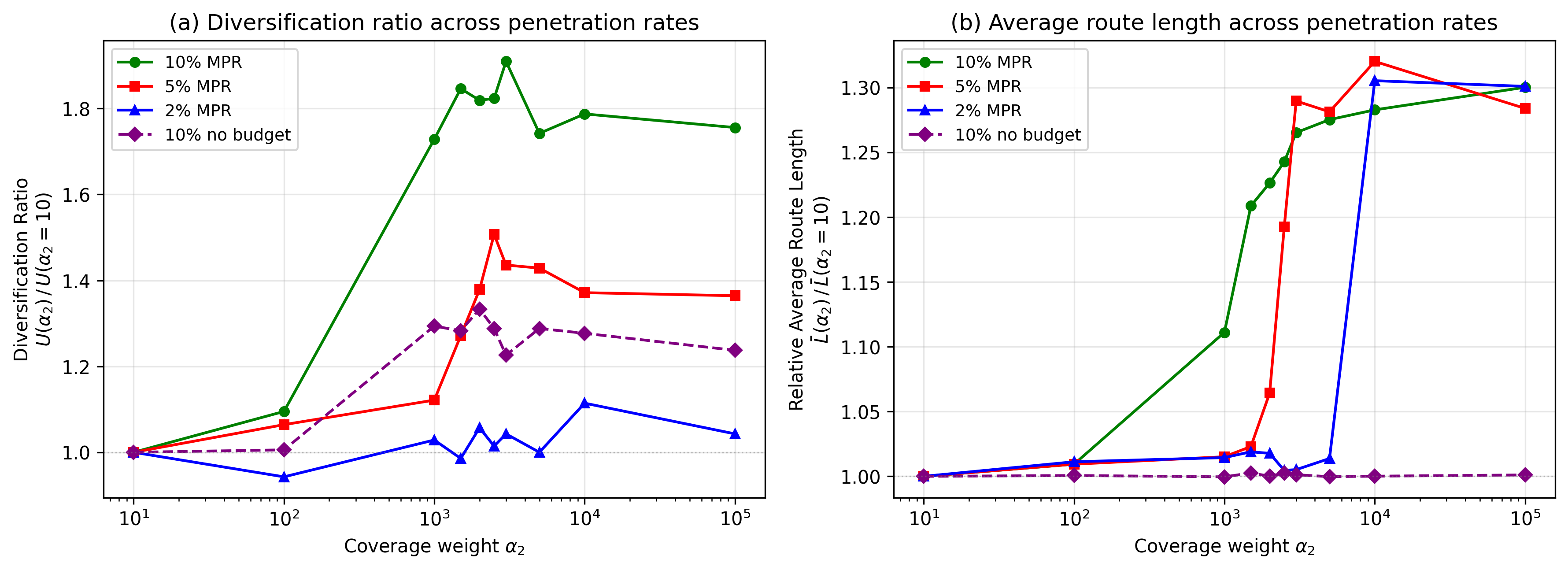}
    \caption{(a)~Diversification ratio $U(\alpha_2)/U(\alpha_2{=}10)$; (b)~Relative average route length $\bar{L}(\alpha_2)/\bar{L}(\alpha_2{=}10)$. Both panels report values relative to the $\alpha_2{=}10$ baseline across MPR settings and the no-budget variant.}
    \label{fig:route_diversity}
\end{figure}

To further quantify the route diversification under three stages, we calculate a diversification ratio, which is defined as the ratio between the number of distinct routes used by the fleet under a certain $\alpha_2$ value and the number of distinct routes used by the fleet when $\alpha_2=10$. The denominator represents the condition where the routes are the least diversified (e.g., travel time oriented shortest paths). The trend of the diversification ratio over $\alpha_2$ value is shown in Figure~\ref{fig:route_diversity}. Figure~\ref{fig:route_diversity} (a) shows that the diversification ratio increases with $\alpha_2$ under all MPRs. Higher MPRs lead to higher diversification ratios. This can be explained by the fact that a higher number of fleet vehicles can explore more route combinations given the same origins and destinations. Note that even without any de-routing budget, the algorithm tries to explore more distinct routes under higher $\alpha_2$ values. Figure~\ref{fig:route_diversity} (b) clearly shows the three stages of the routing strategies. Note that although the diversification ratios under three MPRs all reach a high plateau, the threshold $\alpha_2$ values differ: around 2000 for 10\%, around 5000 for 5\%, and around 10000 for 2\%. This is consistent with the sensing power analysis, where sensing powers also peak around these $\alpha_2$ values.

\subsection{Vehicle Number Weighted Monitoring Objective}
\label{sec:count-based_vehicle_coverage_extension}

In the previous section, we evaluate the model where the importance of each cell is treated equally. However, the number of vehicles within each cell can vary substantially over space and time. From the traffic monitoring perspective, cells containing more vehicles may provide greater observational value. 
Therefore, this section modifies the network coverage term of the original objective function to be vehicle number weighted, by replacing $\sum y_i^t$ with $\sum n_i^t\, y_i^t$ in Equation ~\eqref{eqa:obj_function}, where $n_{i}^{t}$ is the number of vehicles in cell $i$ at time step $t$. Since $n_i^t$ can be directly calculated by the CTM and serves as a constant coefficient multiplier, the model remains an MILP. 





\begin{figure}[pos=tbp]
    \centering
    \begin{subfigure}[b]{0.49\textwidth}
        \centering
        \includegraphics[width=\linewidth]{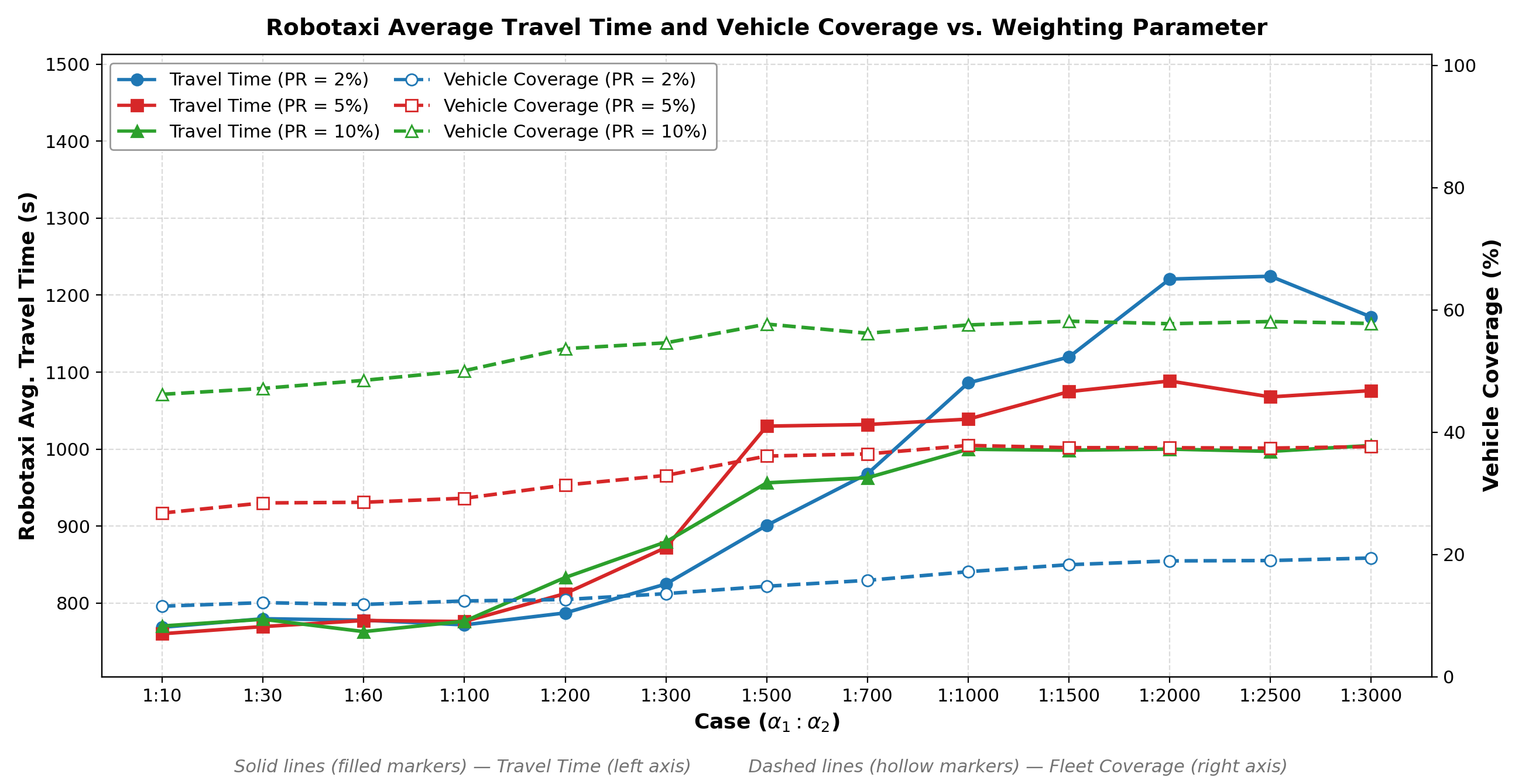}
        \caption{}
        \label{fig:weighted_tt_fc}
    \end{subfigure}
    \hfill
    \begin{subfigure}[b]{0.49\textwidth}
        \centering
        \includegraphics[width=\linewidth]{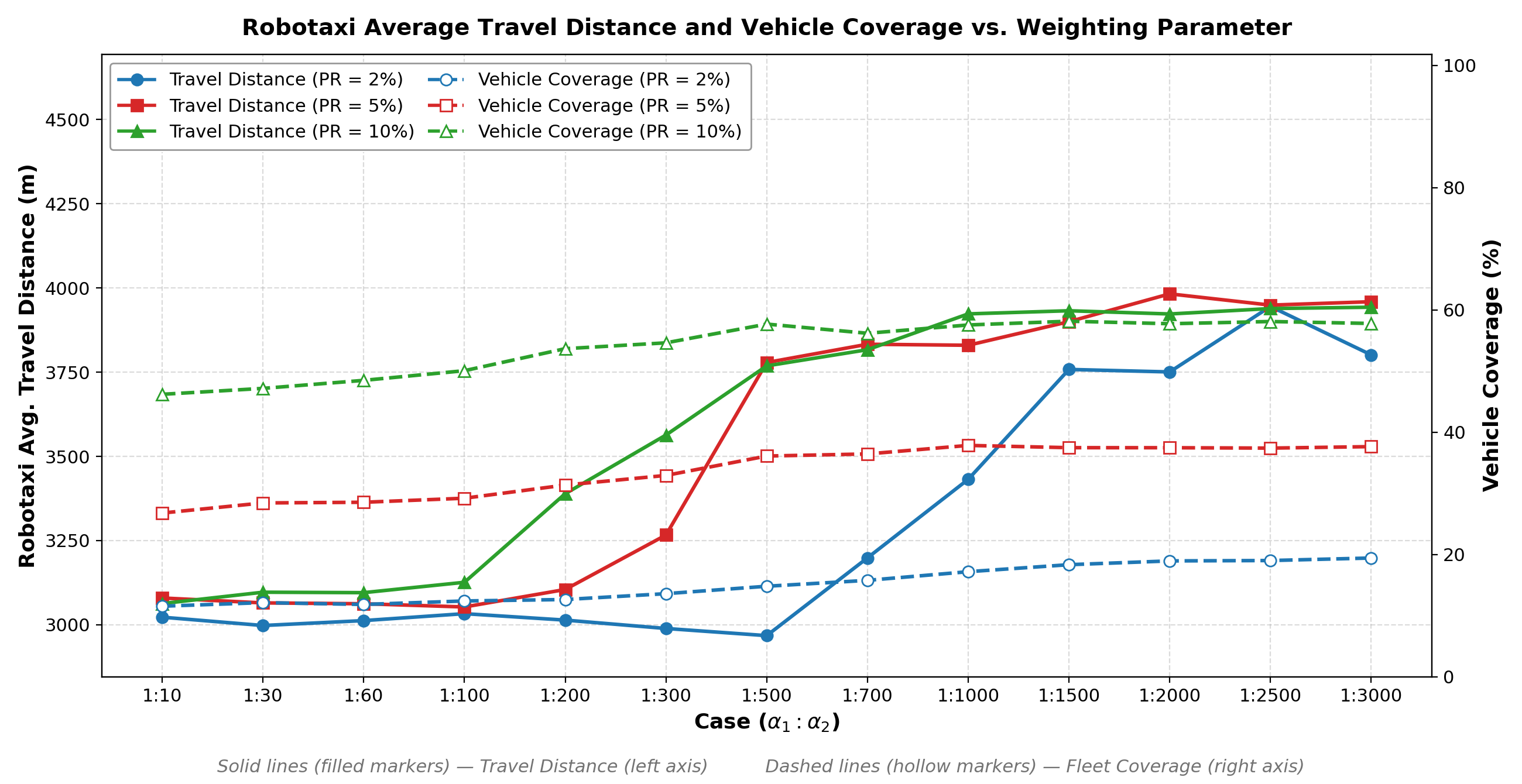}
        \caption{}
        \label{fig:weighted_dist_fc}
    \end{subfigure}
    \caption{Robotaxi average travel time (a) and average travel distance (b), each together with the fleet coverage, under the vehicle weighted objective.} 
    \label{fig:weighted_tt_dist_fc}
\end{figure}

We repeat the experiments with the updated formulation under the same settings and the results are shown in Table~\ref{Table:43_weighted_result}. 
Only the values of $\alpha_2$ are revised due to the magnitude of the modified coverage term in the objective function.
Compared with results in Table~\ref{Table:350_result}, new results show similar performance of the revised objective in increasing the number of observed vehicles with increasing values of $\alpha_2$. However, the average robotaxi travel time pattern shows a slight difference.
Figure~\ref{fig:weighted_tt_fc} plots the curves of average robotaxi travel time and vehicle coverage over different $\alpha_2$ values. The travel time still converges to a certain range between 1000s and 1200s. The converged values are more dispersed than those obtained under the original objective. Further investigation shows that this dispersion is mainly caused by the reduced robotaxi average speed. The vehicle weighted objective routes the robotaxis to more congested cells to increase the observed vehicle number. This is also evidenced by the higher vehicle coverage percentages across all robotaxi MPRs and $\alpha_2$ values. 
Especially when the size of the robotaxi fleet is small (i.e., MPR 2\%), the solution routes vehicles to the most congested links and therefore has the highest average travel time. At higher robotaxi MPRs, the assigned routes are more diverse with more vehicles observing less congested links.  

\begin{table}[H]
\caption{Experiment results under different robotaxi MPRs and objective weights with revised objective function}
\label{Table:43_weighted_result}
\centering
\setlength{\tabcolsep}{2pt}
\renewcommand{\arraystretch}{1.05}
\newcommand{\zcase}[1]{$1:#1$}
\resizebox{\textwidth}{!}{%
\begin{tabular}{llcccc|llcccc}
\hline\hline
\makecell{\textbf{Case} \\ $\alpha_1$:$\alpha_2$} & \textbf{MPR} &
\makecell{\textbf{Avg. TT} \\ \textbf{(s)}} &
\makecell{\textbf{Avg. Dist.} \\ \textbf{(m)}} &
\makecell{\textbf{Network} \\ \textbf{Coverage}} &
\makecell{\textbf{Vehicle} \\ \textbf{Coverage}} &
\makecell{\textbf{Case} \\ $\alpha_1$:$\alpha_2$} & \textbf{MPR} &
\makecell{\textbf{Avg. TT} \\ \textbf{(s)}} &
\makecell{\textbf{Avg. Dist.} \\ \textbf{(m)}} &
\makecell{\textbf{Network} \\ \textbf{Coverage}} &
\makecell{\textbf{Vehicle} \\ \textbf{Coverage}} \\
\hline
$1:10$   & $2\%$  & 768.49  & 3022.73 & 3.23\%  & 11.53\% & $1:10$   & $10\%$ & 770.17  & 3063.14 & 14.18\% & 46.18\% \\
$1:30$   & $2\%$  & 779.61  & 2997.98 & 3.33\%  & 12.11\% & $1:30$   & $10\%$ & 778.91  & 3096.80 & 14.54\% & 47.15\% \\
$1:60$   & $2\%$  & 777.68  & 3012.53 & 3.24\%  & 11.81\% & $1:60$   & $10\%$ & 762.86  & 3095.49 & 15.01\% & 48.47\% \\
$1:100$  & $2\%$  & 771.60  & 3033.14 & 3.37\%  & 12.38\% & $1:100$  & $10\%$ & 776.34  & 3126.29 & 15.58\% & 50.05\% \\
$1:200$  & $2\%$  & 787.15  & 3014.03 & 3.38\%  & 12.62\% & $1:200$  & $10\%$ & 833.15  & 3389.07 & 16.69\% & 53.65\% \\
$1:300$  & $2\%$  & 824.70  & 2989.19 & 3.52\%  & 13.57\% & $1:300$  & $10\%$ & 879.43  & 3563.31 & 17.26\% & 54.60\% \\
$1:500$  & $2\%$  & 900.89  & 2967.92 & 3.74\%  & 14.79\% & $1:500$  & $10\%$ & 956.17  & 3768.53 & 18.33\% & 57.67\% \\
$1:700$  & $2\%$  & 967.94  & 3198.71 & 3.88\%  & 15.75\% & $1:700$  & $10\%$ & 962.78  & 3816.57 & 17.99\% & 56.17\% \\
$1:1000$ & $2\%$  & 1086.01 & 3431.86 & 4.21\%  & 17.17\% & $1:1000$ & $10\%$ & 999.74  & 3922.88 & 18.55\% & 57.53\% \\
$1:1500$ & $2\%$  & 1119.60 & 3758.02 & 4.56\%  & 18.31\% & $1:1500$ & $10\%$ & 998.52  & 3932.16 & 18.72\% & 58.14\% \\
$1:2000$ & $2\%$  & 1220.88 & 3750.58 & 4.69\%  & 18.93\% & $1:2000$ & $10\%$ & 999.98  & 3922.60 & 18.47\% & 57.73\% \\
$1:2500$ & $2\%$  & 1224.54 & 3944.90 & 4.83\%  & 18.99\% & $1:2500$ & $10\%$ & 997.01  & 3938.71 & 18.52\% & 58.09\% \\
$1:3000$ & $2\%$  & 1171.61 & 3801.22 & 4.75\%  & 19.41\% & $1:3000$ & $10\%$ & 1004.67 & 3942.68 & 18.55\% & 57.77\% \\
\hline
$1:10$   & $5\%$ & 760.08  & 3079.70 & 7.86\%  & 26.76\% & \zcase{10}   & $10\%$ no budget & 763.21 & 2944.85 & 13.65\% & 45.00\% \\
$1:30$   & $5\%$ & 769.40  & 3065.06 & 8.03\%  & 28.41\% & \zcase{30}   & $10\%$ no budget & 744.06 & 2951.71 & 13.92\% & 45.60\% \\
$1:60$   & $5\%$ & 777.30  & 3062.55 & 8.17\%  & 28.53\% & \zcase{60}   & $10\%$ no budget & 730.47 & 2945.66 & 14.17\% & 46.43\% \\
$1:100$  & $5\%$ & 775.97  & 3053.04 & 8.22\%  & 29.19\% & \zcase{100}  & $10\%$ no budget & 743.48 & 2941.93 & 14.86\% & 48.75\% \\
$1:200$  & $5\%$ & 812.27  & 3104.86 & 8.67\%  & 31.34\% & \zcase{200}  & $10\%$ no budget & 746.85 & 2935.31 & 15.15\% & 50.45\% \\
$1:300$  & $5\%$ & 872.17  & 3266.53 & 8.99\%  & 32.92\% & \zcase{300}  & $10\%$ no budget & 747.62 & 2938.96 & 15.33\% & 51.27\% \\
$1:500$  & $5\%$ & 1029.82 & 3778.47 & 10.15\% & 36.09\% & \zcase{500}  & $10\%$ no budget & 745.59 & 2934.16 & 15.23\% & 50.86\% \\
$1:700$  & $5\%$ & 1031.91 & 3832.75 & 10.26\% & 36.43\% & \zcase{700}  & $10\%$ no budget & 747.58 & 2941.08 & 15.25\% & 51.43\% \\
$1:1000$ & $5\%$ & 1038.90 & 3829.74 & 10.39\% & 37.82\% & \zcase{1000} & $10\%$ no budget & 745.26 & 2943.14 & 15.17\% & 50.45\% \\
$1:1500$ & $5\%$ & 1074.67 & 3899.78 & 10.47\% & 37.45\% & \zcase{1500} & $10\%$ no budget & 749.03 & 2935.19 & 15.14\% & 50.69\% \\
$1:2000$ & $5\%$ & 1088.46 & 3982.31 & 10.49\% & 37.45\% & \zcase{2000} & $10\%$ no budget & 742.06 & 2932.57 & 15.05\% & 50.24\% \\
$1:2500$ & $5\%$ & 1067.79 & 3948.87 & 10.48\% & 37.38\% & \zcase{2500} & $10\%$ no budget & 749.45 & 2935.53 & 15.35\% & 51.20\% \\
$1:3000$ & $5\%$ & 1076.01 & 3959.01 & 10.52\% & 37.63\% & \zcase{3000} & $10\%$ no budget & 741.84 & 2926.03 & 15.03\% & 50.54\% \\
\hline\hline
\end{tabular}%
}
\end{table}

\begin{figure}[pos=tbp]
    \centering
    \begin{subfigure}[b]{0.49\textwidth}
        \centering
        \includegraphics[width=\linewidth]{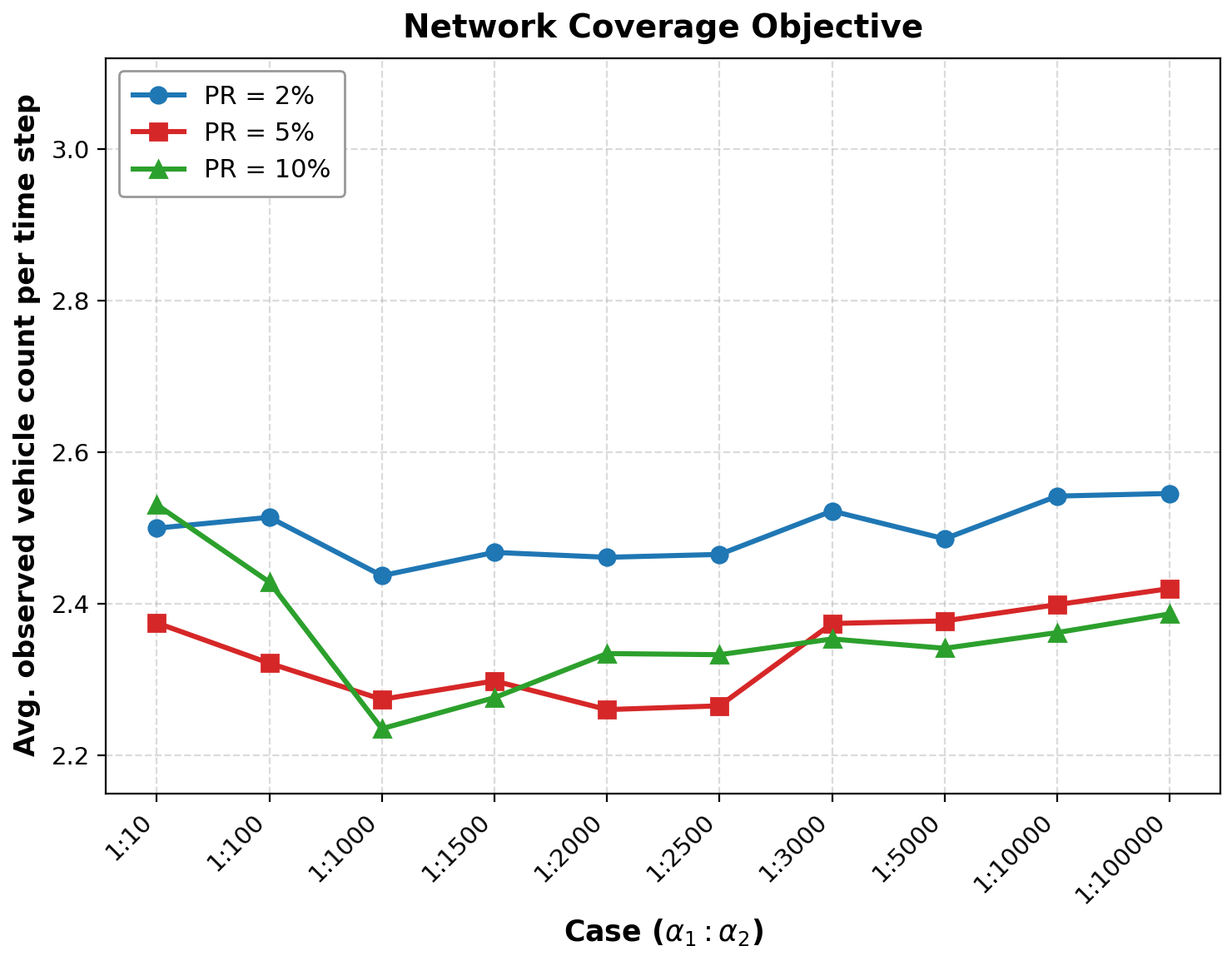}
        \caption{}
        \label{fig:route_traffic_countbased}
    \end{subfigure}
    \hfill
    \begin{subfigure}[b]{0.49\textwidth}
        \centering
        \includegraphics[width=\linewidth]{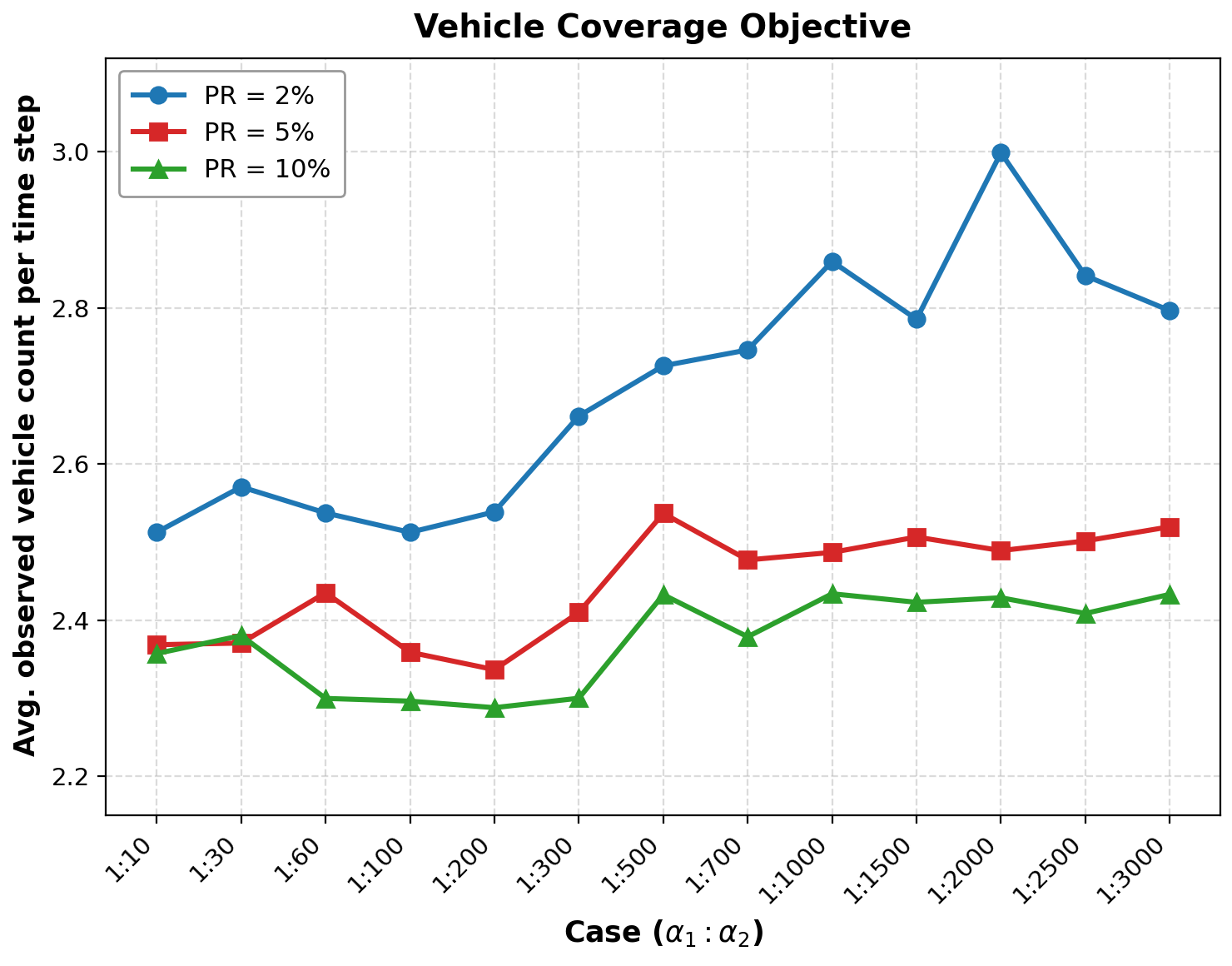}
        \caption{}
        \label{fig:route_traffic_weighted}
    \end{subfigure}
    \caption{Average observed vehicle count per time step under (a) cell based coverage objective, and (b) vehicle weighted coverage objective.} 
    \label{fig:route_traffic}
\end{figure}

We further calculate the average number of observed vehicles per time step for each robotaxi during the entire trip, and the results are shown in Figure~\ref{fig:route_traffic}, with two sub figures representing results from the cell based objective and the vehicle weighted objective, respectively. 
Under the cell based objective, the average number of observed vehicles remains mostly constant under different $\alpha_2$ values (Figure~\ref{fig:route_traffic_countbased}), while under the vehicle weight objective, the observed vehicle number increases with $\alpha_2$ (Figure~\ref{fig:route_traffic_weighted}). Overall, the vehicle weight objective results in higher average observed vehicle numbers crossing all robotaxi MPRs.

In summary, this modification can redirect the sensing effort from cell-time to vehicle-time observations. This allows future deployment to apply task specific adjustments based on the value of observations determined by traffic conditions and policy preferences.


\section{Implications of Real-world Deployment}
\label{sec:implication}

A practical motivation for applying the routing for monitoring framework to robotaxis is the observation that they spend a non-trivial fraction of time without serving passengers, while the same vehicles already carry the perception and communication hardware needed for traffic monitoring. Operating these vehicles as a coordinated sensing layer is therefore an underutilized opportunity. Even when serving passengers, incentives could be offered to encourage passengers to accept routes with slightly longer travel time or travel distance while improving traffic monitoring performance. Such incentive based routing strategies have been studied in literature \citep{benelia2011changing, yang2011managing} and implemented in real world (e.g., Metropia \citep{metropia}). Moreover, monitoring and mobility objectives are not strictly competing. Under certain conditions, the robotaxi average travel time decreases with improved network coverage, creating a win-win situation. The main reason is that the improved coverage reduces errors in travel time prediction, which further leads to better route selections. As a result, robotaxi companies would be willing to contribute their vehicles as drive-by sensors while improving service quality.

\section{Conclusion and Future Work}
\label{sec:Conclusion and Future Work}
In this paper, we proposed a novel routing for monitoring framework for robotaxi fleets that integrated microscopic simulation, traffic state estimation, and dynamic vehicle routing. A SUMO-based simulation environment was developed to simulate vehicle movements for both robotaxis and background vehicles. A cell-based network typology was applied to support both traffic state estimation and vehicle-level detection and monitoring capabilities. The Cell Transmission Model (CTM) was employed to predict traffic states and generate time-dependent travel costs for routing optimization. An MILP formulation was then developed to optimize robotaxi routes under a rolling horizon scheme. The objective function considered both robotaxi total travel time and spatiotemporal network coverage, which represents traffic monitoring performance. A case study was conducted on a $5\times 5$ grid network with symmetric O-D demand and fixed-time traffic signals. Three robotaxi MPRs were evaluated: $2\%$, $5\%$, and $10\%$, each tested with multiple weight parameter combinations. 
Results showed that different weights of the network coverage objective could effectively change the routing strategy. When proper weights are assigned between the two objectives, the average travel time of robotaxis is also reduced, possibly because increased network coverage could improve the accuracy of the traffic state prediction model, which further leads to more accurate time dependent travel time calculation. We further showed that increasing coverage does not monotonically improve the fleet's sensing power. Once the de-routing budget pushes the routing into distance extension only, fleet speeds drop and sensing power degrades, indicating that an effective deployment must balance coverage uniformity, fleet speed, and travel time penalty rather than only maximizing coverage.

This work sets the foundation for robotaxi as drive-by sensors and opens several paths for further extensions, including 1) applying more realistic vehicle perception models to consider uncertainties in the observations; 
2) incorporating robotaxi fleet operations (e.g., passenger pick-up and drop-off) in the routing framework; 
and 3) integrating other traffic monitoring sensing modalities (e.g., fixed location sensors). 
These extensions are particularly important for moving the current work beyond the simplified settings towards a more realistic and deployable framework.


\section{Acknowledgments}
This research is supported in part by the U.S. National Science Foundation (NSF) through Grant CMMI \#2339753, and the U.S. Department of Transportation (USDOT) Region 5 University Transportation Center (Center for Connected and Automated Transportation). The views presented in this paper are those of the authors alone.

\section{CRediT authorship contribution statements}
Yilin Wang: Writing – review \& editing, Writing – original draft, Visualization, Validation, Methodology, Investigation, Formal analysis, Data curation, Conceptualization. Yiheng Feng: Writing – review \& editing, Project administration, Methodology, Investigation, Funding acquisition, Conceptualization.

\section{Declaration of competing interest}
The authors declare that they have no known competing financial interests or personal relationships that could have appeared to influence the work reported in this paper.

\bibliographystyle{cas-model2-names}

\bibliography{reference}

\end{document}